\documentclass{article} 
\usepackage{iclr_preprint,times}

\usepackage{amsmath,amsfonts,bm}

\def\eqref#1{equation~\ref{#1}}

\def\1{\bm{1}}

\DeclareMathAlphabet{\mathsfit}{\encodingdefault}{\sfdefault}{m}{sl}
\SetMathAlphabet{\mathsfit}{bold}{\encodingdefault}{\sfdefault}{bx}{n}

\newcommand{\softmax}{\mathrm{softmax}}

\DeclareMathOperator{\sign}{sign}

\usepackage{hyperref}
\usepackage{url}
\usepackage{amsmath, amsthm}
\usepackage{array}
\usepackage{booktabs}
\usepackage{graphicx}
\usepackage[capitalise]{cleveref} 
\usepackage{caption}   
\usepackage{subcaption}
\usepackage{makecell}
\newtheorem{corollary}{Corollary}

\newtheorem{definition}{Definition}
\newtheorem{proposition}{Proposition}
    
\title{Elucidating the Design Space of FP4 training}

\author{Robert Hu \& Carlo Luschi \& Paul Balanca \\
Graphcore\\
London, UK \\
\texttt{\{roberthu, carlo, paulb\}@graphcore.ai} \\
}

\iclrfinalcopy 
\begin{document}

\maketitle

\begin{abstract}
The increasing computational demands of foundation models have spurred research into low-precision training, with 4-bit floating-point (\texttt{FP4}) formats emerging as a frontier for maximizing hardware throughput. While numerous techniques have been proposed to stabilize \texttt{FP4} training, they often present isolated solutions with varying, and not always clear, computational overheads. This paper aims to provide a unified view of the design space of \texttt{FP4} training. We introduce a comprehensive, quantisation gradient-based framework for microscaling quantization that allows for a theoretical analysis of the computational costs associated with different stabilization methods on both the forward and backward passes. Using a simulator built on this framework, we conduct an extensive empirical study across a wide range of machine learning tasks, including regression, image classification, diffusion models, and language models. By systematically evaluating thousands of combinations of techniques—such as novel gradient approximations, rounding strategies, and scaling methods, we identify which configurations offer the most favourable performance-to-overhead trade-off. We find that the techniques enabling the best trade-off involve carefully combining Hadamard transformations, tensor scaling and stochastic rounding. We further find that using \texttt{UE5M3} as a scaling factor potentially offers a good compromise between range and precision with manageable computational overhead.

\end{abstract}

\section{Introduction}

With the emergence of foundation models \citep{bommasani2021opportunities}, the demand for computational resources has grown proportionally to the parameter count of these models, which often span from billions to trillions of parameters. Many of these models rely on the transformer architecture \citep{vaswani2017attention}, which is ubiquitous across vision, text and video models \citep{Khan_2022}. These models tend to be compute bound by two operations: the attention mechanism \citep{duman2023computational, dolga2024latte} which tends to scale quadratically with sequence length, and the matmul operations from the weight matrices \citep{eleks2025quadraticattention} which are quadratic with respect to size of the hidden dimensions. 

In this paper, we conduct the first large-scale, systematic investigation of the \texttt{FP4} design space. Traditionally, machine learning training uses \texttt{FP32} format which serves as the baseline with the highest accuracy and lowest throughput. Historically, with each subsequent generation of hardware, the precision has been halved — starting from 2016 \texttt{FP16} \citep{micikevicius2018mixedprecisiontraining}, 2022 \texttt{FP8} \citep{noune20228bitnumericalformatsdeep,micikevicius2022fp8formatsdeeplearning,fishman2025scalingfp8trainingtrilliontoken}, and now 2025 with \texttt{FP4} \citep{chmiel2025fp4wayfullyquantized, castro2025quartetnativefp4training}. Halving the precision often allows a doubling of the throughput for matrix multiplication operations \citep{hao2025lowprecisiontraininglargelanguage}, hence with each iteration careful adjustments need to be made to the training procedure to account for the loss of numerical accuracy \citep{tseng2024quipbetterllmquantization, li2025svdquantabsorbingoutlierslowrank}. There have been several previous works accounting for \texttt{FP16} \citep{micikevicius2018mixedprecisiontraining}, \texttt{FP8} \citep{fishman2025scalingfp8trainingtrilliontoken}, and recently \texttt{FP4} \citep{tseng2025trainingllmsmxfp4, chen2025oscillationreducedmxfp4trainingvision,castro2025quartetnativefp4training,hao2025lowprecisiontraininglargelanguage, chmiel2025fp4wayfullyquantized,wang2025optimizinglargelanguagemodel,11038348,su2025characterizationmitigationtraininginstabilities,li2025svdquantabsorbingoutlierslowrank,cao2025metistraininglargelanguage}. \newline While these works introduce new techniques to stabilize \texttt{FP4} training for larger models, they all propose different methodologies with varying computational overhead that are empirically shown to work in isolation through simulations in \texttt{BFLOAT16}. However, a systematic evaluation of the performance-overhead trade-offs has been missing. As an example, \cite{wang2025optimizinglargelanguagemodel} introduces a quantile based pruning and gradient adjustment, both of which are shown to be useful, however both add an additional $\sim\mathcal{O}(n)$ time and memory overhead which cannot be done in low-precision and is non-fusable. It should be further noted that the simulation procedure in \cite{wang2025optimizinglargelanguagemodel} does not adequately quantise the scale, which their description implies is kept in high-precision -- a detail that can significantly impact training stability. Similarly, \cite{tseng2025trainingllmsmxfp4} proposes a block-wise Hadamard transformation, which induces a $\sim\mathcal{O}(n \log l)$ overhead. It should be noted that none of the aforementioned papers simulate \texttt{FP4} training fully, as \cite{wang2025optimizinglargelanguagemodel} only consider quantising weights and activations and not the gradient and \cite{tseng2025trainingllmsmxfp4} only quantises the gradient. Further,  \cite{cao2025metistraininglargelanguage,li2025svdquantabsorbingoutlierslowrank} proposes spectral decomposition techniques to handle outliers, which consequently introduces $\mathcal{O}(mnk)$ time and $\mathcal{O}(k^2)$ memory, which in terms of hardware acceleration is also non-fusable. It is currently not clear whether these additional overheads are necessary in downstream implementations of low-precision matrix multiplication, as some evidence in \cite{chmiel2025fp4wayfullyquantized,11038348} suggests that something as simple as \emph{Stochastic Rounding (SR)} is enough to stabilise \texttt{FP4} training.\newline The goal of this work is to develop a thorough understanding of the quantisation mechanism and how it affects the training procedure and illuminate which techniques offer a worthwhile trade-off in terms of additional overhead vs performance benefit. We summarise the contributions of this paper as follows:

\begin{enumerate}
    \item We propose a quantisation gradient-based framework for \texttt{FP4} quantisation, which is used to derive the computational overhead of conceivably useful techniques (both novel and existing ones) on the forward and backward pass of a quantised linear layer.
    \item We implement the framework as a simulator, running experiments of across various machine learning tasks to gain insight on which combination of techniques offer a reasonable overhead vs. performance benefit. 
\end{enumerate}

We first introduce our unified, gradient-based framework in Section 2, then use it to analyze the design space of scaling, rounding, and gradient approximation techniques in Sections 3, 4, and 5. We survey other relevant methods in Section 6, present our extensive empirical validation in Section 7, and conclude with our key findings.

\section{A common framework for \texttt{FP4} training}

In this section, we detail what happens when we use microscaling formats to quantize a tensor for a linear layer forward pass. Consider a tensor \( \mathbf{X} \in \mathbb{R}^{m \times n} \). We first define $\mathbf{X}$ represented in microscaling format \cite{rouhani2023microscalingdataformatsdeep}.

\begin{definition}
A \emph{micro-scaled block} is defined by a scalar \( s \in \mathbb{R} \) and a vector \( \mathbf{P} = [p_i]_{i=1}^l \) of \( l \) elements. Each value \( x_i \) can be recovered as
\[
x_i = s \cdot p_i.
\]
The parameter \( l \) is a fixed constant known as the \emph{block size}. Given a tensor \( \mathbf{X} \in \mathbb{R}^{m \times n} \) and a block size of \( l  \), the \emph{MX representation} of \( \mathbf{X} \) consists of a collection of tuples
\[
\{ (s_j, \mathbf{P}_j) \}_{j=1}^{(m\cdot n) // l},
\]
where each tuple corresponds to a block of $l$ elements in \( \mathbf{X} \).
\end{definition}
Intuitively, microformat scaling represents partitions of a tensor with a common scale often used to normalise the partition, where the scaled elements $\mathbf{P}_j$ is quantized to a lower precision. We formally detail the quantisation procedure for one partition $\mathbf{X}_p \in \mathbb{R}^l$ below, represented by a transformation $f$:
\[
f(\mathbf{X}_p) = \frac{1}{s_q} Q(s_q \cdot \mathbf{X}_p)
\]
where the components are defined as follows:

\textbf{Outer Scaling Factor ($s_q$):} This factor is a function of $\mathbf{X}_p$. First, an intermediate factor $s(\mathbf{X}_p)$ is computed:
    \[
    s(\mathbf{X}_p) = \frac{\texttt{FP4 max}}{Z(\mathbf{X}_p)}
    \]
    Here, $Z: \mathbb{R}^{l} \to \mathbb{R}$ is a scalar-valued function of the tensor $\mathbf{X}_p$ (e.g., the absolute maximum norm, \texttt{absmax}). This factor is then quantized:
    \[
    s_q = q(s(\mathbf{X}_p))
    \]
    
\textbf{Quantization Function ($Q$):} $Q$ is an element-wise function that quantizes the elements of $\mathbf{X}_p$.

We can now introduce the gradient with respect to an element in $\mathbf{X}$:

\begin{proposition}
Let $f(\mathbf{X}_p) = \frac{1}{s_q} Q(s_q \cdot \mathbf{X}_p)$
 Then, the partial derivative of \( f_{ij} \) with respect to \( \mathbf{X}_{ij} \) is given by:
\begin{equation}
\boxed{
\frac{\partial f_{ij}}{\partial \mathbf{X}_{ij}} = Q'(s_q \mathbf{X}_{ij}) + \frac{\partial s}{\partial \mathbf{X}_{ij}} \left[ \frac{q'(s)}{s_q} \left( \mathbf{X}_{ij} Q'(s_q \mathbf{X}_{ij}) - \frac{1}{s_q} Q(s_q \mathbf{X}_{ij}) \right) \right]
}
\end{equation}
See \Cref{derivation_1} for derivations.
\end{proposition}

In the context of a linear layer with weights \( \mathbf{W} \in \mathbb{R}^{n \times m} \) and input data \( \mathbf{X} \in \mathbb{R}^{b \times m} \), the output $\mathbf{Y} = f(\mathbf{X}) f(\mathbf{W})^{\top}$ would have the corresponding gradients:
\[
\frac{\partial \mathcal{L}}{\partial \mathbf{X}} = \left(\frac{\partial \mathcal{L}}{\partial \mathbf{Y}} \cdot f(\mathbf{W})\right) \odot \frac{\partial f(\mathbf{X})}{\partial \mathbf{X}}, \quad
\frac{\partial \mathcal{L}}{\partial \mathbf{W}} = \left( \left( \frac{\partial \mathcal{L}}{\partial \mathbf{Y}} \right)^{\top} \cdot f(\mathbf{X})\right) \odot \frac{\partial f(\mathbf{W})}{\partial \mathbf{W}}.
\]

Here, \( \mathcal{L} \) denotes the scalar loss, $\odot$ the elementwise product and \( f(\cdot) \) is a differentiable transformation (e.g., MX decomposition or quantization-aware mapping) applied to the inputs and weights. In the next section, we detail different choices in terms of calculating and approximating $\frac{\partial \mathcal{L}}{\partial \mathbf{X}},\frac{\partial \mathcal{L}}{\partial \mathbf{W}}$. We summarise the time and memory overhead of our proposed and existing techniques in \Cref{tab:techniques_summary}.

\section{Tensor scaling in \texttt{FP4} training}

An alternative to applying block-wise scaling directly is to first normalize the entire tensor \cite{blake2023unitscalingoutoftheboxlowprecision,micikevicius2022fp8formatsdeeplearning,NEURIPS2019_65fc9fb4, peng2023fp8lmtrainingfp8large}. The goal of this strategy is to improve the quantization of the scaling factors themselves. In this approach, a tensorwise scaling factor $g$ is computed, used to normalize the tensor, and then multiplied back after the block-wise quantization. While the intent is for $g$ to cancel out, the non-linear nature of the scale quantization function $q(\cdot)$ results in a distinct final transformation.

Let $g = \max_p{\{m_p\}},\quad m_p= Z(\mathbf{X}_p)$ be the global scaling factor for a tensor $\mathbf{X}$, and let $\mathbf{U} = \mathbf{X}/g$ be the globally normalized tensor. The transformation $h(\mathbf{X})$ for an element $\mathbf{X}_{ij}$ within a block $p$ is defined as $h_{ij}(\mathbf{X}_p) = g \cdot f_{ij}(\mathbf{U}_p)$.

Here, $f(\mathbf{U}_p)$ is the block-wise quantization function from Section 2 applied to the normalized block $\mathbf{U}_p$. Its components are functions of $\mathbf{U}_p$:
\begin{enumerate}
    \item \textbf{Ideal Scale:} $s'_p = \frac{\texttt{FP4}_{\text{max}}}{Z(\mathbf{U}_p)} = g \cdot \frac{\texttt{FP4}_{\text{max}}}{Z(\mathbf{X}_p)} = g \cdot s_p$
    \item \textbf{Quantized Scale:} $s'_{q,p} = q(s'_p) = q(g \cdot s_p)$
\end{enumerate}
Substituting these gives the full forward pass expression for an element:
\begin{equation}
\boxed{
h_{ij}(\mathbf{X}) = \frac{g}{q\left(g \cdot s_p\right)} Q\left(q(g \cdot s_p) \cdot \frac{\mathbf{X}_{ij}}{g}\right)
}
\end{equation}
The gradient of this transformation accounts for both the block-wise dependencies and the global dependency on $g$.

\begin{corollary}
\label{coro_1}
    Let $h(\mathbf{X})$ be the quantization with intermediate global normalization. For an element $\mathbf{X}_{ij}$ in block $p$, the partial derivative is:
\begin{equation}
\boxed{
\frac{\partial h_{ij}}{\partial \mathbf{X}_{ij}} = \frac{\partial f_{ij}}{\partial \mathbf{U}_{p,ij}} + \frac{\partial g}{\partial \mathbf{X}_{ij}} \left( f_{ij}(\mathbf{U}_p) - \mathbf{U}_{p,ij} \frac{\partial f_{ij}}{\partial \mathbf{U}_{p,ij}} \right)
}
\end{equation}
where $\frac{\partial f_{ij}}{\partial \mathbf{U}_{p,ij}}$ is the full gradient from the first Theorem, evaluated on the normalized block $\mathbf{U}_p$ with its corresponding scales ($s'_p, s'_{q,p}, k'_p$). See \Cref{derivation_2} for derivations.
\end{corollary}

It should be noted that $g$ in context of the \texttt{MXFP4}(\texttt{FP4} format with \texttt{E8M0} scale) and \texttt{NVFP4} (\texttt{FP4} format with \texttt{E4M3} scale) has overhead complexity $\mathcal{O}((m\cdot n)//l)$ as opposed to $\mathcal{O}(m\cdot n)$ when computing $\texttt{absmax}(\mathbf{X})$, since it suffices to search the scales rather than the entire tensor $\mathbf{X}$.

\subsection{Rounding and Scaling Strategies}

Recent work in low-precision training has highlighted that the rounding strategy for scaling factors can have a profound impact on model stability. For instance, \cite{mishra2025recipespretrainingllmsmxfp8} found that for \texttt{MXFP8} formats, rounding-to-positive-infinity improves signal propagation by reducing the number of saturated values, given the limited range of the scaling factor. As our work considers both \texttt{E4M3} (which has a limited range) and \texttt{E8M0} (which has a wider range), we evaluate both round-to-nearest (RTN) and round-to-positive-infinity in our experiments.

We note that the results in \cite{chmiel2025fp4wayfullyquantized} get \texttt{NVFP4} to converge without any issues with tensor scaling, as they mitigate any overflow by taking $ \tilde s_p = \frac{s_p'}{\texttt{FP4}_{\text{max}}\cdot\texttt{E4M3} _{\text{max}}\cdot0.5}$. This pushes down the effective range of $ \tilde s_p\in[2/\texttt{E4M3}_{\text{max}}, 2/\texttt{E4M3}_{\text{max}} \cdot g)$. While not completely protected from overflow, it's a good rule of thumb to maximize the utilised range of \texttt{E4M3}. We use this technique when applying tensor scaling for \texttt{NVFP4}. We note that this heuristic can be extended to any scale format beyond \texttt{E4M3}, as it effectively rescales the scale factor to utilise its maximum range. 

\textbf{Handling Zero-Valued Scales.} A critical edge case is the handling of zero-valued scaling factors, resulting in division by zero in the dequantisation. \cite{chmiel2025fp4wayfullyquantized} replaces any zero scale with one, which may induce further quantisation errors as small scales are set to 1. We propose rounding the zeros and underflows to the closest representable subnormal value in the target format and saturate overflows to the maximum representable number. We compare the efficacy of both approaches in our experiments.



\textbf{Rounding of the weight tensor} The impact of the rounding strategy has previously been demonstrated in \cite{chmiel2025fp4wayfullyquantized,fitzgibbon2025stochasticroundingrandombits} to have significant impact on the stability of LLM training in low-precision format. The main observations for \texttt{FP4} formats is to use \emph{round-to-nearest} (RTN) for the forward pass and \emph{stochastic rounding} (SR) in backwards pass \citep{chmiel2025fp4wayfullyquantized,11038348}, specifically on the activation and gradient tensors. We follow the quantisation procedure in \cite{rouhani2023microscalingdataformatsdeep}, which considers 6 quantisations for a forward and backward pass in a linear layer. We benchmark against the proposed strategy in \cite{chmiel2025fp4wayfullyquantized} and additionally consider SR on the activations in the forward pass as well.

\textbf{Rounding of the scales} We also experiment with stochastic rounding in the scaling factor as well. We motivate this design choice with the observation that \texttt{E8M0} has very large intervals between each number, leading to potential bias, which can mitigated more effectively at the scaling factor.

\section{Differentiable Relaxations for Quantization}

\textbf{Approximating $Q'(\mathbf{X})$ and $q'(x)$} In \cite{wang2025optimizinglargelanguagemodel}, they take $\frac{\partial f_{ij}}{\partial \mathbf{W}_{ij}} \approx Q'(s\cdot \mathbf{W}_{ij})$. However since $Q$ is a quantisation function which is not differentiable, they approximate $Q(x)\approx \frac{\delta}{2} \cdot\left(1+\operatorname{sign}\left(\frac{2 x}{\delta}-1\right) \cdot\left|\frac{2 x}{\delta}-1\right|^{\frac{1}{w}}\right)$, with gradient $Q'(x)=\frac{1}{w} \cdot\left|\frac{2 x}{\delta}-1\right|^{\frac{1}{w}-1}$. They propose $w=5$ in their implementation. There are some potential flaws with the proposed parametrisation, as calculating power of fractionals tends to be computationally expensive and require $\mathcal{O}(w)$ cycles. This leads to the overall complexity of $\mathcal{O}(nmw\log_2(k) )$, where $\log(k)$ comes from finding the interval $x$ belongs to on the \texttt{E2M1} grid using binary search, with $k$ being the grid size. We thus propose an alternative differentiable relaxation to $Q(x)$.


\textbf{Linear Spline approximation} A linear spline is a continuous piecewise linear function defined over a set of sorted knots $t_0, \dots, t_n$. These knots partition the domain into $n$ intervals $I_i = [t_i, t_{i+1})$. The function's continuity is ensured by having the linear segments connect at the knots.

The forward and backward passes evaluate the spline and its derivative. For an input $x \in [t_i, t_{i+1})$, the spline is a line segment, $S(x) = a_i (x - t_i) + b_i$ (\textbf{Forward pass}) with $S'(x) = a_i$ (\textbf{Backward pass}). Here, $b_i$ represents the value of the spline at knot $t_i$ (i.e., $S(t_i)$), and $a_i$ is the slope of the line segment over the interval $[t_i, t_{i+1})$.

\begin{figure}[htbp]
    \centering 

    \begin{subfigure}[b]{0.32\linewidth}
        \centering
        \includegraphics[width=\linewidth]{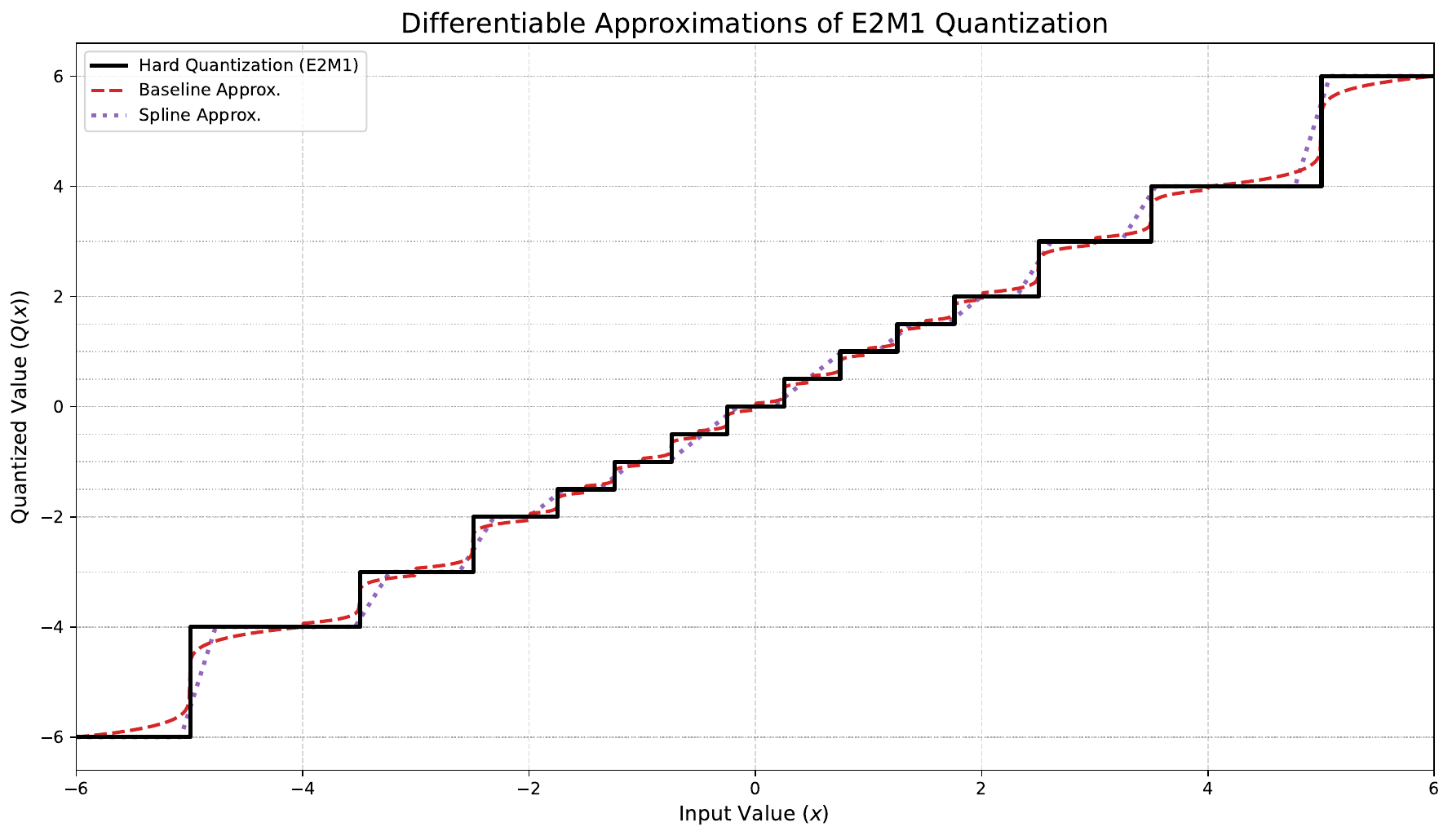}
        \caption{Approx. function $Q_{\text{approx}}(x)$.}
        \label{subfig:quant_func}
    \end{subfigure}
    \hfill 
    \begin{subfigure}[b]{0.32\linewidth}
        \centering
        \includegraphics[width=\linewidth]{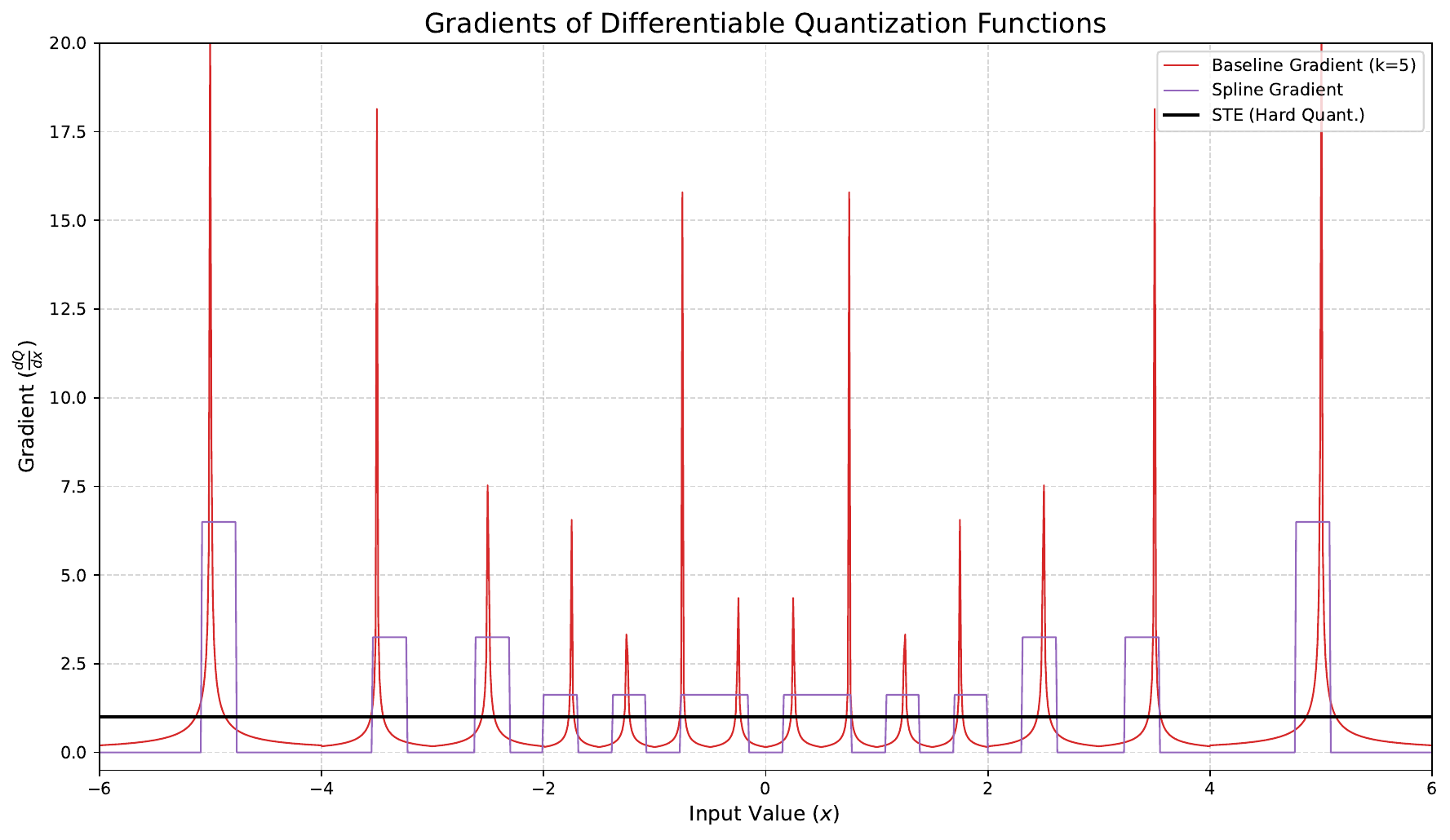}
        \caption{Gradient of $Q'_{\text{approx}}(x)$.}
        \label{subfig:quant_grad}
    \end{subfigure}
    \hfill 
    \begin{subfigure}[b]{0.32\linewidth}
        \centering
        \includegraphics[width=\linewidth]{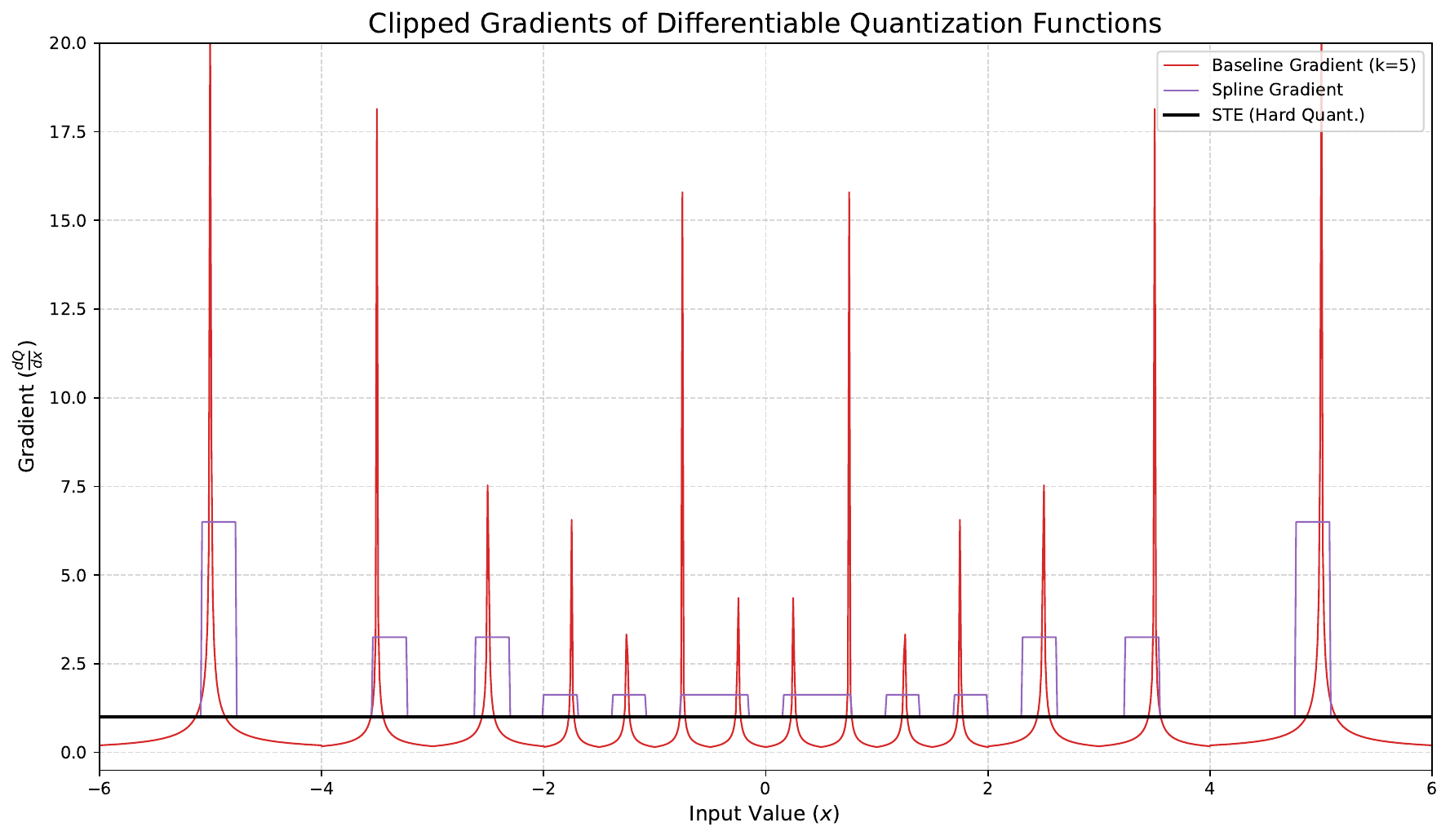}
        \caption{Clipped Gradient of $Q'_{\text{approx}}(x)$.}
        \label{subfig:quant_grad_clipped}
    \end{subfigure}
    \caption{Approximations of $Q(x)$ and their corresponding gradients, assuming ties-to-even rounding. We refer to \cite{wang2025optimizinglargelanguagemodel} as the baseline.  }
    \label{fig:quant_approx_and_grad}
    \vspace{-1em}
\end{figure}

We illustrate our proposed differentiable quantization approximation and its corresponding gradient in \cref{fig:quant_approx_and_grad}. The function is shown in \cref{subfig:quant_func}, and its gradient is depicted in \cref{subfig:quant_grad}. We found that applying the quantisation gradient in the backwards pass sometimes would mask out the gradient entirely, hence we propose clipping $Q'(x)$ from below to prevent multiplying the gradient with 0 (\Cref{subfig:quant_grad_clipped}). The overall complexity overhead of the spline approximation is thus $\mathcal{O}(nm\log_2(k))$.

 Note that we need to save the unquantised matrix $\mathbf{X}$ for the backwards pass to evaluate $Q'(\mathbf{X})$, adding $\mathcal{O}(mn)$ memory overhead.

\section{Gradient Adjustment for Scaling Factor Quantization}

The gradient adjustment techniques used for weights and activations can also be applied to the quantization of the scaling factor $q(s)$. However, the relatively high dynamic range required for scaling factors introduces additional complexity. To find a decent trade-off between accuracy and complexity, we first analyze the regions where the quantization error is most significant. We measure this error using the relative deviation, defined as the ratio $s/s_q$. A value of this ratio far from 1 indicates a large quantization error.

\begin{figure}[htbp!]
    \centering
    \begin{subfigure}[b]{0.49\linewidth}
        \centering
        \includegraphics[width=\linewidth]{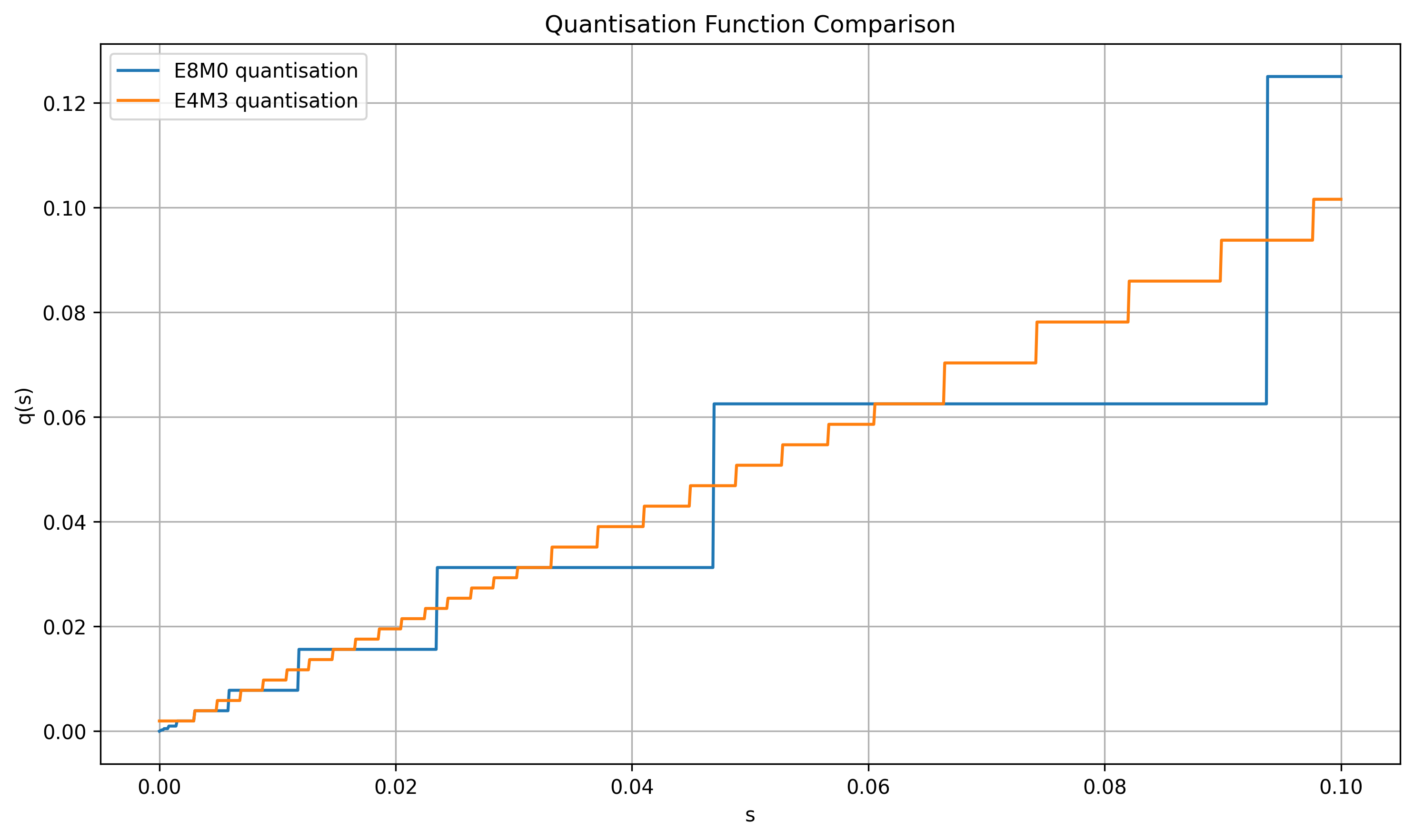}
        \caption{Quantization functions $s_q = q(s)$.}
        \label{subfig:sf_quant_func}
    \end{subfigure}
    \hfill
    \begin{subfigure}[b]{0.49\linewidth}
        \centering
        \includegraphics[width=\linewidth]{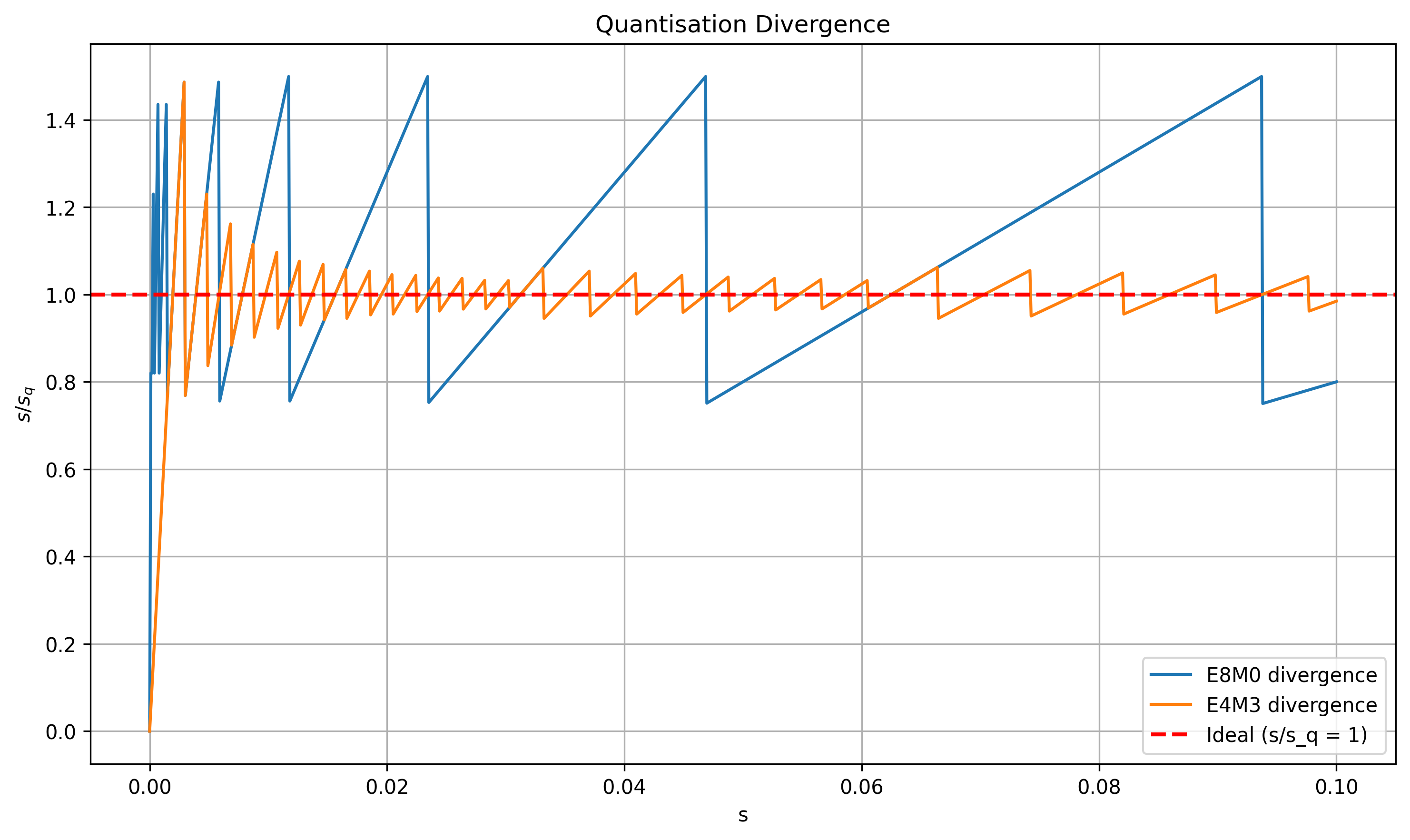}
        \caption{Relative deviation $s/s_q$.}
        \label{subfig:sf_quant_div}
    \end{subfigure}
    \caption{Comparison of quantization for \texttt{E4M3} and \texttt{E8M0} scaling factors. Figure (a) shows the quantization step functions. Figure (b) shows the relative deviation, which is most pronounced for small values of the scaling factor $s$.}
    \label{fig:scaling_factor_quant}
\end{figure}

\Cref{fig:scaling_factor_quant} illustrates the quantization functions for the \texttt{E4M3} and \texttt{E8M0} scale formats and their corresponding relative deviations. The quantization function itself is shown in \cref{subfig:sf_quant_func}, while the error is plotted in \cref{subfig:sf_quant_div}.

As illustrated in \cref{subfig:sf_quant_div}, the largest relative deviation occurs for small-magnitude scaling factors, especially within the first few representable values of the \texttt{E4M3} scale format. Based on this observation, we can choose to apply the gradient adjustment selectively, targeting only the range where the quantization error is highest when computing the $q'(s)$ term.


\subsection{Gradient adjustment of \texttt{absmax}: Adjusting for $Z(\mathbf{X}_p)$ and $s'(\mathbf{X}_p)$}

First, we establish the general relationship between the gradient of the scaling factor, $\frac{\partial s}{\partial \mathbf{X}}$, and the gradient of the normalization function, $\frac{\partial Z}{\partial \mathbf{X}}$.

\begin{proposition}
Given the scaling factor $s(\mathbf{X}) = \frac{\texttt{FP4 max}}{Z(\mathbf{X})}$, its element-wise gradient with respect to an element $\mathbf{X}_{ij}$ is given by:
\[
\frac{\partial s}{\partial \mathbf{X}_{ij}} = -\frac{\texttt{FP4 max}}{Z(\mathbf{X})^2} \frac{\partial Z}{\partial \mathbf{X}_{ij}}
\]
See \Cref{derivation_3} for derivations.
\end{proposition}

The following corollaries provide the specific form of this gradient for two common choices of the normalization function $Z(\mathbf{X})$.

\begin{corollary}
\label{col_1}
If the normalization function $Z(\mathbf{X})$ is \texttt{absmax}, $Z(\mathbf{X}) = \max_{k,l} |\mathbf{X}_{kl}|$, then the gradient of the scaling factor is non-zero only for the element with the maximum absolute value:
\begin{equation}
\frac{\partial s}{\partial \mathbf{X}_{ij}} = -\frac{\texttt{FP4 max}}{Z(\mathbf{X})^2} \left( \sign(\mathbf{X}_{i^*j^*}) \cdot \delta_{ii^*}\delta_{jj^*} \right)
\end{equation}
where $(i^*, j^*)$ is the index of the maximum absolute value element and $\delta$ is the Kronecker delta.
See \Cref{derivation_4} for derivations.
\end{corollary}

\begin{corollary}
If the normalization function $Z(\mathbf{X})$ is the smooth LogSumExp approximation of the max function, $Z(\mathbf{X}) = \frac{1}{\beta} \log\left(\sum_{k,l} e^{\beta|\mathbf{X}_{kl}|}\right)$, the gradient of the scaling factor is a dense gradient given by:
\begin{equation}
\frac{\partial s}{\partial \mathbf{X}_{ij}} = -\frac{\texttt{FP4 max}}{Z(\mathbf{X})^2} \left( \softmax(\beta|\mathbf{X}|)_{ij} \cdot \sign(\mathbf{X}_{ij}) \right)
\end{equation}
See \Cref{derivation_5} for derivations.
\end{corollary}

We consider four configurations for calculating the gradients with respect to the scaling factors, summarized in Table~\ref{tab:gradientTable}. Alongside the standard `Absmax` and `Softmax` approaches, we introduce a `Hybrid` method. This approach uses the computationally efficient `absmax` function in the forward pass but approximates its gradient with the dense `softmax` derivative during the backward pass. This is intended to propagate gradient information to more elements without incurring the forward-pass cost of the LogSumExp operation.

\begin{table}[htbp]
\centering
\caption{Gradient configurations for the block-wise scale $s(\mathbf{X})$ and global scale $g(\mathbf{X})$. The Straight-Through Estimator (STE) gradient is a heuristic approximation, as detailed in the text.}
\label{tab:gradientTable}
\renewcommand{\arraystretch}{1.8}
\resizebox{\textwidth}{!}{%
\begin{tabular}{>{\bfseries}l p{3cm} p{5.5cm} p{3cm} p{5.5cm}}
\toprule
\textbf{Configuration} & \textbf{Scaling Function} \( Z(\mathbf{X}) \) & \textbf{Gradient} \( \frac{\partial s}{\partial \mathbf{X}_{ij}} \) & \textbf{Global Scaling Function} \( g(\mathbf{X}) \) & \textbf{Gradient} \( \frac{\partial g}{\partial \mathbf{X}_{ij}} \) \\
\midrule
STE &
\( \max_{k,l} |\mathbf{X}_{kl}| \) &
\( 1 \) &
\( \max_{k,l} |\mathbf{X}_{kl}| \) &
\( 1 \) \\
\addlinespace
Absmax &
\( \max_{k,l} |\mathbf{X}_{kl}| \) &
\( -\frac{\text{FP4 max}}{(Z(\mathbf{X}))^2} \left( \operatorname{sign}(\mathbf{X}_{i^*j^*}) \cdot \delta_{ii^*} \delta_{jj^*} \right) \) &
\( \max_{k,l} |\mathbf{X}_{kl}| \) &
\( \operatorname{sign}(\mathbf{X}_{i^*j^*}) \cdot \delta_{ii^*} \delta_{jj^*} \) \\
\addlinespace
Softmax &
\( \frac{1}{\beta} \log\left( \sum_{k,l} e^{\beta|\mathbf{X}_{kl}|} \right) \) &
\( -\frac{\text{FP4 max}}{(Z(\mathbf{X}))^2} \left( \operatorname{softmax}(\beta|\mathbf{X}|)_{ij} \cdot \operatorname{sign}(\mathbf{X}_{ij}) \right) \) &
\( \frac{1}{\beta} \log\left( \sum_{k,l} e^{\beta|\mathbf{X}_{kl}|} \right) \) &
\( \operatorname{softmax}(\beta|\mathbf{X}|)_{ij} \cdot \operatorname{sign}(\mathbf{X}_{ij}) \) \\
\addlinespace
Hybrid &
\( \max_{k,l} |\mathbf{X}_{kl}| \) &
\( -\frac{\text{FP4 max}}{(Z(\mathbf{X}))^2} \left( \operatorname{softmax}(\beta|\mathbf{X}|)_{ij} \cdot \operatorname{sign}(\mathbf{X}_{ij}) \right) \) &
\( \max_{k,l} |\mathbf{X}_{kl}| \) &
\( \operatorname{softmax}(\beta|\mathbf{X}|)_{ij} \cdot \operatorname{sign}(\mathbf{X}_{ij}) \) \\
\bottomrule
\end{tabular}%
}
\end{table}

For any softmax-based configuration, we must either compute or save the softmax for the backwards pass, incurring additional time and memory complexity. For \emph{Absmax}, it suffices to save the index of the maximum value. For the STE case of \( \frac{\partial s}{\partial \mathbf{X}_{ij}} \), we are effectively setting the entire second term from Equation (1), \( \left[ \frac{q'(s)}{s_q} \left( \mathbf{X}_{ij} Q'(s_q \mathbf{X}_{ij}) - \frac{1}{s_q} Q(s_q \mathbf{X}_{ij}) \right) \right] \), to be equal to 1. This provides a simple alternative to completely omit the $\mathcal{O}(3mn)$ extra computation of this extra gradient term, treating the complex scaling derivative as a direct pass-through. We have an \texttt{ignore} option for the \( \frac{\partial g}{\partial \mathbf{X}_{ij}} \) term, which means setting the corresponding update term from \Cref{coro_1} to zero: $\frac{\partial g}{\partial \mathbf{X}_{ij}} \left( f_{ij}(\mathbf{U}_p) - \mathbf{U}_{p,ij} \frac{\partial f_{ij}}{\partial \mathbf{U}_{p,ij}} \right) = 0$, skipping the extra $\mathcal{O}(4mn)$ work and saving memory.

\section{Other techniques}

\textbf{Optimizer centric} Recently, \cite{huang2025stablespamtrain4bitstably} proposed StableSPAM, which modifies the Adam optimiser by bounding the momentum term with a moving average statistic. This is motivated by the observation that in low precision, the gradient norms tend to explode during training, meaning more careful normalisation of the momentum norm and bounding of large values is needed to stabilise training. While their optimiser is primarily tailored around LLMs, we explore the impact of combining StableSPAM with existing rounding and gradient adjustment based techniques for general purpose ML workloads.

\textbf{Loss scaling} We consider loss scaling as a technique to propagate signal when the range of the precision is very limited following \cite{micikevicius2018mixedprecisiontraining}. We implement the automated loss scaling technique, which adjusts the loss scaling scale dynamically during training. 

\textbf{Outlier concentration} During quantisation, outliers in high-precision may induce quantisation error as they impact the scaling during quantisation. Recent work by \cite{tseng2024quipbetterllmquantization,tseng2025trainingllmsmxfp4} proposes applying $Q(\mathbf{HS}\mathbf{X}_p)$ to concentrate outliers towards the median of the data. Here $\mathbf{HS}$ is the random Hadamard transform applied to each block $\mathbf{X}_p$ of size $l$ elements, inducing a $\mathcal{O}(\frac{mn}{l} l\cdot\log(l))$ compute overhead. It should be noted that this operation is fusable, and can be done on-the-fly with warp shuffle operations. We consider applying Hadamard transformation in both the forward and backward pass and only the backward pass, akin to \cite{tseng2024quipbetterllmquantization,tseng2025trainingllmsmxfp4,castro2025quartetnativefp4training}.

\textbf{Spectral decomposition} In \cite{li2025svdquantabsorbingoutlierslowrank,cao2025metistraininglargelanguage}, they propose to use spectral decomposition techniques to alleviate the difficulty of quantising outliers in low-precision. This is done by decomposing the tensor into a low-rank representation using \emph{singular value decomposition} (SVD), where the low-rank components are then quantised instead. As this is non-fusable, and has prohibitive time complexity overhead $\mathcal{O}(mnk)$ (with $k$ referring to a chosen lower rank), we do not consider it for our simulations as the Hadamard transformation offers a more seamless alternative in the pre-training setting.

\begin{table}[h!]
\centering
\caption{Summary of FP4 Training Techniques and Overheads. Here we assume each operation is applied to a tensor with $n$ elements, which can be partitioned to $n//l$ blocks with block size $l$.}
\label{tab:techniques_summary}
\renewcommand{\arraystretch}{1} 
\small 
\resizebox{\textwidth}{!}{%

\begin{tabular}{>{\bfseries}l p{2cm} p{1.8cm} c p{2.5cm}}
\toprule
\textbf{Technique} & \textbf{Compute Overhead} & \textbf{Additional Memory} & \textbf{Fuseability} & \textbf{Comment} \\
\midrule
Straight-Through Estimator (STE) & \(\mathcal{O}(1)\) & None & Yes &  \\
Baseline $Q'(\mathbf{X}))$\cite{wang2025optimizinglargelanguagemodel} & $\mathcal{O}(n\cdot w \log{k} )$ &  $\mathcal{O}(n)$ & No & $w=5$\\
Spline $Q'(\mathbf{X}))$ & \(\mathcal{O}(n\log k)\) & $\mathcal{O}(n)$ & No &  \\
Stochastic Rounding \cite{fitzgibbon2025stochasticroundingrandombits} & \(\mathcal{O}(n)\) & None & Yes & \\
Stochastic Rounding Scale & \(\mathcal{O}(n//l)\) & None & Yes & \\
Global Tensor Scaling \cite{blake2023unitscalingoutoftheboxlowprecision} & \(\mathcal{O}(n)\) & $\mathcal{O}(1)$ & Yes & Rescale in full prec.  \\
Global Scaling Gradient (\Cref{coro_1}) & \(\mathcal{O}(3n)\) & \(\mathcal{O}(3n)\) & No &Save ex. tensor \\
Differentiable Scale (Absmax) & \(\mathcal{O}(4n)\) & \(\mathcal{O}(3n)\) & No &   \\
Differentiable Scale (Softmax) & \(\mathcal{O}(4n)\) & \(\mathcal{O}(4n)\) & No & Softmax backw. \\
Scale Gradient Adjustment  & \(\mathcal{O}(n)\) &  \(\mathcal{O}(n//l)\) & No & Only for Diff. Scale \\
Outlier concentration (Hadamard) \cite{tseng2025trainingllmsmxfp4} & \(\mathcal{O}(n \cdot  \log l)\) & None & Yes & On-the-fly possible \\
StableSPAM Optimizer \cite{huang2025stablespamtrain4bitstably} & $\mathcal{O}(n)$ & \(\mathcal{O}(1)\) & No & \\
Dynamic Loss Scaling \cite{micikevicius2018mixedprecisiontraining} & \(\mathcal{O}(n)\) & \(\mathcal{O}(1)\) & No &Mult. each tensor \\
SVD techniques \cite{li2025svdquantabsorbingoutlierslowrank,cao2025metistraininglargelanguage} & \(\mathcal{O}(nk)\) & \(\mathcal{O}(k^2)\) & No & \\ 
\bottomrule
\end{tabular}
}
\end{table}

\section{Experiments}

\textbf{Experimental design and selection strategy} We consider the search space in Appendix \cref{tab:full_sweep_compact}, which totals to over 20,000 different parameter combinations, an infeasible search space for larger models. Consequently, our strategy is to do larger sweeps for smaller models that are faster to train and using the results to derive insights and prune the search space for larger models. We run the experiments in the order described in Appendix \cref{tab:exp_summary}.

\textbf{Performance--efficiency score}  We define an efficiency score $S(c)=\frac{G(c)}{1 + \Omega(c)}$, for a configuration $c$, that balances relative performance gain $G(c) = (M_\text{ref} - M_c)/M_\text{ref}$ against a complexity penalty $\Omega(c) = \sum_{t \in \mathcal{T}_c} w_t$. Here, $\mathcal{T}_c$ is the set of non-standard techniques used, $w_t$ their overhead points, and the $+1$ ensures a well-defined score for baseline configurations. Scores are split by positive/negative gain per format, guiding pruning toward configurations that maximize performance with minimal added complexity (see \Cref{par:appendix_complexity} for more details). We consider validation loss for $M$ when we calculate the score.

\subsection{Results}
\begin{figure}[htp]
\centering
\caption{Training and validation performance curves for selected models and datasets.}
\label{fig:key_models}

\begin{subfigure}[b]{0.33\textwidth}
    \centering
    \includegraphics[width=\textwidth]{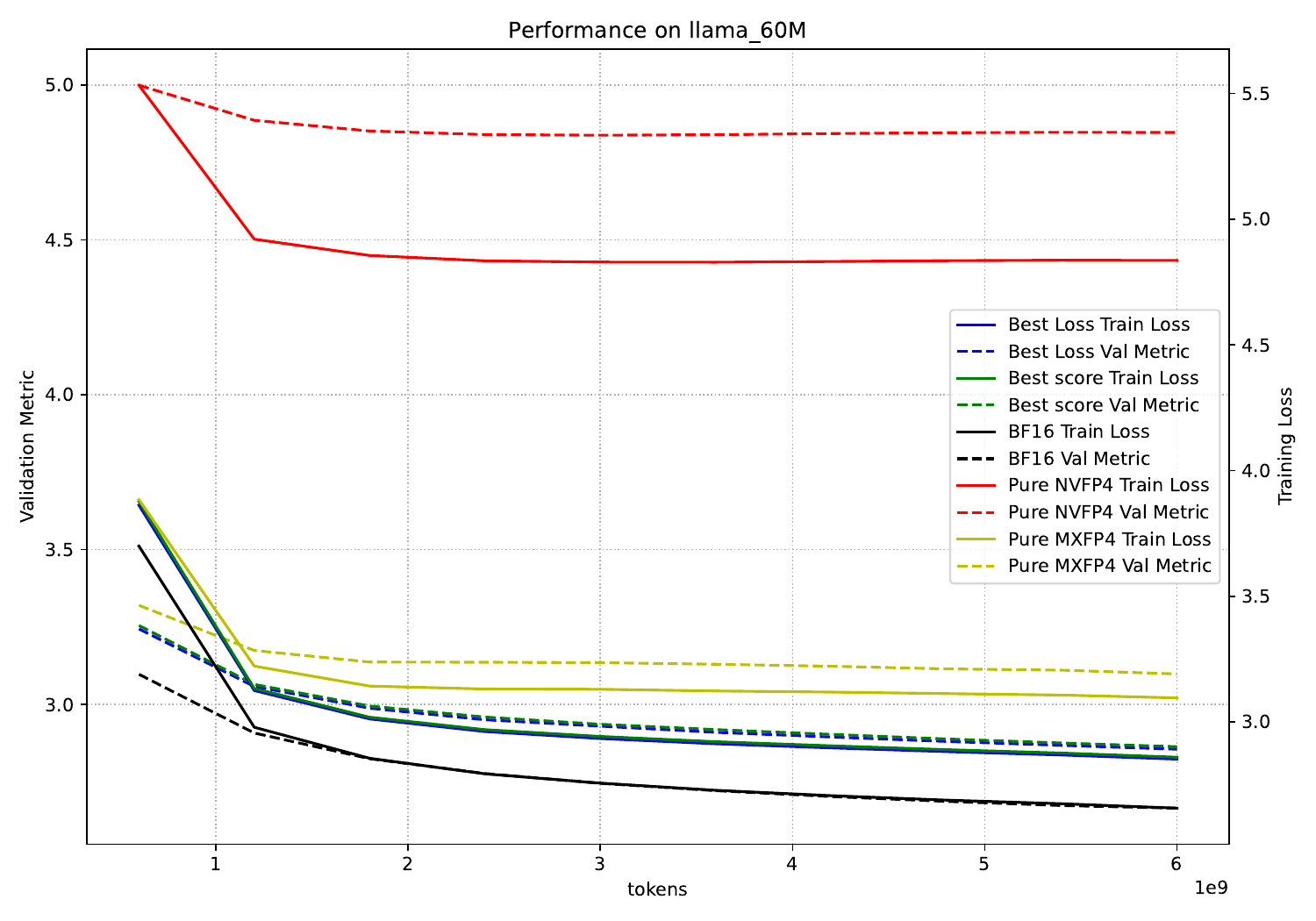}
    \caption{Llama 60M}
    \label{fig:llama_60m}
\end{subfigure}%
\begin{subfigure}[b]{0.33\textwidth}
    \centering
    \includegraphics[width=\textwidth]{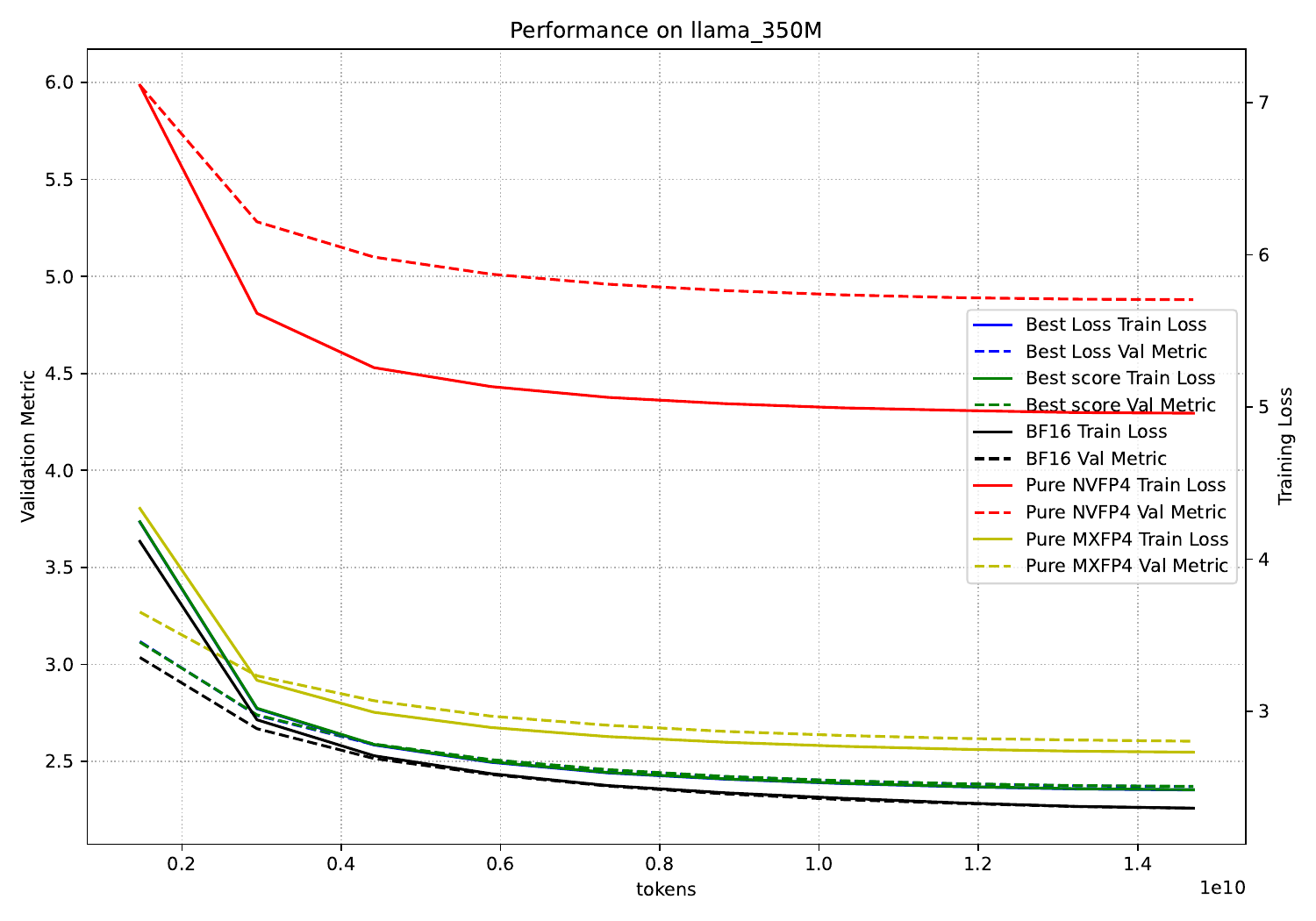}
    \caption{Llama 350M}
    \label{fig:llama_350m}
\end{subfigure}%
\begin{subfigure}[b]{0.33\textwidth}
    \centering
    \includegraphics[width=\textwidth]{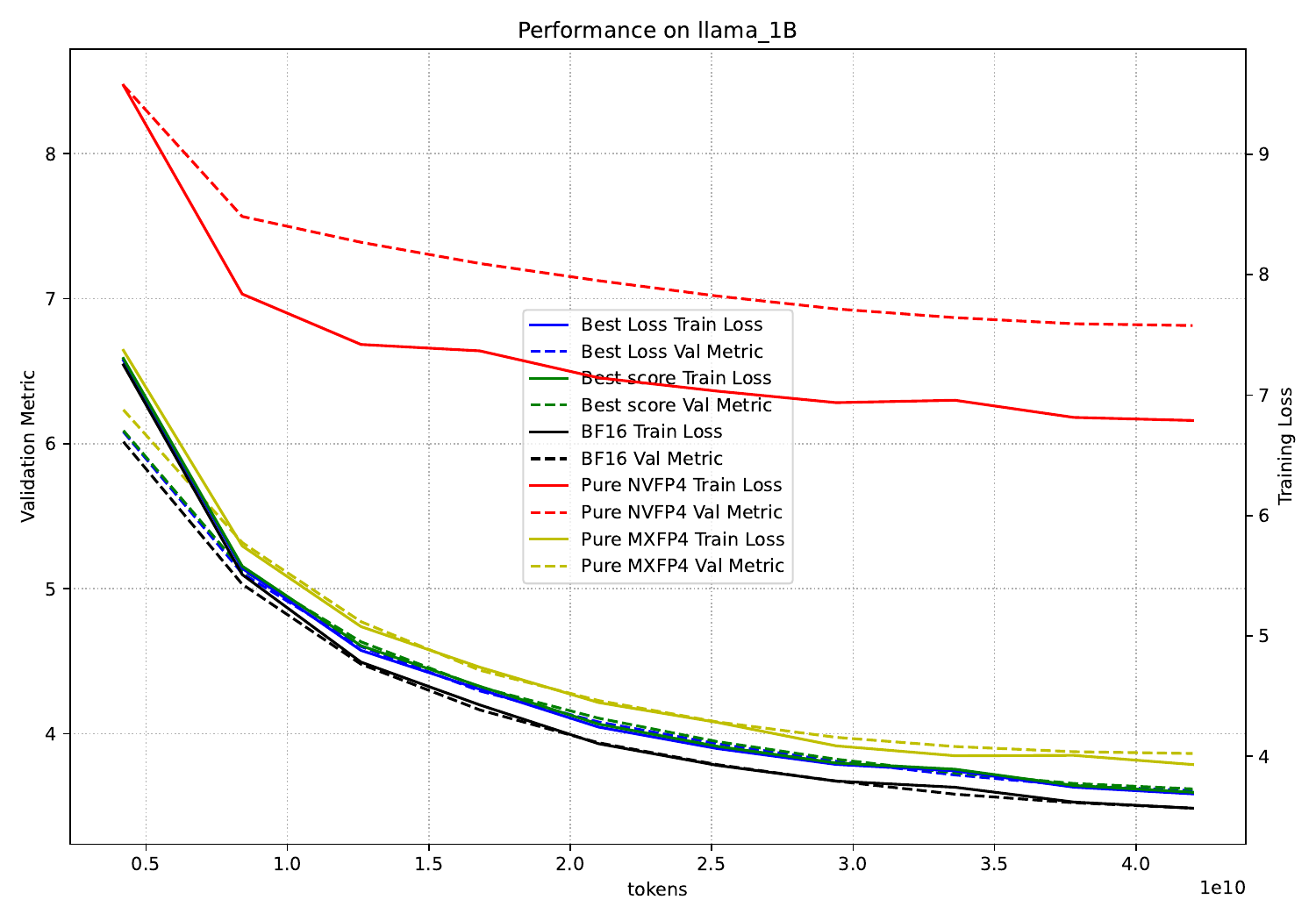}
    \caption{Llama 1B}
    \label{fig:llama_1b}
\end{subfigure}

\begin{subfigure}[b]{0.33\textwidth}
    \centering
    \includegraphics[width=\textwidth]{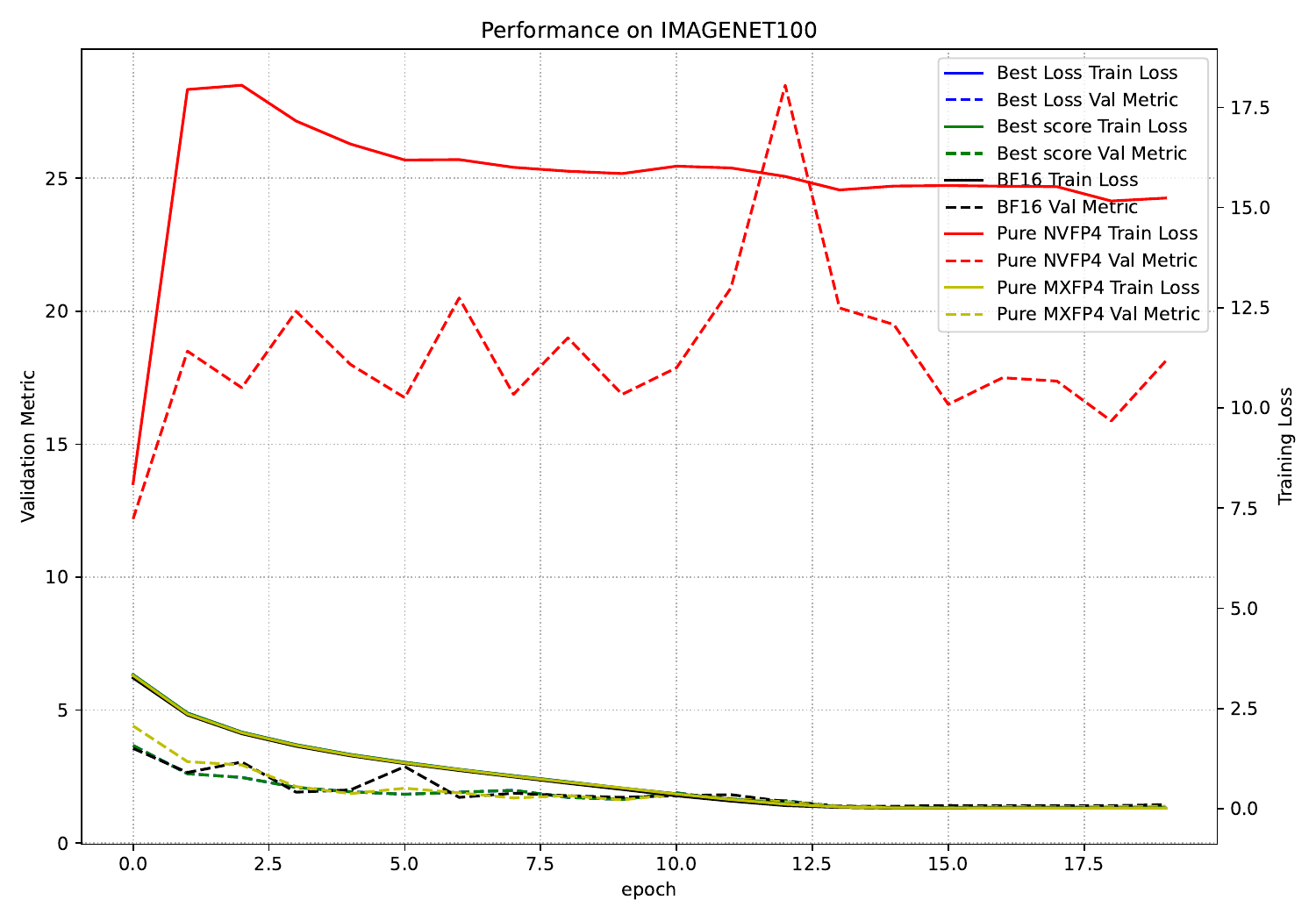}
    \caption{ImageNet-100}
    \label{fig:imagenet100}
\end{subfigure}%
\begin{subfigure}[b]{0.33\textwidth}
    \centering
    \includegraphics[width=\textwidth]{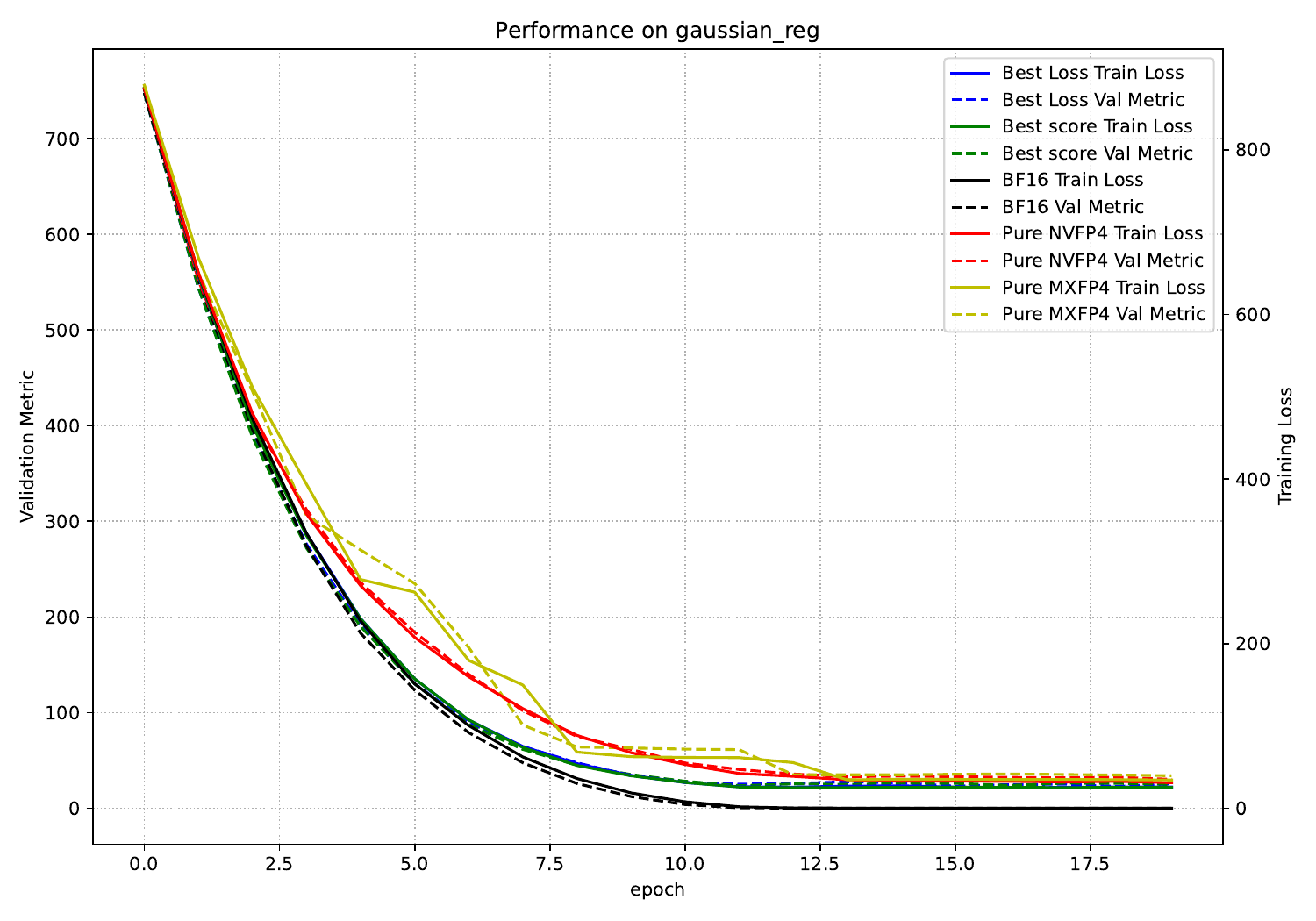}
    \caption{Gaussian Reg.}
    \label{fig:gaussian_reg}
\end{subfigure}%
\begin{subfigure}[b]{0.33\textwidth}
    \centering
    \includegraphics[width=\textwidth]{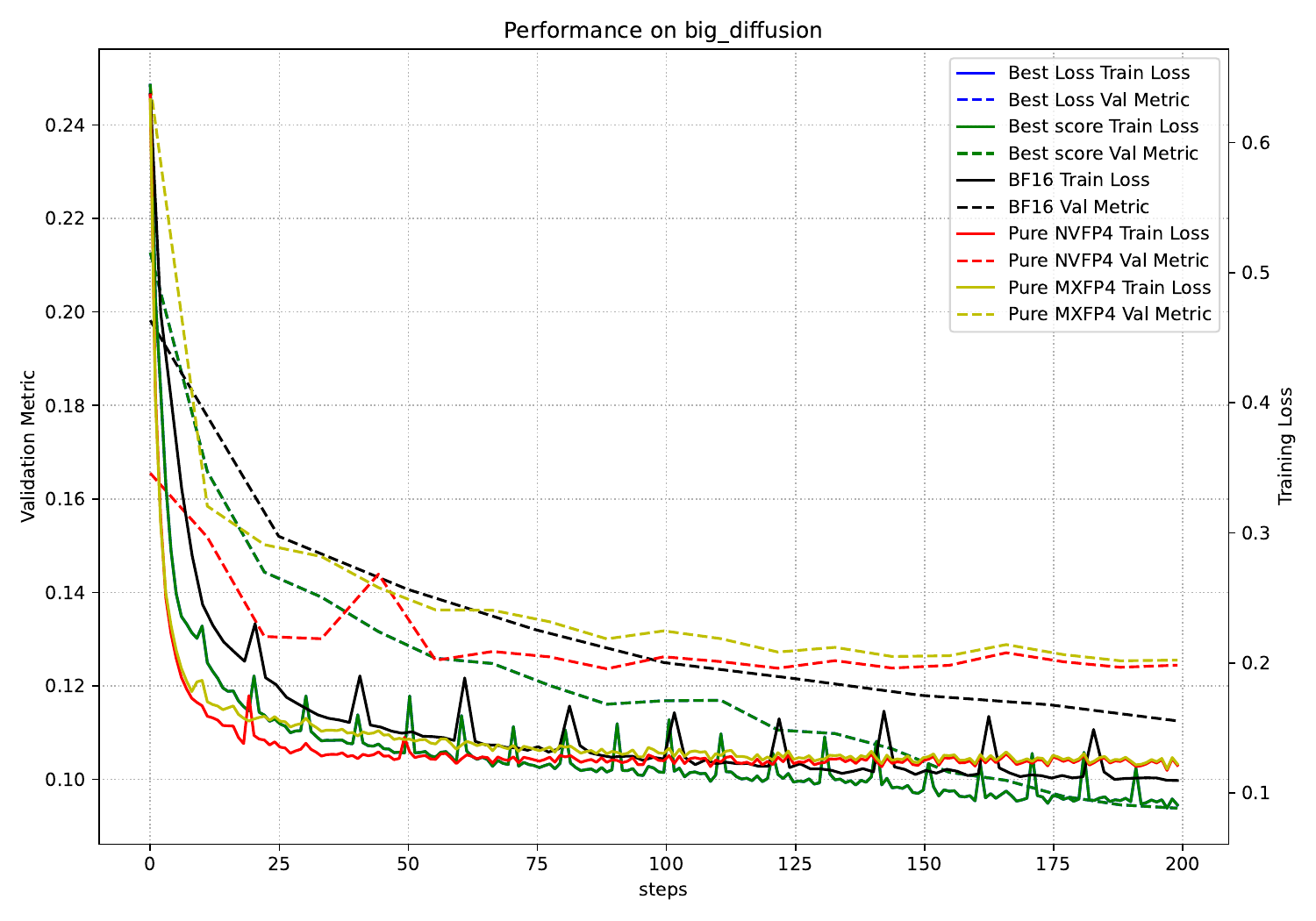}
    \caption{Big Diffusion}
    \label{fig:big_diffusion}
\end{subfigure}
\end{figure}

We loss curves of ImageNet-100, Gaussian regression, U-net large (big\_diffusion) and Llama 60M, 350M and 1B in \Cref{fig:key_models}. For detailed results of each dataset we refer to \Cref{additional_experiments,ue5m3_results}. Based on our learnings from experiments, we present three guiding principles when training in FP4.

\textbf{Principle 1: Gradient Stability Outweighs Unbiasedness} Across all our experiments (\Cref{additional_experiments}) we found that none of the proposed gradient adjustment had any significant positive effect on training stability compared to STE and consequently we were unable to match the findings in \cite{wang2025optimizinglargelanguagemodel}. As a possible explanation for this, consider the \texttt{absmax} gradient (\Cref{col_1}), which is a single non-zero entry per block. This mathematically enforces a sparse, high-variance update signal that may introduce high-variance, impacting momentum based optimisers such as Adam or StableSPAM. We also observed this when experimenting with $Q'(\mathbf{X})$, that adding this gradient without lower and upper bound clipping of the relaxation (see \Cref{subfig:quant_grad_clipped}) ended up masking out the downstream signal. Consequently using STE, which offers a dense and stable update and led to more stable training in the low-precision context. 

\textbf{Principle 2: Scale Representation is the Primary Bottleneck} We find throughout our experiments and ablations studies that the range of the scaling factor has a profound effect on training stability, especially demonstrated in larger language models and the ImageNet-100 runs in Appendix \Cref{tab:regression_results,tab:llm_results}. As many of our results contradicted findings in \cite{chmiel2025fp4wayfullyquantized}, we ran ablation studies in \Cref{sswiglu} investigating the additional impact of \texttt{SmoothSwiGLU}\footnote{This operation adds an approximate $\sim\mathcal{O}(n)$ non-fusable overhead due to additional \texttt{absmax} normalisation.} \citep{fishman2025scalingfp8trainingtrilliontoken} on language models and ablating the range of \texttt{E4M3} scaling, by replacing it with \texttt{E8M3} in \Cref{E8M3_results}. Our findings suggest that \texttt{E4M3}, despite applying tensor scaling, did not converge due to its range limitation. We speculate that a potential sweet spot exists between \texttt{E8M0} and \texttt{E4M3}. We then experiment with \texttt{UE5M3} in \Cref{ue5m3_results}, a format that has increased range and additional precision and find that it indeed consistently outperforms \texttt{E8M0} on language modelling. The caveat however is that it requires tensor scaling and SR in the backwards pass to achieve this performance. We further find throughout our experiments that nan-handling, also depends on scale format and that it although doesn't have a big impact, it matters.

\begin{figure}[htp!]
\centering
\caption{Pareto-frontier plots for each dataset, $\Omega(c)$ on the x-axis and $S(c)$ on the y-axis. $S(c)=0$ implies the configuration $c$ matches \texttt{BFLOAT16} performance.}
\label{fig:main_pareto}

\begin{subfigure}[b]{0.33\textwidth}
    \centering
    \includegraphics[width=\textwidth]{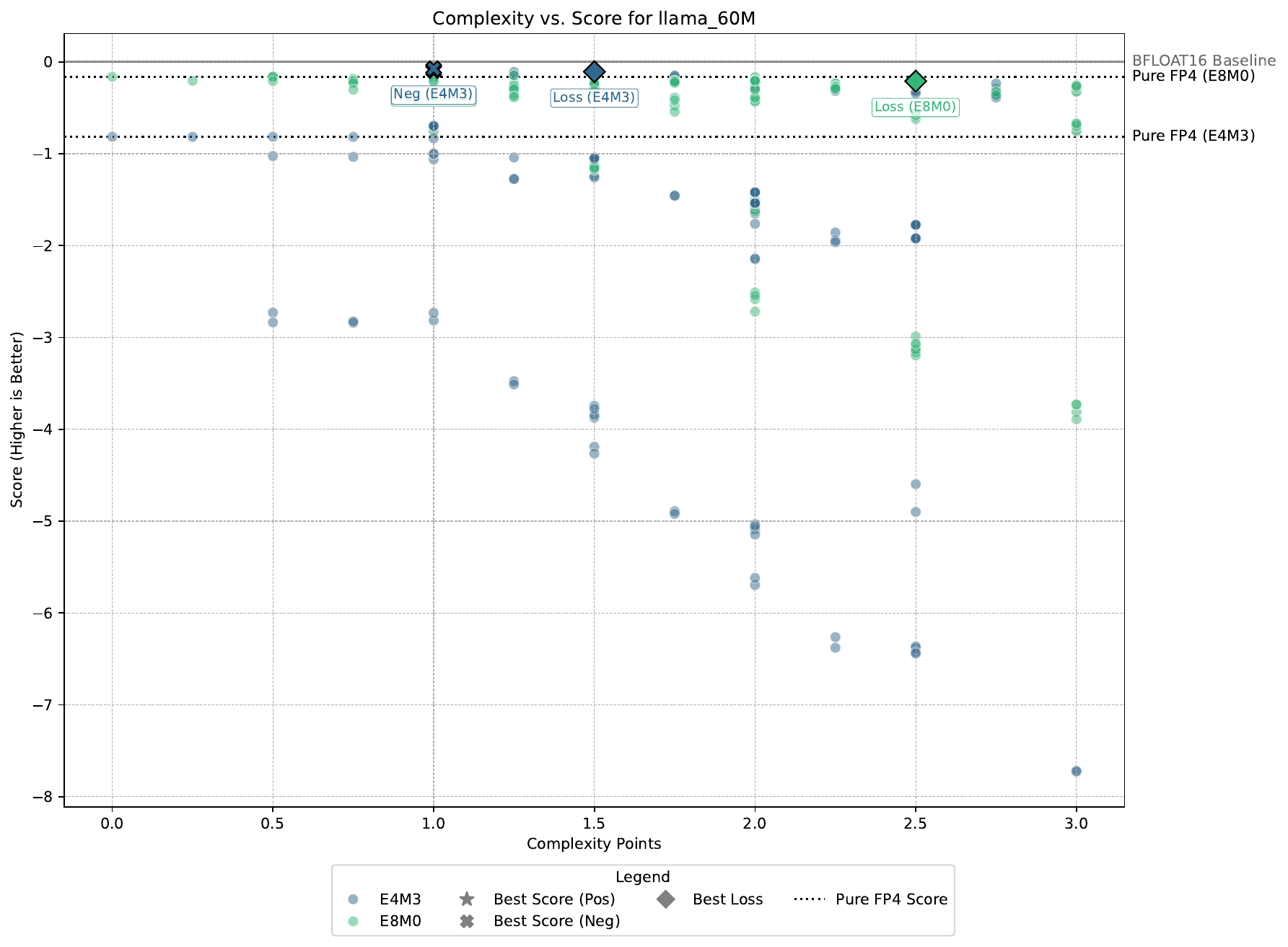}
    \caption{Llama 60M}
    \label{fig:llama_60m}
\end{subfigure}%
\begin{subfigure}[b]{0.33\textwidth}
    \centering
    \includegraphics[width=\textwidth]{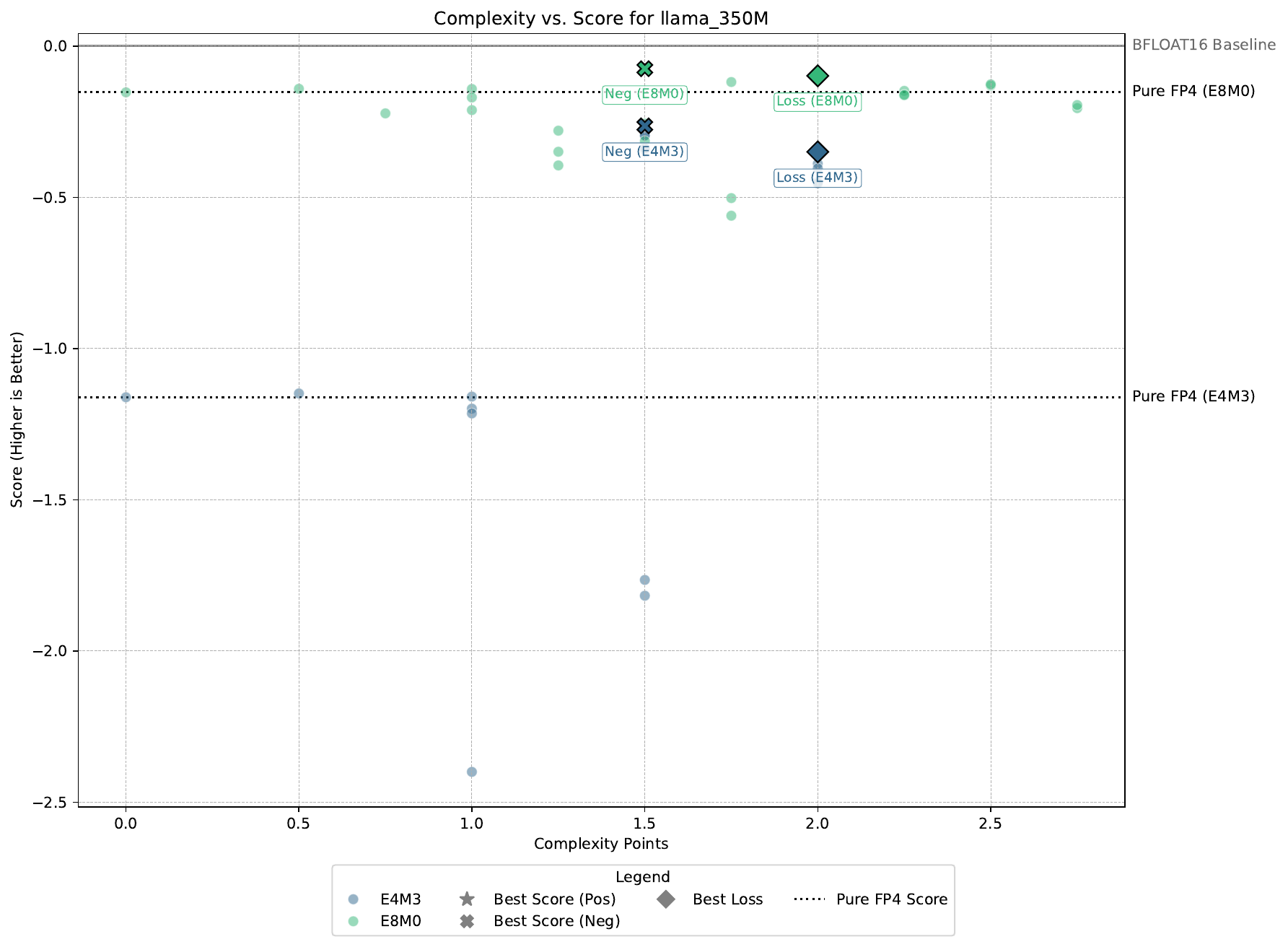}
    \caption{Llama 350M}
    \label{fig:llama_350m}
\end{subfigure}%
\begin{subfigure}[b]{0.33\textwidth}
    \centering
    \includegraphics[width=\textwidth]{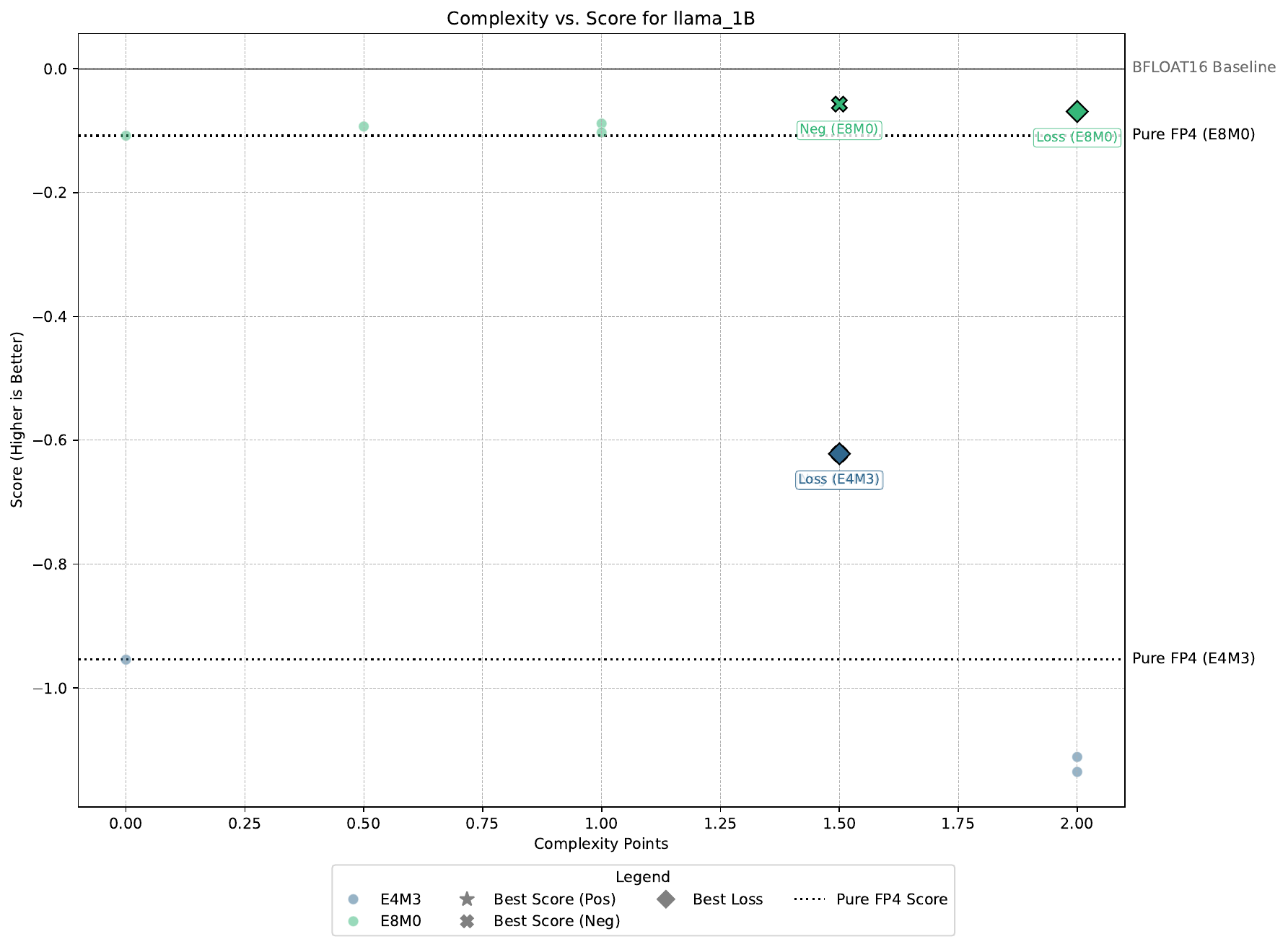}
    \caption{Llama 1B}
    \label{fig:llama_1b}
\end{subfigure}

\begin{subfigure}[b]{0.33\textwidth}
    \centering
    \includegraphics[width=\textwidth]{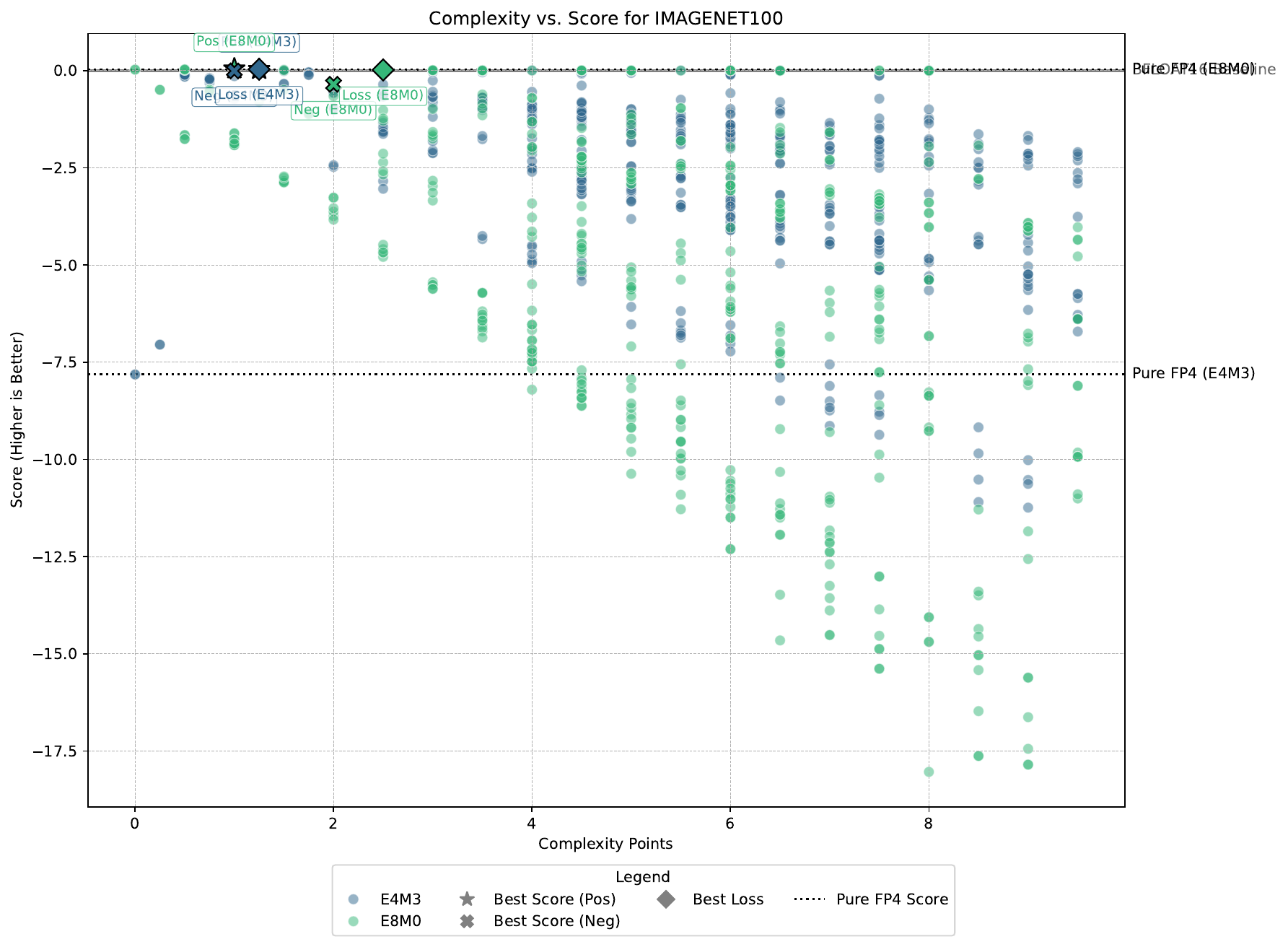}
    \caption{ImageNet-100}
    \label{fig:imagenet100}
\end{subfigure}%
\begin{subfigure}[b]{0.33\textwidth}
    \centering
    \includegraphics[width=\textwidth]{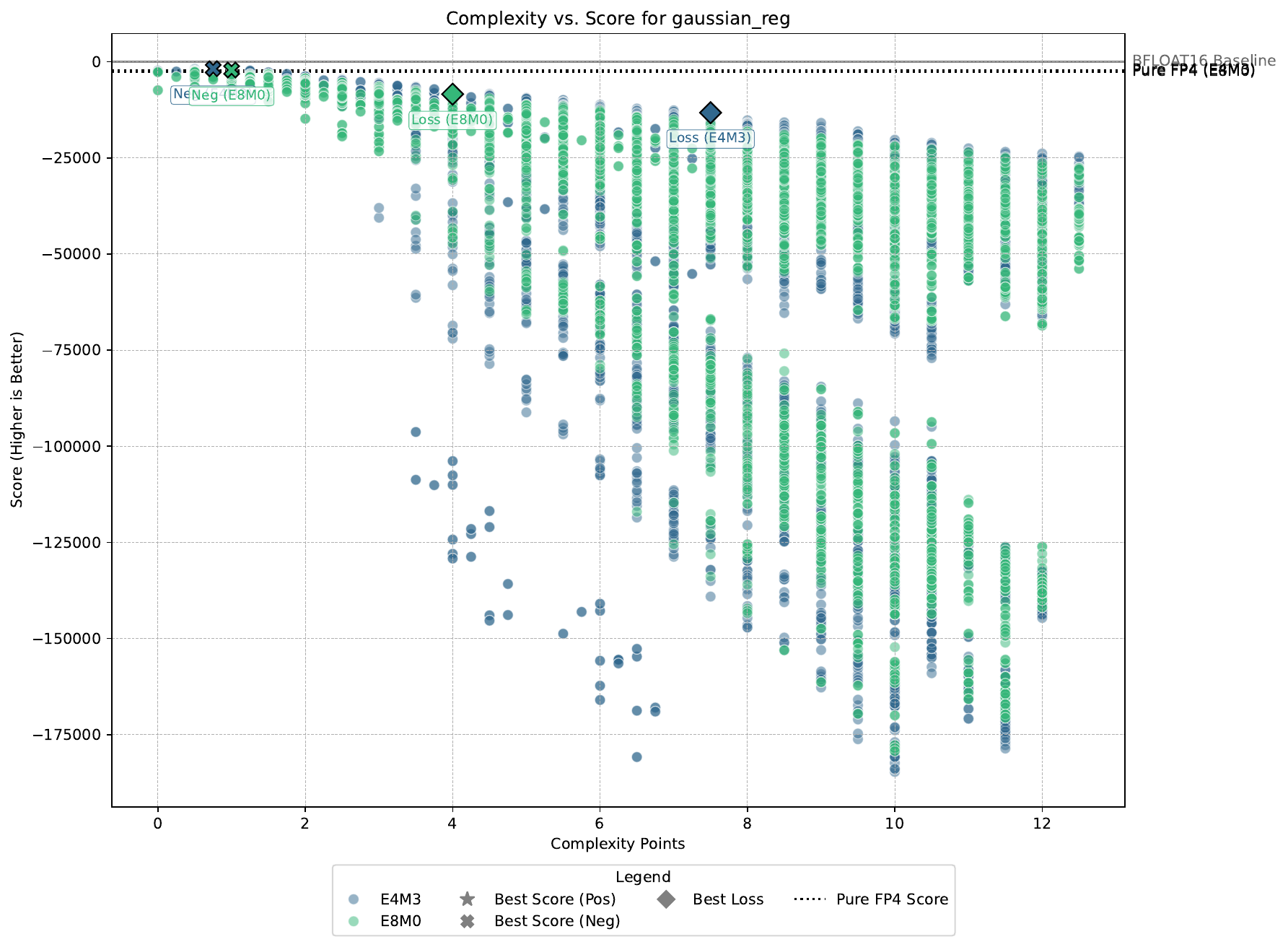}
    \caption{Gaussian Reg.}
    \label{fig:gaussian_reg}
\end{subfigure}%
\begin{subfigure}[b]{0.33\textwidth}
    \centering
    \includegraphics[width=\textwidth]{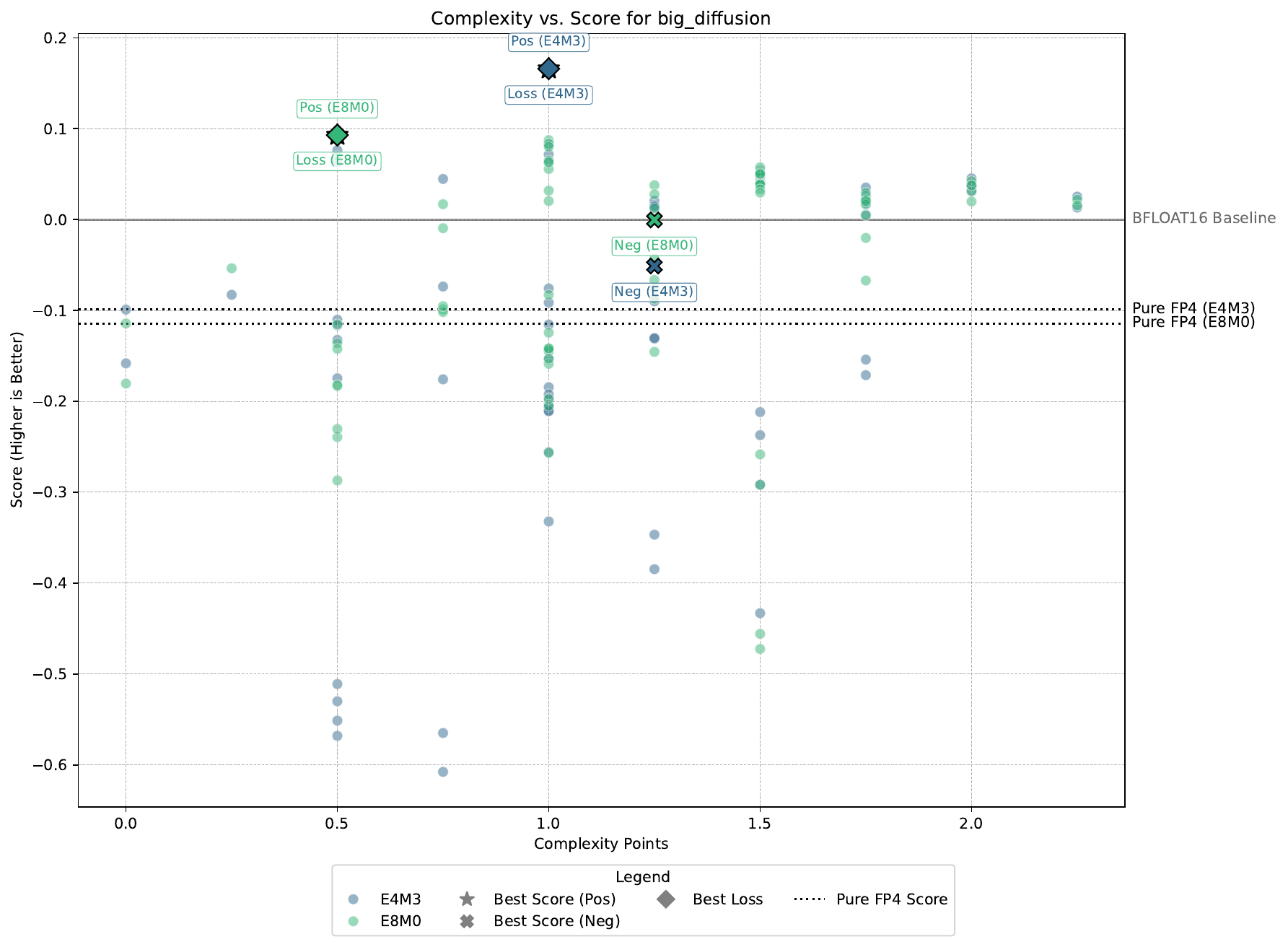}
    \caption{Big Diffusion}
    \label{fig:big_diffusion}
\end{subfigure}

\begin{subfigure}[b]{0.25\textwidth}
    \centering
    \includegraphics[width=\textwidth]{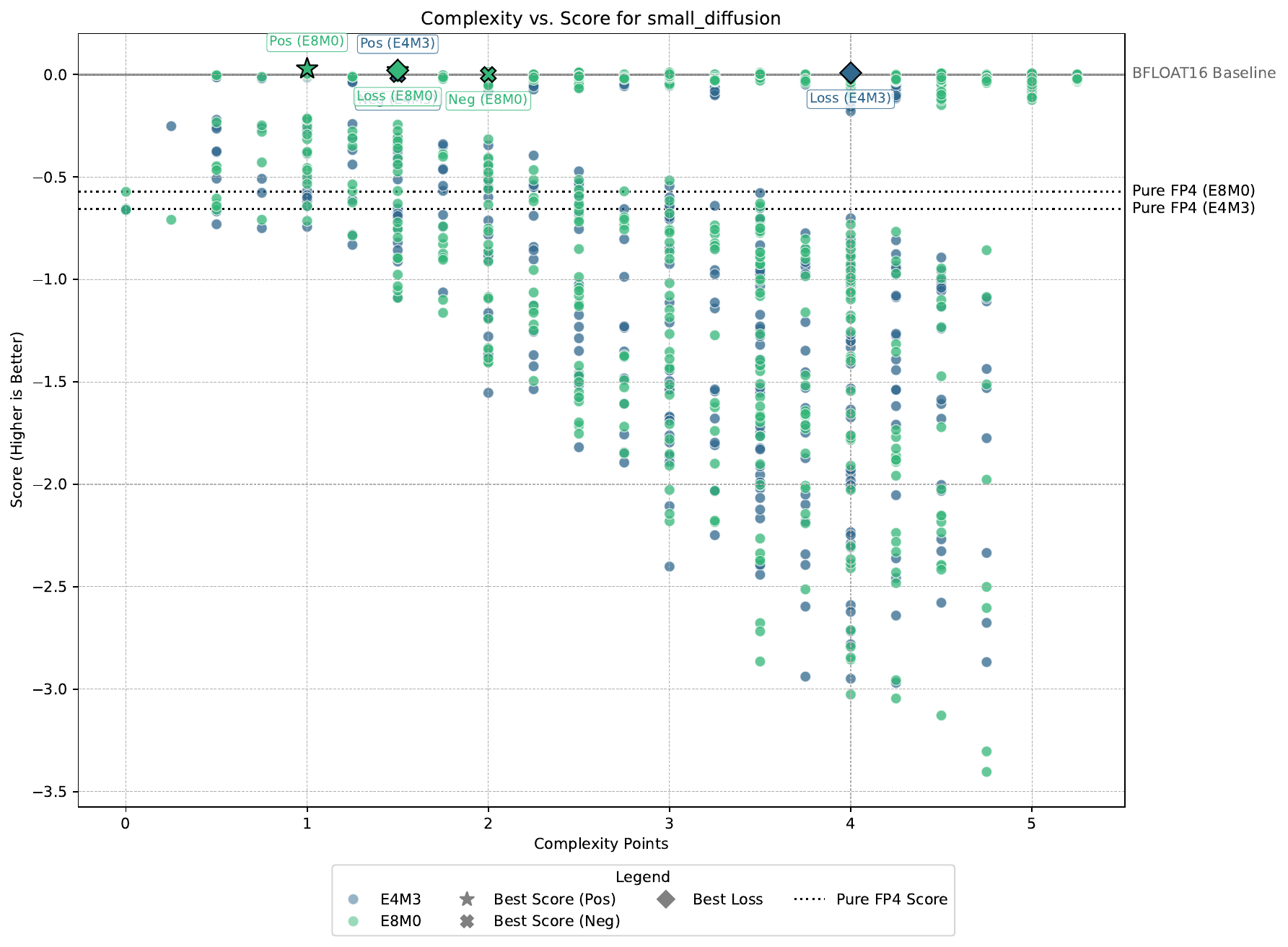}
    \caption{Small Diffusion}
    \label{fig:small_diffusion}
\end{subfigure}\hfill
\begin{subfigure}[b]{0.25\textwidth}
    \centering
    \includegraphics[width=\textwidth]{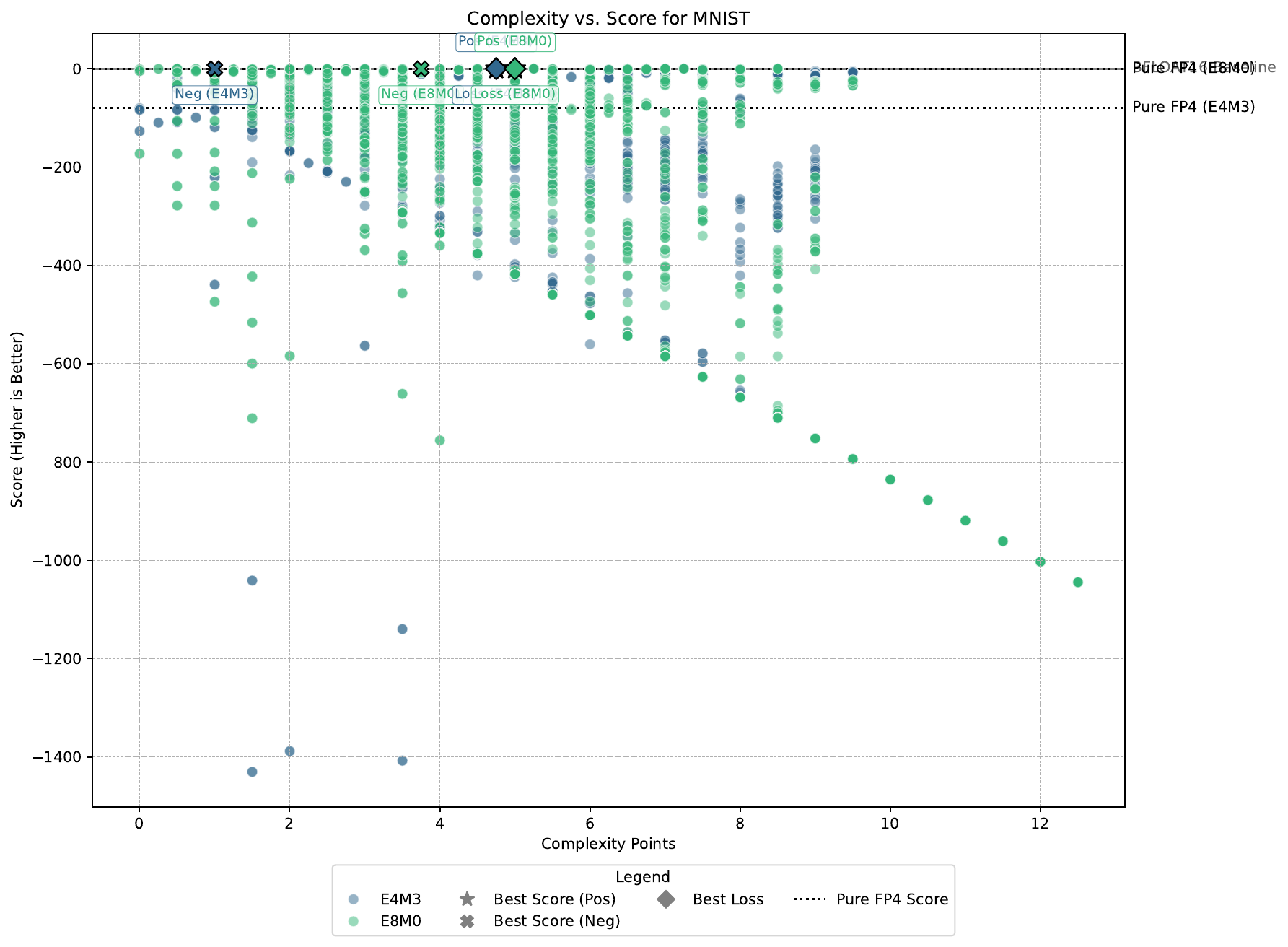}
    \caption{MNIST}
    \label{fig:mnist}
\end{subfigure}\hfill
\begin{subfigure}[b]{0.25\textwidth}
    \centering
    \includegraphics[width=\textwidth]{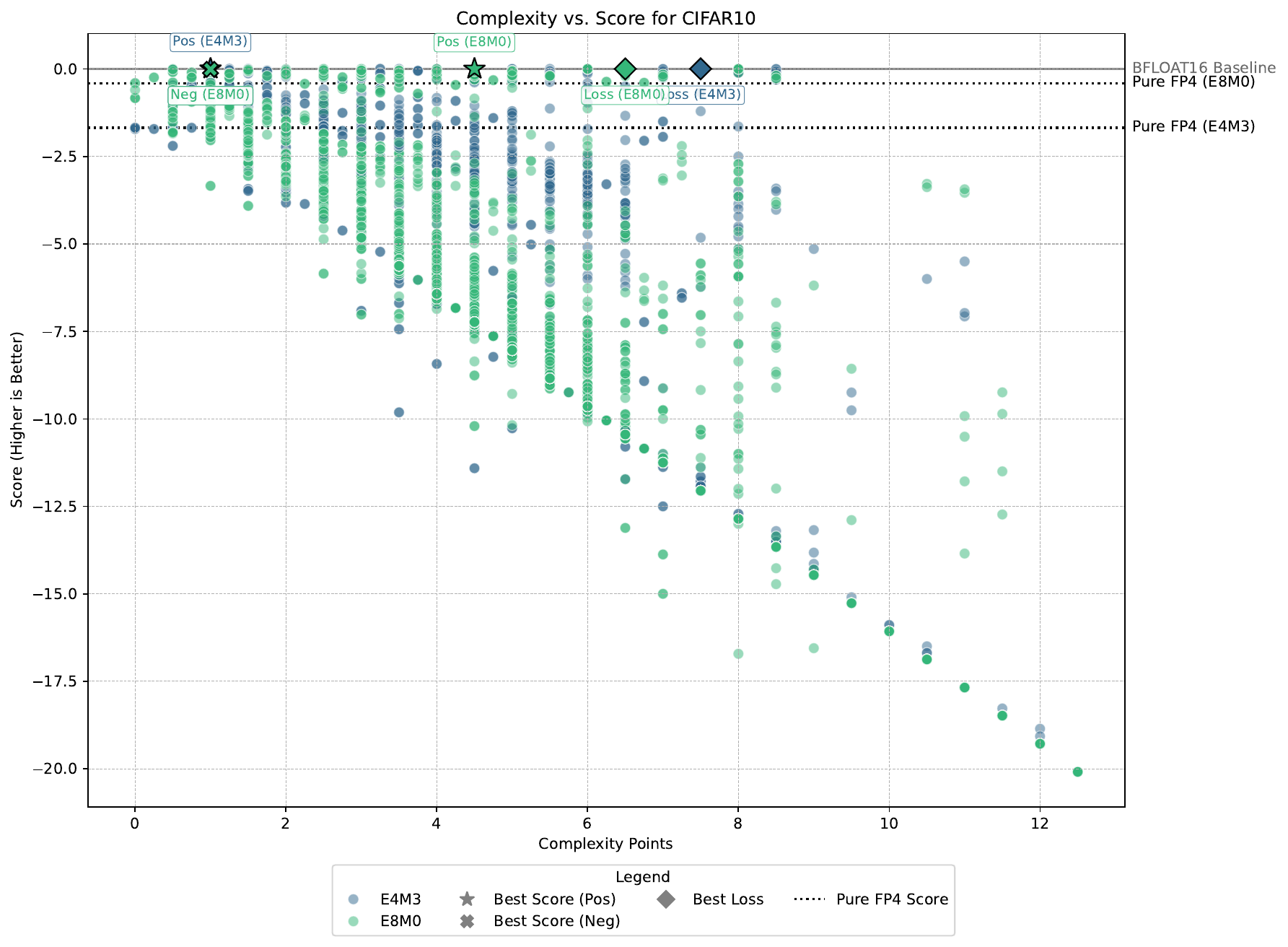}
    \caption{CIFAR-10}
    \label{fig:cifar10}
\end{subfigure}\hfill
\begin{subfigure}[b]{0.25\textwidth}
    \centering
    \includegraphics[width=\textwidth]{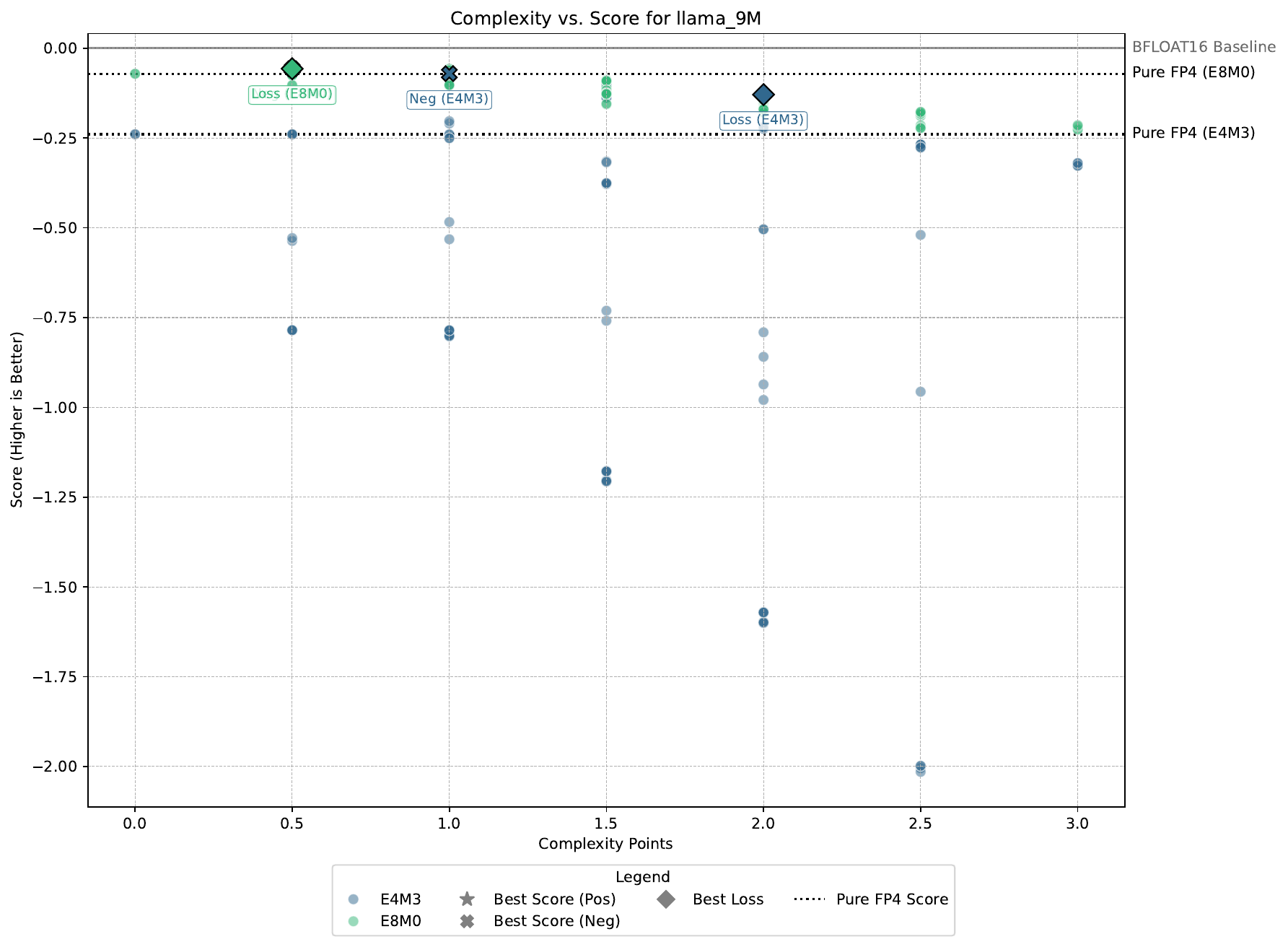}
    \caption{Llama 9M}
    \label{fig:llama_9m}
\end{subfigure}

\end{figure}

\textbf{Principle 3: The Performance-Overhead Frontier is Sparse}  From our sweeps over thousands of configurations, we find that only a handful of techniques such as Hadamard transforms, tensor scaling, stochastic rounding and optimiser choice provide a consistent, positive return on their computational overhead. We illustrate this in Pareto-frontier plots in \Cref{fig:main_pareto} for each dataset and Appendix \Cref{fig:ue5m3_pareto} for the \texttt{UE5M3} experiments. We overall observe that less complex configurations achieve better scores, and adding complexity yields diminishing returns. One can achieve lower loss, but often at a steep cost with respect to increased overhead, as observed in classification tasks.

\textbf{Conclusion and further work} We've proposed a novel framework for deriving the exact gradient updates for a linear layer under micro-scaling quantisation. While differentiable \texttt{absmax} gradients and quantisation gradients provided a benefit on smaller classification tasks, we found they offered no improvement or were detrimental for larger diffusion and language models, suggesting they can often be omitted to reduce overhead without sacrificing performance in these domains. Stochastic rounding of the scale showed little success beyond small models. We further find that the range of \texttt{NVFP4} hampers its performance on language models, and that the format might require additional overhead inducing adjustments beyond what is presented in \cite{chmiel2025fp4wayfullyquantized} for language models up to 1B. We find that \texttt{UE5M3} scale yields better results than \texttt{MXFP4}, offering a compromise between range and precision, however requiring tensor scaling and SR to work for LLM training, introducing, albeit manageable, overhead. For further research on hardware supporting FP4 training, we'd recommend starting out with \texttt{MXFP4} and implementing fusable operations such as Hadamard transformation, SR, Tensor scaling, being mindful of nan-handling, carefully selecting the optimiser and finally exploring different scaling formats such as \texttt{UE5M3}. Finally, our work highlights that FP4 training dynamics may not be consistent across model scales, and leave this critical direction to further research.

\clearpage

\bibliography{iclr2025_conference}
\bibliographystyle{iclr2025_conference}
\clearpage
\appendix

\subsection{LLM use disclosure}

LLMs were used in writing this paper. LLMs were used to:

\begin{enumerate}
    \item Polish the writing, wording and condensing text 
    \item Parse tedious mathematical derivations into latex
    \item Parse tedious figures and tables into latex
    \item Helping write some of the code
\end{enumerate}

\subsection{Experimental results}
\label{additional_experiments}

We provide detailed experimental results in this section.

\textbf{Linear regression} We first consider a linear regression task with synthetically generated data $
y = X \cdot  w_{\text{true}} \quad \text{for} \quad X \in \mathbb{R}^{100000 \times 1024}, \; w_{\text{true}} \in \mathbb{R}^{1024 \times 1} \quad \text{with} \quad X_{ij}, (w_{\text{true}})_i \stackrel{\text{iid}}{\sim} \mathcal{N}(0, 1)
$. We present our results in \Cref{tab:regression_results}. We find that the StableSPAM optimiser finds a perfect solution. We find that stochastic rounding applied to an \texttt{E4M3} scale yielded the best trade-off in results. Additional gradients resulting from the \texttt{absmax} normalisation were not found to be helpful. Overall the StableSPAM optimiser remains the superior choice for regression.

\begin{table}[htp!]
\caption{Experimental results}
\label{tab:regression_results}
\resizebox{\textwidth}{!}{%
\begin{tabular}{llrrllllllllrlllrrl}
\toprule
Dataset & Source & \makecell{Val\\loss} & \makecell{Train\\loss} & Scale & \makecell{Block \\ size} & \makecell{Max \\ grad.} & \makecell{Quant.\\grad} & Hadamard & \makecell{Scale\\grad} & SR & Optimiser & \makecell{Loss\\scaling} & Round mode & \makecell{Tensor\\scaling} & \makecell{Tensor \\grad} & \makecell{Complexity\\points} & Score & \makecell{NaN\\mode} \\
\midrule
IMAGENET100 & Baseline & 1.383 & 0.014 & N/A & N/A & N/A & N/A & N/A & N/A & N/A & Adam & False & N/A & N/A & N/A & N/A & N/A & N/A \\
IMAGENET100 & Baseline & 1.750 & 0.078 & N/A & N/A & N/A & N/A & N/A & N/A & N/A & StableSPAM & False & N/A & N/A & N/A & N/A & N/A & N/A \\
IMAGENET100 & Best Score (Neg) & 1.391 & 0.018 & E4M3 & 16 & STE & STE & N/A & STE & None\_exact & Adam & True & TowardPositive & True & ignore & 1.000 & -0.006 & nearest\_subnormal \\
IMAGENET100 & Best Score (Neg) & 1.625 & 0.087 & E8M0 & 32 & STE & STE & N/A & STE & all\_activation\_exact & StableSPAM & True & TiesToEven & True & ignore & 2.000 & -0.350 & nearest\_subnormal \\
IMAGENET100 & Best Score (Pos) & 1.320 & 0.015 & E4M3 & 16 & STE & STE & N/A & STE & IntelFP4\_exact & Adam & True & Stochastic & False & N/A & 1.250 & 0.036 & nearest\_subnormal \\
IMAGENET100 & Best Score (Pos) & 1.312 & 0.014 & E8M0 & 32 & STE & STE & N/A & STE & None\_exact & Adam & True & TowardPositive & True & ignore & 1.000 & 0.051 & nearest\_subnormal \\
IMAGENET100 & Best loss MXFP4 & 1.312 & 0.014 & E8M0 & 32 & STE & STE & N/A & spline & None\_exact & Adam & True & TowardPositive & True & ignore & 2.500 & 0.020 & nearest\_subnormal \\
IMAGENET100 & Best loss NVFP4 & 1.320 & 0.015 & E4M3 & 16 & STE & STE & N/A & STE & IntelFP4\_exact & Adam & True & Stochastic & False & N/A & 1.250 & 0.036 & to\_one \\
IMAGENET100 & Pure FP4 & 12.188 & 8.112 & E4M3 & 16 & STE & STE & N/A & STE & None\_exact & Adam & False & TiesToEven & False & N/A & 0.000 & -7.814 & nearest\_subnormal \\
IMAGENET100 & Pure FP4 & 1.344 & 0.014 & E8M0 & 32 & STE & STE & N/A & STE & None\_exact & Adam & False & TiesToEven & False & N/A & 0.000 & 0.028 & nearest\_subnormal \\
big\_diffusion & Baseline & 0.135 & 0.128 & N/A & N/A & N/A & N/A & N/A & N/A & N/A & Adam & False & N/A & N/A & N/A & N/A & N/A & N/A \\
big\_diffusion & Baseline & 0.113 & 0.110 & N/A & N/A & N/A & N/A & N/A & N/A & N/A & StableSPAM & False & N/A & N/A & N/A & N/A & N/A & N/A \\
big\_diffusion & Best Score (Neg) & 0.117 & 0.113 & E4M3 & 16 & STE & STE & N/A & STE & IntelFP4\_exact & Adam & True & Stochastic & False & N/A & 1.250 & -0.051 & nearest\_subnormal \\
big\_diffusion & Best Score (Neg) & 0.113 & 0.108 & E8M0 & 32 & STE & STE & N/A & STE & all\_activation\_exact & StableSPAM & False & Stochastic & False & N/A & 1.250 & -0.000 & nearest\_subnormal \\
big\_diffusion & Best Score (Pos) & 0.094 & 0.088 & E4M3 & 16 & STE & STE & N/A & STE & None\_exact & StableSPAM & False & TiesToEven & True & ignore & 1.000 & 0.166 & nearest\_subnormal \\
big\_diffusion & Best Score (Pos) & 0.102 & 0.095 & E8M0 & 32 & STE & STE & N/A & STE & None\_exact & StableSPAM & False & TiesToEven & False & N/A & 0.500 & 0.093 & nearest\_subnormal \\
big\_diffusion & Best loss MXFP4 & 0.102 & 0.095 & E8M0 & 32 & STE & STE & N/A & STE & None\_exact & StableSPAM & False & TiesToEven & False & N/A & 0.500 & 0.093 & nearest\_subnormal \\
big\_diffusion & Best loss NVFP4 & 0.094 & 0.088 & E4M3 & 16 & STE & STE & N/A & STE & None\_exact & StableSPAM & False & TiesToEven & True & ignore & 1.000 & 0.166 & nearest\_subnormal \\
big\_diffusion & Pure FP4 & 0.124 & 0.117 & E4M3 & 16 & STE & STE & N/A & STE & None\_exact & Adam & False & TowardPositive & False & N/A & 0.000 & -0.099 & nearest\_subnormal \\
big\_diffusion & Pure FP4 & 0.125 & 0.118 & E8M0 & 32 & STE & STE & N/A & STE & None\_exact & Adam & False & TowardPositive & False & N/A & 0.000 & -0.114 & nearest\_subnormal \\
gaussian\_reg & Baseline & 25.250 & 25.224 & N/A & N/A & N/A & N/A & N/A & N/A & N/A & Adam & False & N/A & N/A & N/A & N/A & N/A & N/A \\
gaussian\_reg & Baseline & 0.013 & 0.013 & N/A & N/A & N/A & N/A & N/A & N/A & N/A & StableSPAM & False & N/A & N/A & N/A & N/A & N/A & N/A \\
gaussian\_reg & Best Score (Neg) & 23.500 & 24.875 & E4M3 & 16 & STE & STE & N/A & STE & None\_exact & StableSPAM & False & Stochastic & False & N/A & 0.750 & -1815.151 & nearest\_subnormal \\
gaussian\_reg & Best Score (Neg) & 27.875 & 27.928 & E8M0 & 32 & STE & STE & N/A & STE & None\_exact & StableSPAM & False & TowardPositive & True & ignore & 1.000 & -2153.264 & nearest\_subnormal \\
gaussian\_reg & Best loss MXFP4 & 27.250 & 30.474 & E8M0 & 32 & STE & spline & backward\_exact & STE & None\_exact & StableSPAM & False & TowardPositive & True & ignore & 4.000 & -8419.849 & nearest\_subnormal \\
gaussian\_reg & Best loss NVFP4 & 22.875 & 24.467 & E4M3 & 16 & STE & spline & backward\_exact & STE & None\_exact & StableSPAM & True & TiesToEven & True & absmax & 7.500 & -13251.368 & nearest\_subnormal \\
gaussian\_reg & Pure FP4 & 30.125 & 30.947 & E4M3 & 16 & STE & STE & N/A & STE & None\_exact & Adam & False & TiesToEven & False & N/A & 0.000 & -2327.151 & nearest\_subnormal \\
gaussian\_reg & Pure FP4 & 34.000 & 34.303 & E8M0 & 32 & STE & STE & N/A & STE & None\_exact & Adam & False & TiesToEven & False & N/A & 0.000 & -2626.623 & nearest\_subnormal \\
\bottomrule
\end{tabular}
}
\end{table}


\textbf{Image classification} We find in \Cref{tab:regression_results} and Appendix \cref{tab:addtional_results}, that any max relaxation had no effect on performance. Here we find that in some cases using \texttt{absmax} in the tensor scaling gradient tends to help. We generally find that stochastic rounding, combined with tensor scaling and loss scaling leads to the most effective improvements. Here Adam seems to work better overall. We observe that \texttt{NVFP4} does not work out of the box while \texttt{MXFP4} does.


\textbf{Diffusion} We find in \Cref{tab:regression_results} and and Appendix \cref{tab:addtional_results} that any application of absmax gradients does not have any positive effect. We find that it suffices to use loss and tensor scaling combined with the StableSPAM optimiser to achieve a good performance.

\textbf{LLM} We present our results in \Cref{tab:llm_results} and \Cref{fig:key_models}. We are unable to reproduce the findings of \cite{chmiel2025fp4wayfullyquantized}, where we contrastingly find \texttt{MXFP4} to outperform \texttt{NVFP4} for LLM training. We note that \cite{fishman2025scalingfp8trainingtrilliontoken} additionally uses  \texttt{SmoothSwiGLU} \cite{fishman2025scalingfp8trainingtrilliontoken} in their experiments which induces a non-fusable $\sim\mathcal{O}(n)$ overhead as it requires the \texttt{absmax} along one dimension of the tensor, which we have omitted in our main experiments. We include this in further experiments in \Cref{sswiglu} and find that it marginally improves performance, but still fails for the 1B model.

In contrast to \cite{tseng2025trainingllmsmxfp4, castro2025quartetnativefp4training}, we did not find that the combination of Hadamard transformation and SR yielded a significantly better result for \texttt{MXFP4}, suggesting that SR can possibly be omitted to reduce overhead. We do not find that the relaxation of quantisation gradients proposed in \cite{zhou2025efficientpretrainingexploringfp4} had any impact on stabilizing the training of LLMs.

\begin{table}[htp!]
\caption{LLM results}
\label{tab:llm_results}
\resizebox{\textwidth}{!}{%
\begin{tabular}{llrrllllllllrlllrrl}
\toprule
Dataset & Source & \makecell{Val\\loss} & \makecell{Train\\loss} & Scale & \makecell{Block \\ size} & \makecell{Max \\ grad.} & \makecell{Quant.\\grad} & Hadamard & \makecell{Scale\\grad} & SR & Optimiser & \makecell{Loss\\scaling} & Round mode & \makecell{Tensor\\scaling} & \makecell{Tensor \\grad} & \makecell{Complexity\\points} & Score & \makecell{NaN\\mode} \\
\midrule
llama\_1B & Baseline & 3.578 & 3.682 & N/A & N/A & N/A & N/A & N/A & N/A & N/A & Adam & False & N/A & N/A & N/A & N/A & N/A & N/A \\
llama\_1B & Baseline & 3.487 & 3.569 & N/A & N/A & N/A & N/A & N/A & N/A & N/A & StableSPAM & False & N/A & N/A & N/A & N/A & N/A & N/A \\
llama\_1B & Best Score (Neg) & 4.933 & 4.957 & E4M3 & 16 & STE & STE & N/A & STE & None\_exact & StableSPAM & True & TiesToEven & True & ignore & 1.500 & -0.622 & nearest\_subnormal \\
llama\_1B & Best Score (Neg) & 3.620 & 3.701 & E8M0 & 32 & STE & STE & all\_exact & STE & None\_exact & StableSPAM & False & TiesToEven & False & N/A & 1.500 & -0.057 & nearest\_subnormal \\
llama\_1B & Best loss MXFP4 & 3.608 & 3.688 & E8M0 & 32 & STE & STE & all\_exact & STE & IntelFP4\_exact & StableSPAM & False & TiesToEven & False & N/A & 2.000 & -0.069 & nearest\_subnormal \\
llama\_1B & Best loss NVFP4 & 4.933 & 4.957 & E4M3 & 16 & STE & STE & N/A & STE & None\_exact & StableSPAM & True & TiesToEven & True & ignore & 1.500 & -0.622 & nearest\_subnormal \\
llama\_1B & Pure FP4 & 6.815 & 6.789 & E4M3 & 16 & STE & STE & N/A & STE & None\_exact & Adam & False & TiesToEven & False & N/A & 0.000 & -0.954 & nearest\_subnormal \\
llama\_1B & Pure FP4 & 3.864 & 3.932 & E8M0 & 32 & STE & STE & N/A & STE & None\_exact & Adam & False & TiesToEven & False & N/A & 0.000 & -0.108 & nearest\_subnormal \\
llama\_350M & Baseline & 2.269 & 2.375 & N/A & N/A & N/A & N/A & N/A & N/A & N/A & Adam & False & N/A & N/A & N/A & N/A & N/A & N/A \\
llama\_350M & Baseline & 2.258 & 2.363 & N/A & N/A & N/A & N/A & N/A & N/A & N/A & StableSPAM & False & N/A & N/A & N/A & N/A & N/A & N/A \\
llama\_350M & Best Score (Neg) & 2.655 & 2.783 & E4M3 & 16 & STE & STE & N/A & STE & IntelFP4\_exact & StableSPAM & False & TiesToEven & True & ignore & 1.500 & -0.264 & nearest\_subnormal \\
llama\_350M & Best Score (Neg) & 2.371 & 2.485 & E8M0 & 32 & STE & STE & all\_exact & STE & None\_exact & StableSPAM & False & TiesToEven & False & N/A & 1.500 & -0.075 & nearest\_subnormal \\
llama\_350M & Best loss MXFP4 & 2.369 & 2.483 & E8M0 & 32 & STE & STE & all\_exact & STE & None\_exact & StableSPAM & False & TiesToEven & True & ignore & 2.000 & -0.098 & nearest\_subnormal \\
llama\_350M & Best loss NVFP4 & 2.653 & 2.781 & E4M3 & 16 & STE & STE & N/A & STE & IntelFP4\_exact & StableSPAM & True & TiesToEven & True & ignore & 2.000 & -0.350 & nearest\_subnormal \\
llama\_350M & Pure FP4 & 4.880 & 4.958 & E4M3 & 16 & STE & STE & N/A & STE & None\_exact & Adam & False & TiesToEven & False & N/A & 0.000 & -1.161 & nearest\_subnormal \\
llama\_350M & Pure FP4 & 2.603 & 2.731 & E8M0 & 32 & STE & STE & N/A & STE & None\_exact & Adam & False & TiesToEven & False & N/A & 0.000 & -0.153 & nearest\_subnormal \\
llama\_60M & Baseline & 2.665 & 2.657 & N/A & N/A & N/A & N/A & N/A & N/A & N/A & Adam & False & N/A & N/A & N/A & N/A & N/A & N/A \\
llama\_60M & Baseline & 2.983 & 3.028 & N/A & N/A & N/A & N/A & N/A & N/A & N/A & StableSPAM & False & N/A & N/A & N/A & N/A & N/A & N/A \\
llama\_60M & Best Score (Neg) & 2.864 & 2.860 & E4M3 & 16 & STE & STE & N/A & STE & IntelFP4\_exact & Adam & False & TiesToEven & True & ignore & 1.000 & -0.074 & nearest\_subnormal \\
llama\_60M & Best Score (Neg) & 2.917 & 2.908 & E8M0 & 32 & STE & STE & all\_exact & STE & None\_exact & Adam & False & TiesToEven & False & N/A & 1.000 & -0.094 & nearest\_subnormal \\
llama\_60M & Best loss MXFP4 & 2.889 & 2.880 & E8M0 & 32 & STE & STE & all\_exact & STE & None\_exact & StableSPAM & True & TiesToEven & True & ignore & 2.500 & -0.210 & to\_one \\
llama\_60M & Best loss NVFP4 & 2.856 & 2.852 & E4M3 & 16 & STE & STE & N/A & STE & IntelFP4\_exact & StableSPAM & False & TiesToEven & True & ignore & 1.500 & -0.107 & nearest\_subnormal \\
llama\_60M & Pure FP4 & 4.838 & 4.829 & E4M3 & 16 & STE & STE & N/A & STE & None\_exact & Adam & False & TiesToEven & False & N/A & 0.000 & -0.815 & nearest\_subnormal \\
llama\_60M & Pure FP4 & 3.099 & 3.096 & E8M0 & 32 & STE & STE & N/A & STE & None\_exact & Adam & False & TiesToEven & False & N/A & 0.000 & -0.163 & nearest\_subnormal \\
\bottomrule
\end{tabular}
}
\end{table}

\paragraph{Exploring \texttt{UE5M3} scale format} We find in ablation studies (see \Cref{E8M3_results}) for \texttt{E4M3}, that the limiting factor during LLM training (with tensor scaling only) is the range of the exponent. We explore whether an alternative format like \texttt{UE5M3} can achieve better performance than \texttt{MXFP4} in \Cref{ue5m3_results}. Our results suggest that \texttt{UE5M3} offers a good compromise, with improved performance compared to \texttt{E8M0} scale on language modelling tasks. A caveat however is that \texttt{UE5M3} needs tensor scaling and SR in the backwards pass to stabilise, and exhibits instability in its pure form, unlike \texttt{MXFP4}. There is thus a computational overhead needed for the increased precision. We note that the best nan-handling strategy changes to ``to\_one''.

\begin{figure}[htbp!]
\centering
\caption{Training and validation performance curves for other datasets.}
\label{fig:other_models}

\begin{subfigure}[b]{0.25\textwidth}
    \centering
    \includegraphics[width=\textwidth]{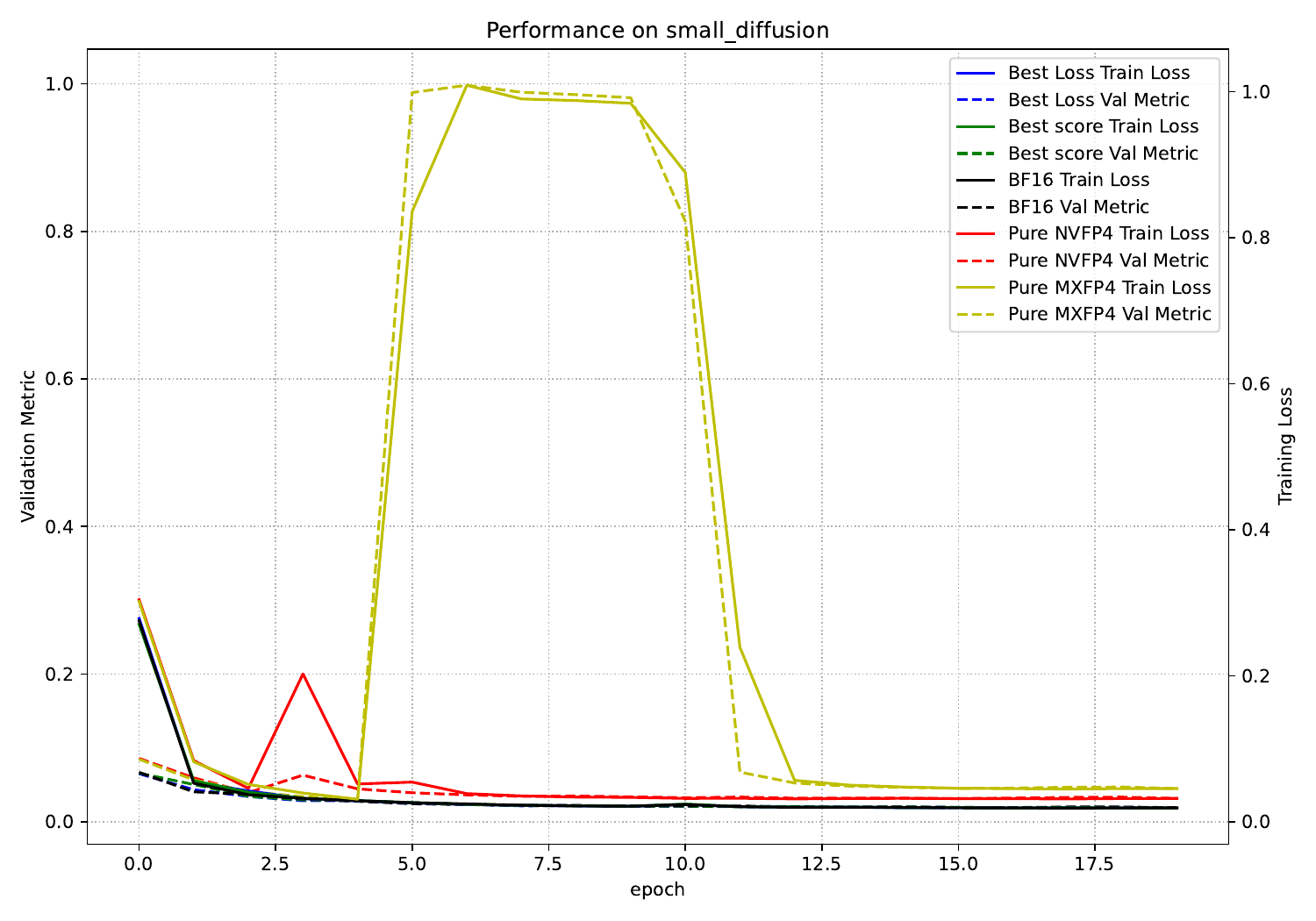}
    \caption{Small Diffusion}
    \label{fig:small_diffusion}
\end{subfigure}\hfill
\begin{subfigure}[b]{0.25\textwidth}
    \centering
    \includegraphics[width=\textwidth]{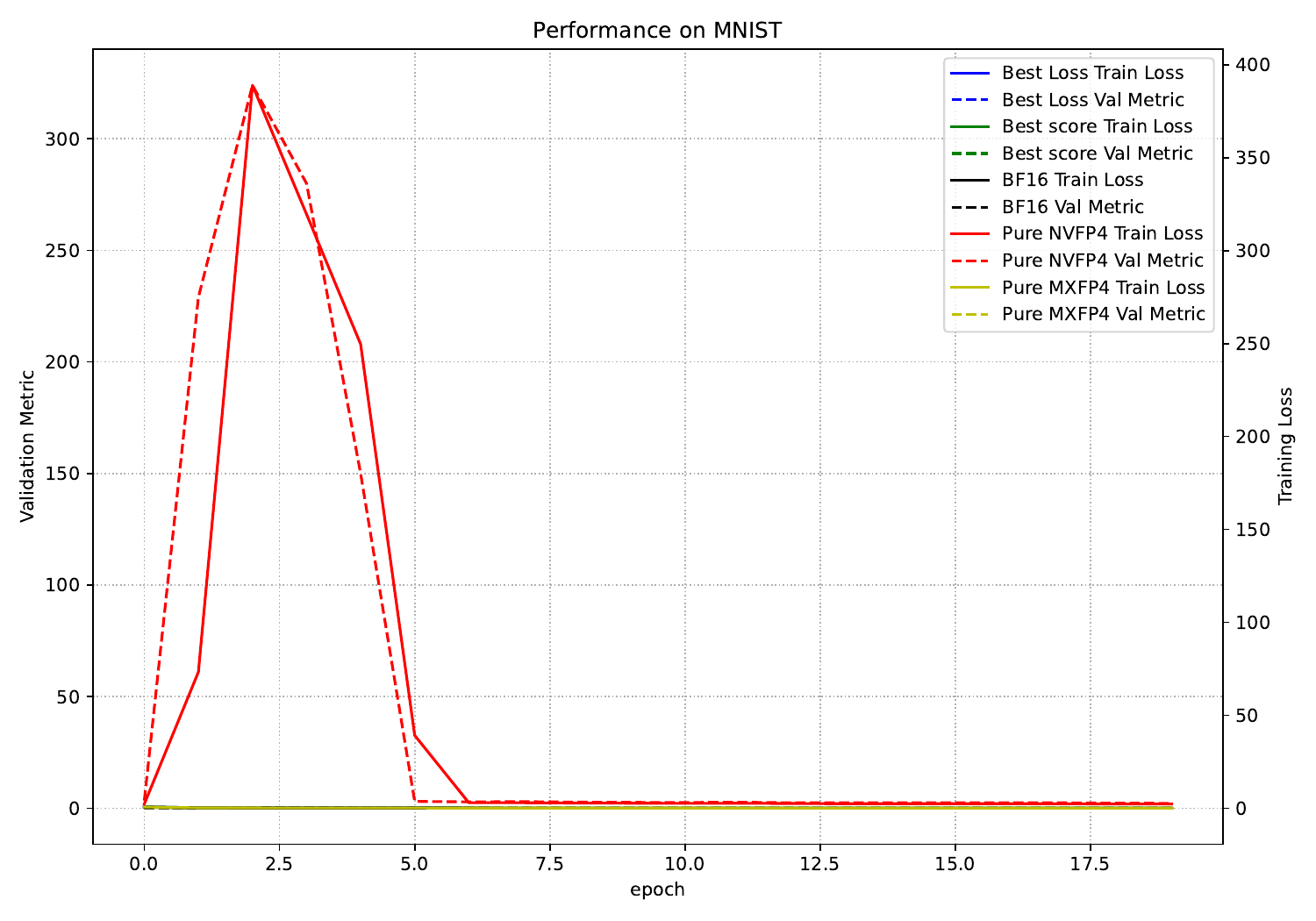}
    \caption{MNIST}
    \label{fig:mnist}
\end{subfigure}\hfill
\begin{subfigure}[b]{0.25\textwidth}
    \centering
    \includegraphics[width=\textwidth]{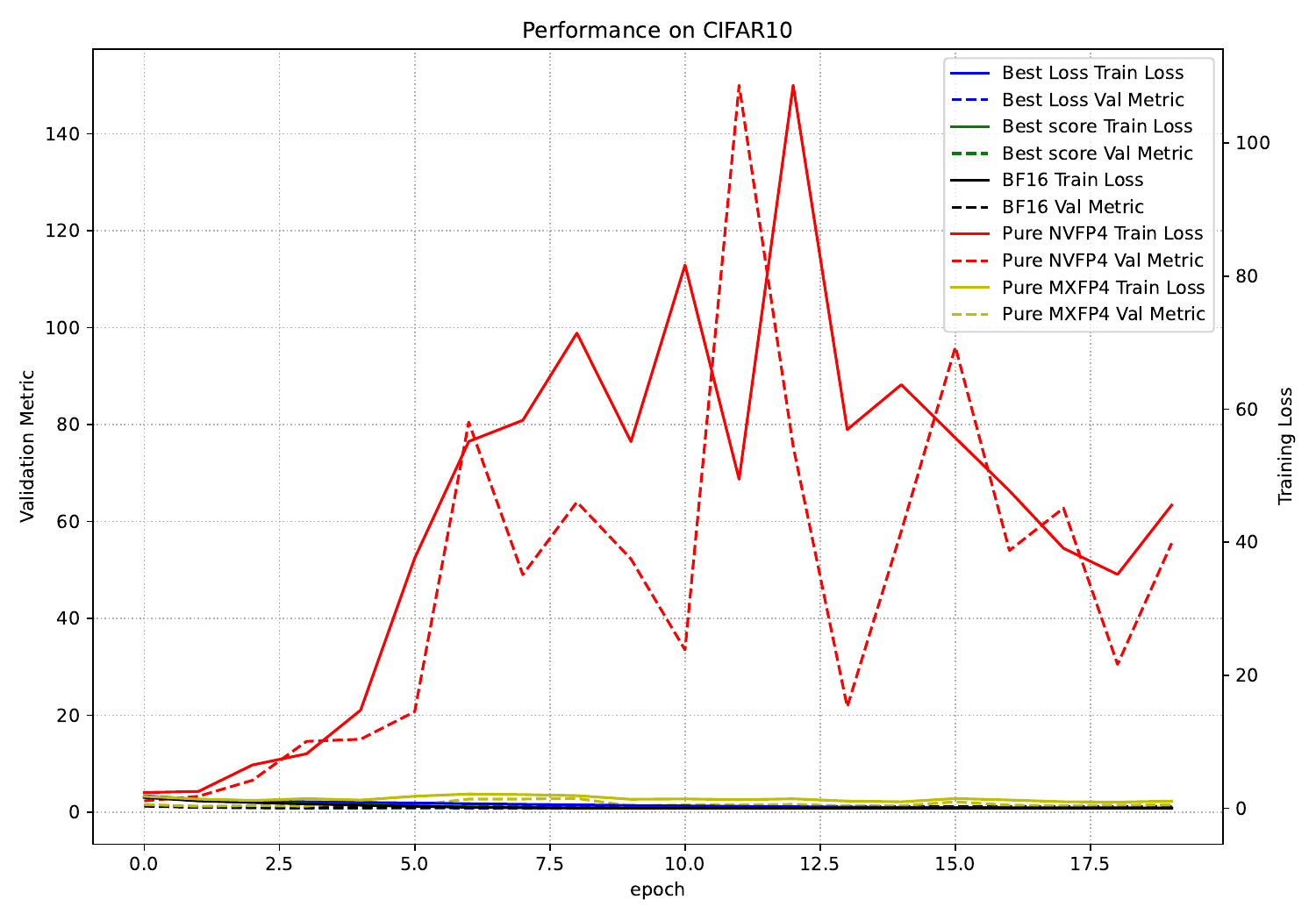}
    \caption{CIFAR-10}
    \label{fig:cifar10}
\end{subfigure}\hfill
\begin{subfigure}[b]{0.25\textwidth}
    \centering
    \includegraphics[width=\textwidth]{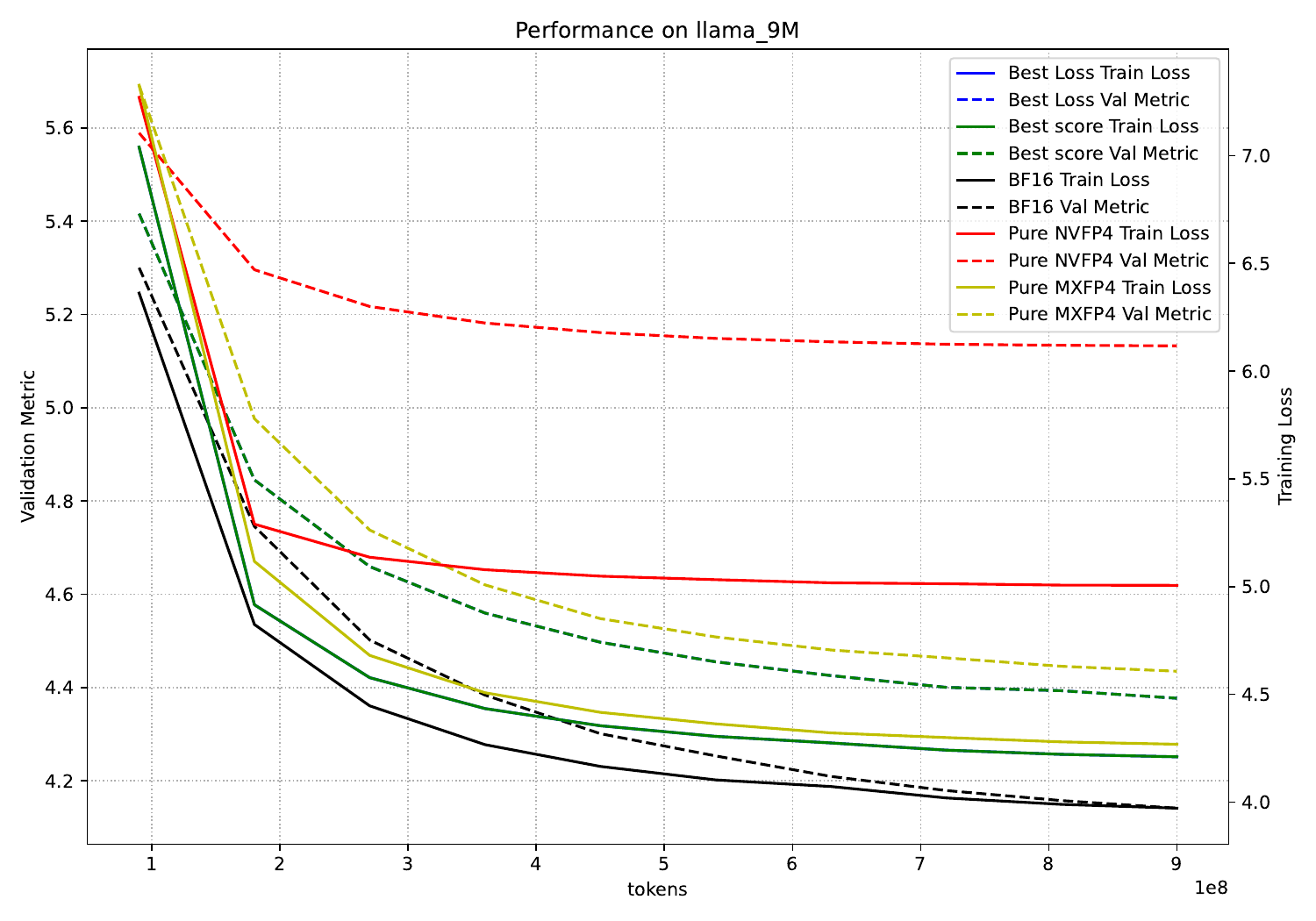}
    \caption{Llama 9M}
    \label{fig:llama_9m}
\end{subfigure}

\end{figure}

\paragraph{Additional dataset results} We present the additional results for MNIST, CIFAR10, Llama 9M and Small U-net (CIFAR 10) in \Cref{fig:other_models} and \Cref{tab:addtional_results}.

\begin{table}[htpb!]
\caption{Additional experimental results}
\label{tab:addtional_results}
\resizebox{\textwidth}{!}{%
\begin{tabular}{llrrllllllllrlllrrl}
\toprule
Dataset & Source & \makecell{Val\\loss} & \makecell{Train\\loss} & Scale & \makecell{Block \\ size} & \makecell{Max \\ grad.} & \makecell{Quant.\\grad} & Hadamard & \makecell{Scale\\grad} & SR & Optimiser & \makecell{Loss\\scaling} & Round mode & \makecell{Tensor\\scaling} & \makecell{Tensor \\grad} & \makecell{Complexity\\points} & Score & \makecell{NaN\\mode} \\
\midrule
CIFAR10 & Baseline & 0.875 & 0.003 & N/A & N/A & N/A & N/A & N/A & N/A & N/A & Adam & False & N/A & N/A & N/A & N/A & N/A & N/A \\
CIFAR10 & Baseline & 0.895 & 0.027 & N/A & N/A & N/A & N/A & N/A & N/A & N/A & StableSPAM & False & N/A & N/A & N/A & N/A & N/A & N/A \\
CIFAR10 & Best Score (Neg) & 0.883 & 0.005 & E4M3 & 16 & STE & STE & N/A & STE & None\_exact & Adam & True & TowardPositive & True & ignore & 1.000 & -0.009 & nearest\_subnormal \\
CIFAR10 & Best Score (Neg) & 0.883 & 0.003 & E8M0 & 32 & STE & STE & N/A & STE & IntelFP4\_exact & Adam & False & TiesToEven & True & ignore & 1.000 & -0.009 & nearest\_subnormal \\
CIFAR10 & Best Score (Pos) & 0.867 & 0.003 & E4M3 & 16 & STE & STE & N/A & STE & all\_activation\_exact & Adam & True & TiesToEven & False & N/A & 1.000 & 0.009 & nearest\_subnormal \\
CIFAR10 & Best Score (Pos) & 0.855 & 0.040 & E8M0 & 32 & STE & STE & N/A & STE & all\_activation\_exact & Adam & True & TiesToEven & True & absmax & 4.500 & 0.005 & nearest\_subnormal \\
CIFAR10 & Best loss MXFP4 & 0.855 & 0.040 & E8M0 & 32 & STE & spline & N/A & STE & all\_activation\_exact & Adam & True & TiesToEven & True & absmax & 6.500 & 0.003 & nearest\_subnormal \\
CIFAR10 & Best loss NVFP4 & 0.836 & 0.037 & E4M3 & 16 & STE & spline & N/A & STE & None\_exact & Adam & True & TowardPositive & True & absmax & 7.500 & 0.006 & nearest\_subnormal \\
CIFAR10 & Pure FP4 & 2.344 & 2.354 & E4M3 & 16 & STE & STE & N/A & STE & None\_exact & Adam & False & TowardPositive & False & N/A & 0.000 & -1.679 & nearest\_subnormal \\
CIFAR10 & Pure FP4 & 1.227 & 0.911 & E8M0 & 32 & STE & STE & N/A & STE & None\_exact & Adam & False & TiesToEven & False & N/A & 0.000 & -0.402 & nearest\_subnormal \\
MNIST & Baseline & 0.027 & 0.016 & N/A & N/A & N/A & N/A & N/A & N/A & N/A & Adam & False & N/A & N/A & N/A & N/A & N/A & N/A \\
MNIST & Baseline & 0.028 & 0.004 & N/A & N/A & N/A & N/A & N/A & N/A & N/A & StableSPAM & False & N/A & N/A & N/A & N/A & N/A & N/A \\
MNIST & Best Score (Neg) & 0.027 & 0.008 & E4M3 & 16 & STE & STE & N/A & STE & None\_exact & StableSPAM & True & TowardPositive & False & N/A & 1.000 & -0.005 & nearest\_subnormal \\
MNIST & Best Score (Neg) & 0.027 & 0.007 & E8M0 & 32 & STE & spline & N/A & STE & None\_exact & StableSPAM & True & Stochastic & True & ignore & 3.750 & -0.051 & nearest\_subnormal \\
MNIST & Best Score (Pos) & 0.021 & 0.009 & E4M3 & 16 & STE & STE & N/A & STE & None\_exact & StableSPAM & True & Stochastic & True & absmax & 4.750 & 0.050 & nearest\_subnormal \\
MNIST & Best Score (Pos) & 0.021 & 0.006 & E8M0 & 32 & STE & STE & N/A & STE & IntelFP4\_exact & StableSPAM & True & TiesToEven & True & absmax & 5.000 & 0.043 & nearest\_subnormal \\
MNIST & Best loss MXFP4 & 0.021 & 0.006 & E8M0 & 32 & STE & STE & N/A & STE & IntelFP4\_exact & StableSPAM & True & TiesToEven & True & absmax & 5.000 & 0.043 & nearest\_subnormal \\
MNIST & Best loss NVFP4 & 0.021 & 0.009 & E4M3 & 16 & STE & STE & N/A & STE & None\_exact & StableSPAM & True & Stochastic & True & absmax & 4.750 & 0.050 & to\_one \\
MNIST & Pure FP4 & 2.188 & 2.258 & E4M3 & 16 & STE & STE & N/A & STE & None\_exact & Adam & False & TiesToEven & False & N/A & 0.000 & -80.086 & to\_one \\
MNIST & Pure FP4 & 0.047 & 0.044 & E8M0 & 32 & STE & STE & N/A & STE & None\_exact & Adam & False & TiesToEven & False & N/A & 0.000 & -0.738 & nearest\_subnormal \\
llama\_9M & Baseline & 4.183 & 4.013 & N/A & N/A & N/A & N/A & N/A & N/A & N/A & Adam & False & N/A & N/A & N/A & N/A & N/A & N/A \\
llama\_9M & Baseline & 4.141 & 3.972 & N/A & N/A & N/A & N/A & N/A & N/A & N/A & StableSPAM & False & N/A & N/A & N/A & N/A & N/A & N/A \\
llama\_9M & Best Score (Neg) & 4.433 & 4.271 & E4M3 & 16 & STE & STE & N/A & STE & IntelFP4\_exact & Adam & False & TiesToEven & True & ignore & 1.000 & -0.071 & nearest\_subnormal \\
llama\_9M & Best Score (Neg) & 4.377 & 4.210 & E8M0 & 32 & STE & STE & N/A & STE & None\_exact & StableSPAM & False & TiesToEven & False & N/A & 0.500 & -0.057 & nearest\_subnormal \\
llama\_9M & Best loss MXFP4 & 4.377 & 4.210 & E8M0 & 32 & STE & STE & N/A & STE & None\_exact & StableSPAM & False & TiesToEven & False & N/A & 0.500 & -0.057 & nearest\_subnormal \\
llama\_9M & Best loss NVFP4 & 4.408 & 4.245 & E4M3 & 16 & STE & STE & N/A & STE & IntelFP4\_exact & StableSPAM & True & TiesToEven & True & ignore & 2.000 & -0.129 & nearest\_subnormal \\
llama\_9M & Pure FP4 & 5.133 & 5.006 & E4M3 & 16 & STE & STE & N/A & STE & None\_exact & Adam & False & TiesToEven & False & N/A & 0.000 & -0.239 & to\_one \\
llama\_9M & Pure FP4 & 4.435 & 4.268 & E8M0 & 32 & STE & STE & N/A & STE & None\_exact & Adam & False & TiesToEven & False & N/A & 0.000 & -0.071 & nearest\_subnormal \\
small\_diffusion & Baseline & 0.029 & 0.029 & N/A & N/A & N/A & N/A & N/A & N/A & N/A & Adam & False & N/A & N/A & N/A & N/A & N/A & N/A \\
small\_diffusion & Baseline & 0.019 & 0.019 & N/A & N/A & N/A & N/A & N/A & N/A & N/A & StableSPAM & False & N/A & N/A & N/A & N/A & N/A & N/A \\
small\_diffusion & Best Score (Neg) & 0.019 & 0.019 & E4M3 & 16 & STE & STE & N/A & STE & None\_exact & StableSPAM & True & TiesToEven & True & ignore & 1.500 & -0.001 & nearest\_subnormal \\
small\_diffusion & Best Score (Neg) & 0.019 & 0.019 & E8M0 & 32 & STE & STE & all\_exact & STE & None\_exact & StableSPAM & False & TiesToEven & True & ignore & 2.000 & -0.000 & nearest\_subnormal \\
small\_diffusion & Best Score (Pos) & 0.018 & 0.019 & E4M3 & 16 & STE & STE & N/A & STE & IntelFP4\_exact & StableSPAM & True & TowardPositive & False & N/A & 1.500 & 0.020 & nearest\_subnormal \\
small\_diffusion & Best Score (Pos) & 0.018 & 0.019 & E8M0 & 32 & STE & STE & N/A & STE & IntelFP4\_exact & StableSPAM & False & TiesToEven & False & N/A & 1.000 & 0.028 & nearest\_subnormal \\
small\_diffusion & Best loss MXFP4 & 0.018 & 0.019 & E8M0 & 32 & STE & STE & N/A & STE & IntelFP4\_exact & StableSPAM & False & TiesToEven & True & ignore & 1.500 & 0.020 & to\_one \\
small\_diffusion & Best loss NVFP4 & 0.018 & 0.019 & E4M3 & 16 & STE & baseline & N/A & STE & IntelFP4\_exact & StableSPAM & True & TiesToEven & True & ignore & 4.000 & 0.008 & nearest\_subnormal \\
small\_diffusion & Pure FP4 & 0.031 & 0.031 & E4M3 & 16 & STE & STE & N/A & STE & None\_exact & Adam & False & TiesToEven & False & N/A & 0.000 & -0.656 & nearest\_subnormal \\
small\_diffusion & Pure FP4 & 0.030 & 0.031 & E8M0 & 32 & STE & STE & N/A & STE & None\_exact & Adam & False & TiesToEven & False & N/A & 0.000 & -0.572 & nearest\_subnormal \\
\bottomrule
\end{tabular}
}
\end{table}

\subsection{Testing \texttt{SmoothSwiGLU}, tensor scaling and SR}
\label{sswiglu}

\begin{figure}[htbp!]
\centering
\caption{Training and validation performance curves LLama with SSwiGLU. The gap between \text{BFLOAT16} still grows with model size depsite tensor scaling and SR.}
\label{fig:sswiglu}

\begin{subfigure}[b]{0.25\textwidth}
    \centering
    \includegraphics[width=\textwidth]{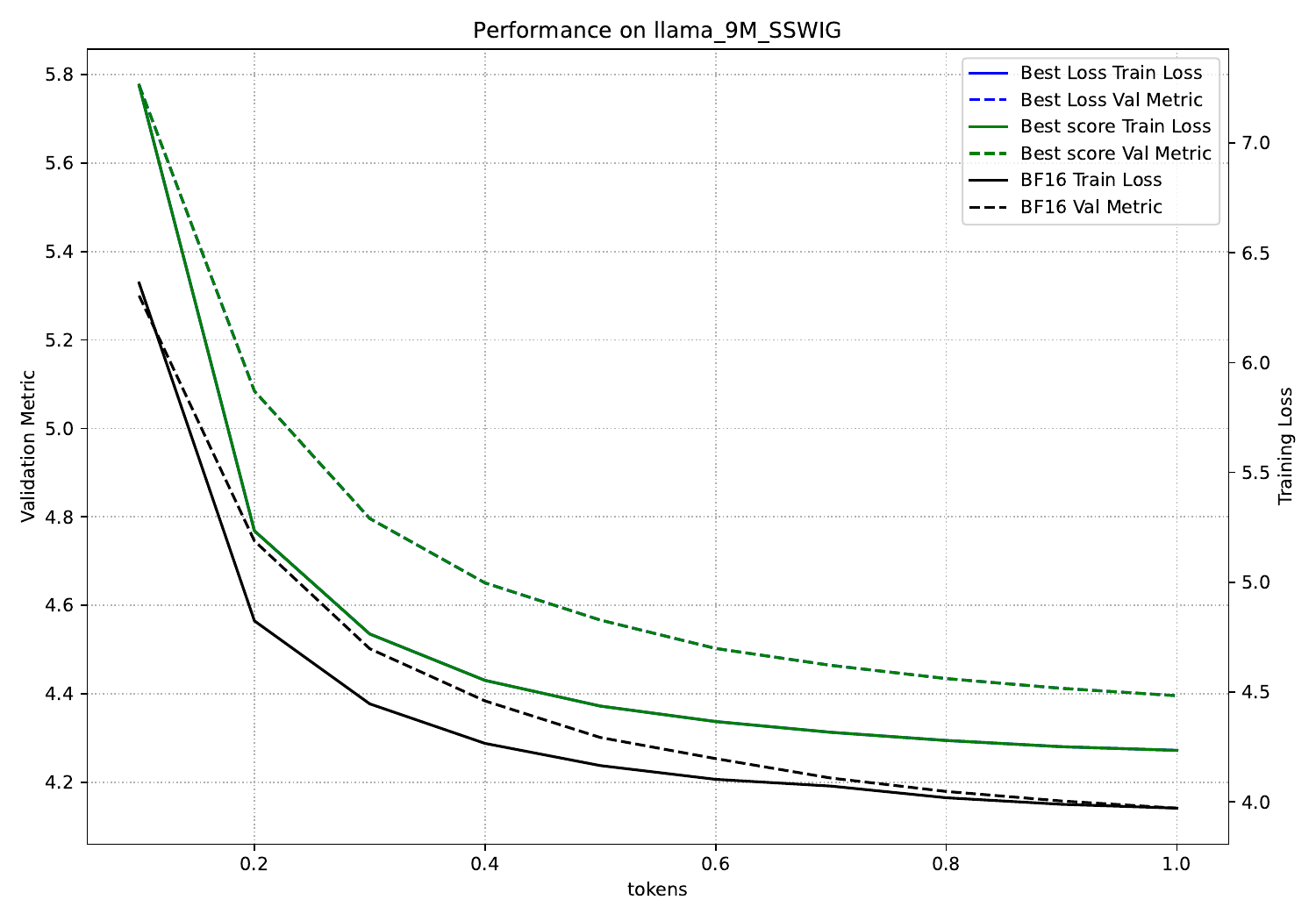}
    \caption{Llama 9M}
\end{subfigure}\hfill
\begin{subfigure}[b]{0.25\textwidth}
    \centering
    \includegraphics[width=\textwidth]{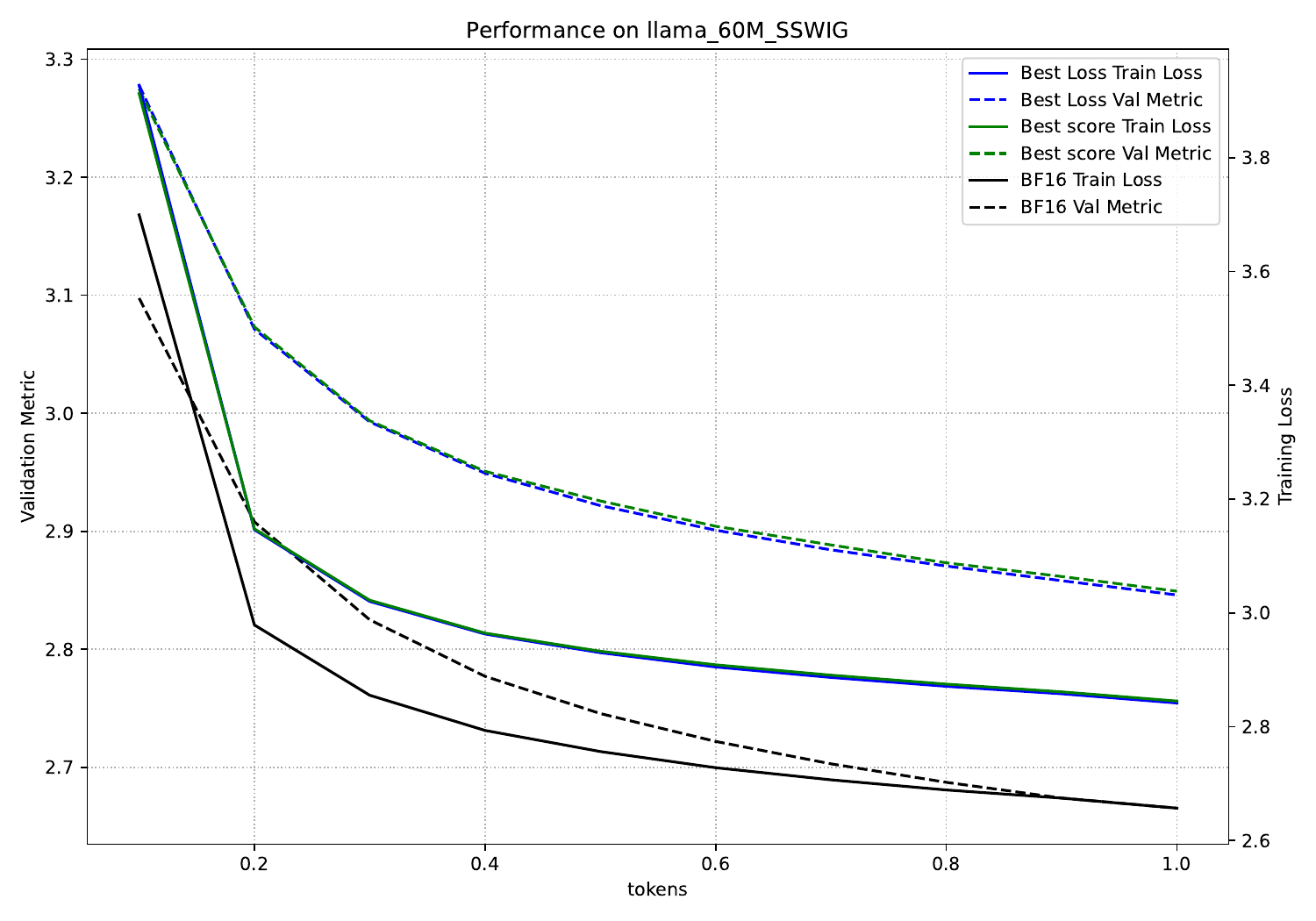}
    \caption{Llama 60M}
\end{subfigure}\hfill
\begin{subfigure}[b]{0.25\textwidth}
    \centering
    \includegraphics[width=\textwidth]{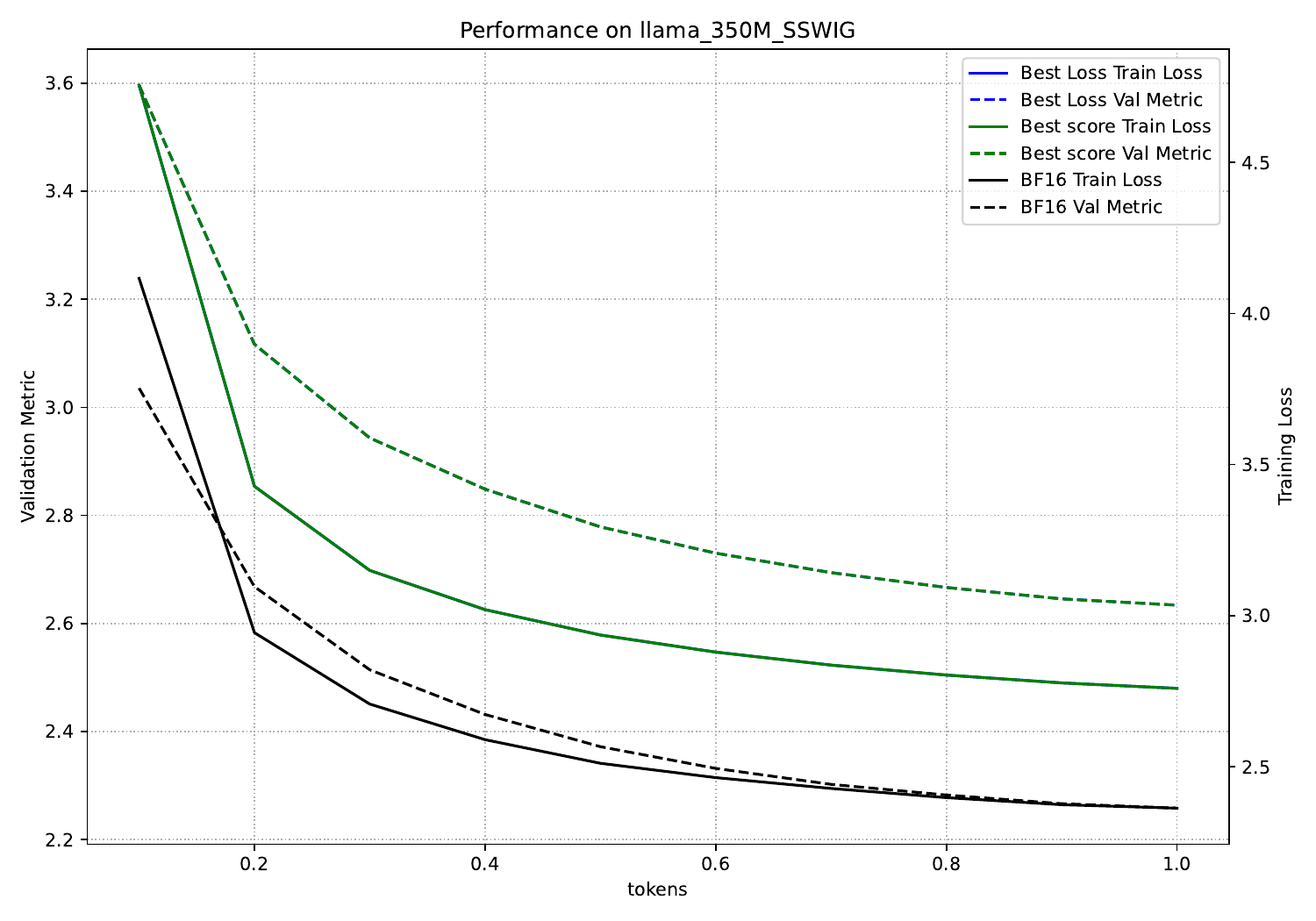}
    \caption{Llama 350M}
\end{subfigure}\hfill
\begin{subfigure}[b]{0.25\textwidth}
    \centering
    \includegraphics[width=\textwidth]{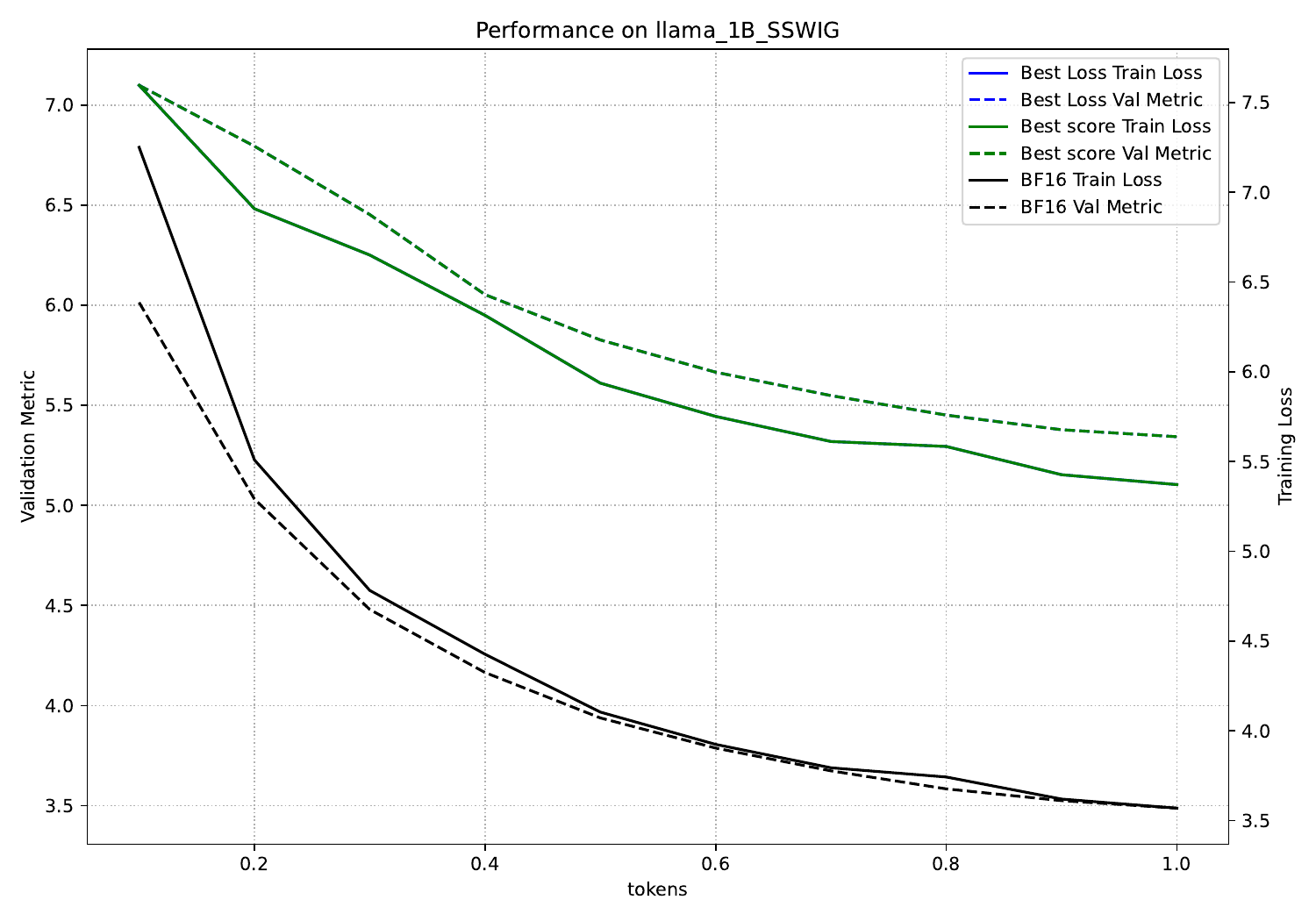}
    \caption{Llama 1B}
\end{subfigure}

\end{figure}

We replicate the results in \cite{chmiel2025fp4wayfullyquantized} more exactly by adding the \texttt{SmoothSwiGLU} in \cite{fishman2025scalingfp8trainingtrilliontoken}. We could not replicate their indicated results on models up to LLama 1B in \Cref{fig:sswiglu} and \Cref{tab:ablation_swig}.

\begin{table}[htpb!]
\caption{SWIG results}
\label{tab:ablation_swig}
\resizebox{\textwidth}{!}{%

\begin{tabular}{llrrllllllllrlllrrl}
\toprule
Dataset & Source & \makecell{Val\\loss} & \makecell{Train\\loss} & Scale & \makecell{Block \\ size} & \makecell{Max \\ grad.} & \makecell{Quant.\\grad} & Hadamard & \makecell{Scale\\grad} & SR & Optimiser & \makecell{Loss\\scaling} & Round mode & \makecell{Tensor\\scaling} & \makecell{Tensor \\grad} & \makecell{Complexity\\points} & Score & \makecell{NaN\\mode} \\
\midrule
llama\_1B & Baseline & 3.578 & 3.682 & N/A & N/A & N/A & N/A & N/A & N/A & N/A & Adam & False & N/A & N/A & N/A & N/A & N/A & N/A \\
llama\_1B & Baseline & 3.487 & 3.569 & N/A & N/A & N/A & N/A & N/A & N/A & N/A & StableSPAM & False & N/A & N/A & N/A & N/A & N/A & N/A \\
llama\_1B\_SSWIG & Best Score (Neg) & 5.343 & 5.372 & E4M3 & 16 & STE & STE & None\_exact & STE & None\_exact & StableSPAM & False & TiesToEven & True & ignore & 1.000 & -0.532 & to\_one \\
llama\_1B\_SSWIG & Best loss NVFP4 & 5.343 & 5.372 & E4M3 & 16 & STE & STE & None\_exact & STE & None\_exact & StableSPAM & False & TiesToEven & True & ignore & 1.000 & -0.532 & to\_one \\
llama\_350M & Baseline & 2.269 & 2.375 & N/A & N/A & N/A & N/A & N/A & N/A & N/A & Adam & False & N/A & N/A & N/A & N/A & N/A & N/A \\
llama\_350M & Baseline & 2.258 & 2.363 & N/A & N/A & N/A & N/A & N/A & N/A & N/A & StableSPAM & False & N/A & N/A & N/A & N/A & N/A & N/A \\
llama\_350M\_SSWIG & Best Score (Neg) & 2.634 & 2.760 & E4M3 & 16 & STE & STE & None\_exact & STE & IntelFP4\_exact & StableSPAM & False & TiesToEven & True & ignore & 1.500 & -0.250 & to\_one \\
llama\_350M\_SSWIG & Best loss NVFP4 & 2.634 & 2.760 & E4M3 & 16 & STE & STE & None\_exact & STE & IntelFP4\_exact & StableSPAM & False & TiesToEven & True & ignore & 1.500 & -0.250 & to\_one \\
llama\_60M & Baseline & 2.665 & 2.657 & N/A & N/A & N/A & N/A & N/A & N/A & N/A & Adam & False & N/A & N/A & N/A & N/A & N/A & N/A \\
llama\_60M & Baseline & 2.983 & 3.028 & N/A & N/A & N/A & N/A & N/A & N/A & N/A & StableSPAM & False & N/A & N/A & N/A & N/A & N/A & N/A \\
llama\_60M\_SSWIG & Best Score (Neg) & 2.849 & 2.845 & E4M3 & 16 & STE & STE & None\_exact & STE & IntelFP4\_exact & StableSPAM & False & TiesToEven & True & ignore & 1.500 & -0.103 & nearest\_subnormal \\
llama\_60M\_SSWIG & Best loss NVFP4 & 2.846 & 2.842 & E4M3 & 16 & STE & STE & None\_exact & STE & IntelFP4\_exact & StableSPAM & True & TiesToEven & True & ignore & 2.000 & -0.136 & to\_one \\
llama\_9M & Baseline & 4.183 & 4.013 & N/A & N/A & N/A & N/A & N/A & N/A & N/A & Adam & False & N/A & N/A & N/A & N/A & N/A & N/A \\
llama\_9M & Baseline & 4.141 & 3.972 & N/A & N/A & N/A & N/A & N/A & N/A & N/A & StableSPAM & False & N/A & N/A & N/A & N/A & N/A & N/A \\
llama\_9M\_SSWIG & Best Score (Neg) & 4.396 & 4.235 & E4M3 & 16 & STE & STE & None\_exact & STE & IntelFP4\_exact & StableSPAM & False & TiesToEven & True & ignore & 1.500 & -0.092 & nearest\_subnormal \\
llama\_9M\_SSWIG & Best loss NVFP4 & 4.396 & 4.235 & E4M3 & 16 & STE & STE & None\_exact & STE & IntelFP4\_exact & StableSPAM & False & TiesToEven & True & ignore & 1.500 & -0.092 & nearest\_subnormal \\
\bottomrule
\end{tabular}
}
\end{table}

\subsection{Changing the scale to \texttt{E8M3}}
\label{E8M3_results}
\begin{figure}[htpb!]
\centering
\caption{Training and validation performance curves for \texttt{E8M3}.}
\label{fig:e8m3_ablation}

\begin{subfigure}[b]{0.5\textwidth}
    \centering
    \includegraphics[width=\textwidth]{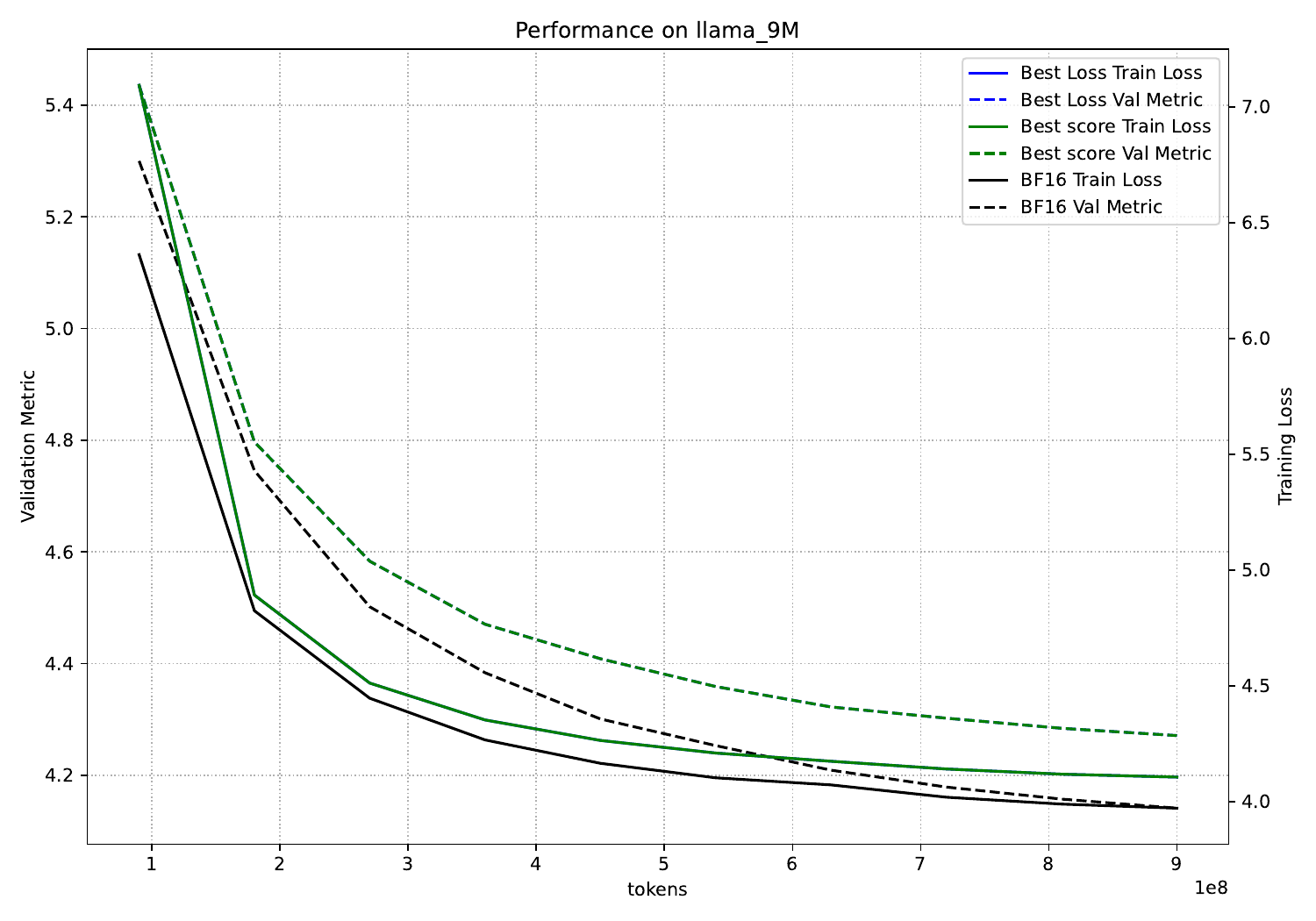}
    \caption{Llama 9M}
\end{subfigure}\hfill
\begin{subfigure}[b]{0.5\textwidth}
    \centering
    \includegraphics[width=\textwidth]{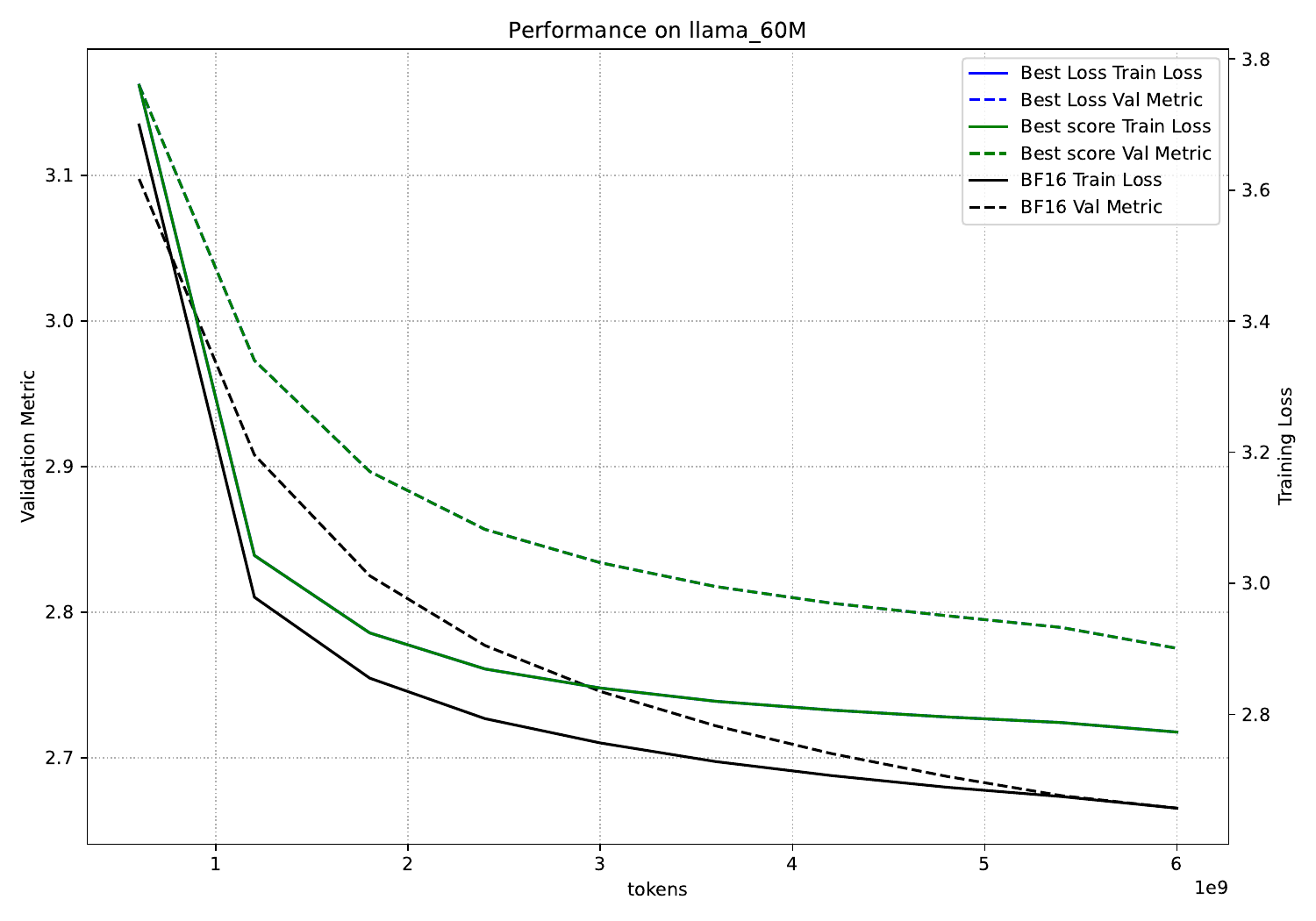}
    \caption{Llama 60M}
\end{subfigure}\hfill
\end{figure}

During our experiments, we noticed that \texttt{E4M3} did not match the performance of \texttt{E8M0}, even with tensor scaling. We speculated that the range of \texttt{E4M3} was the issue and decided to verify this with an ablation study using \texttt{E8M3} to test this hypothesis. We presents the results in \Cref{tab:ablation_e8m3} and \Cref{fig:e8m3_ablation}.
\begin{table}[htpb!]
\caption{E8M3 ablation results}
\label{tab:ablation_e8m3}
\resizebox{\textwidth}{!}{%
\begin{tabular}{llrrlrllllllrlllrrl}
\toprule
Dataset & Source & \makecell{Val\\loss} & \makecell{Train\\loss} & Scale & \makecell{Block \\ size} & \makecell{Max \\ grad.} & \makecell{Quant.\\grad} & Hadamard & \makecell{Scale\\grad} & SR & Optimiser & \makecell{Loss\\scaling} & Round mode & \makecell{Tensor\\scaling} & \makecell{Tensor \\grad} & \makecell{Complexity\\points} & Score & \makecell{NaN\\mode} \\
\midrule
llama\_60M & Baseline & 2.665 & 2.657 & N/A & N/A & N/A & N/A & N/A & N/A & N/A & Adam & False & N/A & N/A & N/A & N/A & N/A & N/A \\
llama\_60M & Baseline & 2.983 & 3.028 & N/A & N/A & N/A & N/A & N/A & N/A & N/A & StableSPAM & False & N/A & N/A & N/A & N/A & N/A & N/A \\
llama\_60M & Best Score (Neg) & 2.775 & 2.773 & E8M3 & 16.000 & STE & STE & None\_exact & STE & IntelFP4\_exact & StableSPAM & False & TiesToEven & False & N/A & 1.000 & -0.041 & nearest\_subnormal \\
llama\_60M & Best loss E8M3 & 2.775 & 2.773 & E8M3 & 16.000 & STE & STE & None\_exact & STE & IntelFP4\_exact & StableSPAM & False & TiesToEven & False & N/A & 1.000 & -0.041 & nearest\_subnormal \\
llama\_60M & Pure FP4 & 2.851 & 2.848 & E8M3 & 16.000 & STE & STE & None\_exact & STE & None\_exact & Adam & False & TiesToEven & False & N/A & 0.000 & -0.070 & nearest\_subnormal \\
llama\_9M & Baseline & 4.183 & 4.013 & N/A & N/A & N/A & N/A & N/A & N/A & N/A & Adam & False & N/A & N/A & N/A & N/A & N/A & N/A \\
llama\_9M & Baseline & 4.141 & 3.972 & N/A & N/A & N/A & N/A & N/A & N/A & N/A & StableSPAM & False & N/A & N/A & N/A & N/A & N/A & N/A \\
llama\_9M & Best Score (Neg) & 4.271 & 4.106 & E8M3 & 16.000 & STE & STE & None\_exact & STE & IntelFP4\_exact & StableSPAM & False & TiesToEven & False & N/A & 1.000 & -0.031 & nearest\_subnormal \\
llama\_9M & Best loss E8M3 & 4.271 & 4.106 & E8M3 & 16.000 & STE & STE & None\_exact & STE & IntelFP4\_exact & StableSPAM & False & TiesToEven & False & N/A & 1.000 & -0.031 & nearest\_subnormal \\
llama\_9M & Pure FP4 & 4.320 & 4.156 & E8M3 & 16.000 & STE & STE & None\_exact & STE & None\_exact & Adam & False & TiesToEven & False & N/A & 0.000 & -0.043 & nearest\_subnormal \\
\bottomrule
\end{tabular}
}
\end{table}

Confirming that the limiting factor of \texttt{E4M3} is the range, we next speculate that an 8-bit numerical scaling format in-between \texttt{E4M3} and \texttt{E8M0} might offer a good trade-off between range and precision.

\subsection{\texttt{UE5M3} results}
\label{ue5m3_results}

\begin{figure}[htpb!]
\centering
\caption{Training and validation performance curves for selected models and datasets of \texttt{UE5M3} experiments.}
\label{fig:UE5M3_plots}

\begin{subfigure}[b]{0.33\textwidth}
    \centering
    \includegraphics[width=\textwidth]{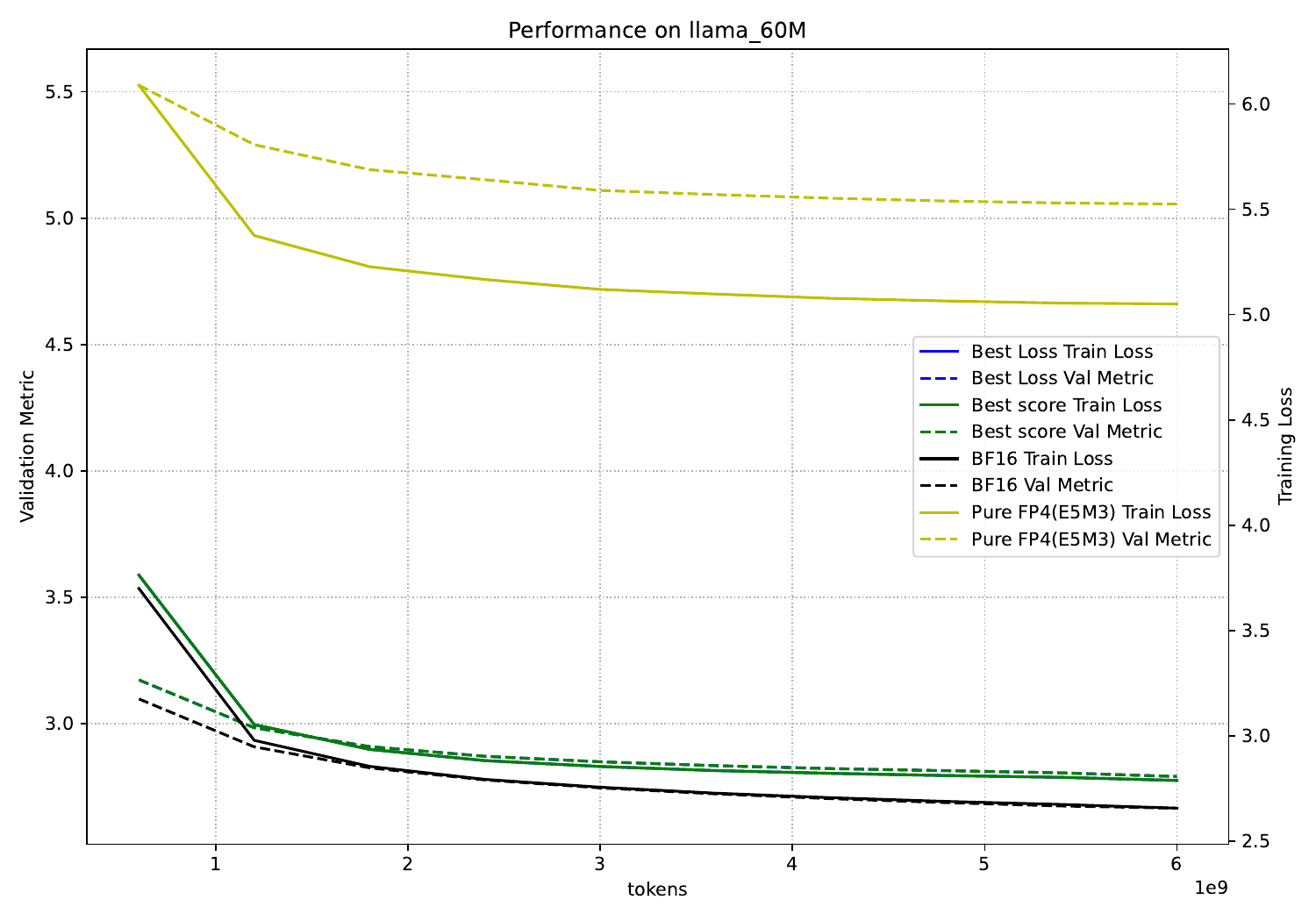}
    \caption{Llama 60M}
    \label{fig:llama_60m}
\end{subfigure}%
\begin{subfigure}[b]{0.33\textwidth}
    \centering
    \includegraphics[width=\textwidth]{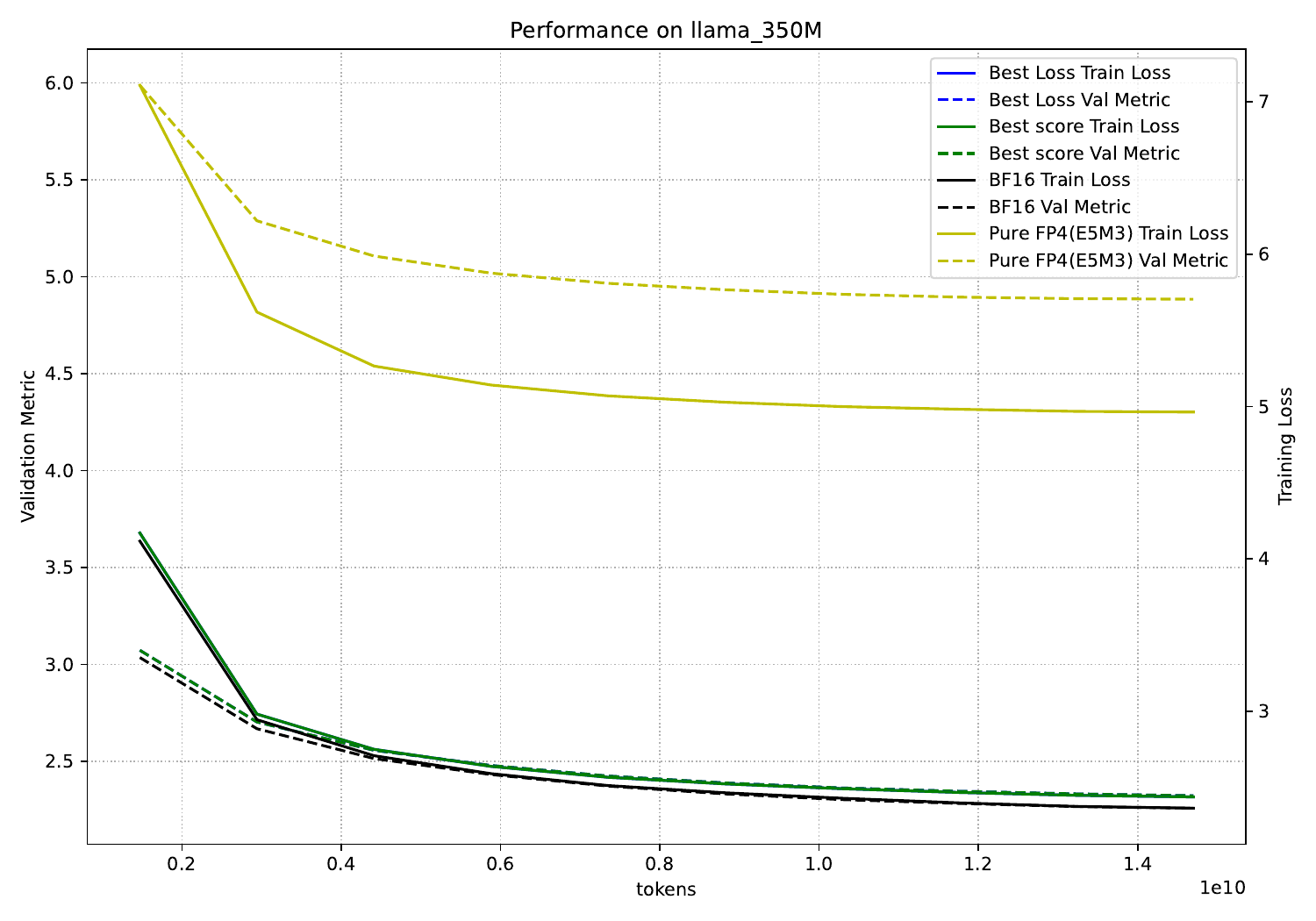}
    \caption{Llama 350M}
    \label{fig:llama_350m}
\end{subfigure}%
\begin{subfigure}[b]{0.33\textwidth}
    \centering
    \includegraphics[width=\textwidth]{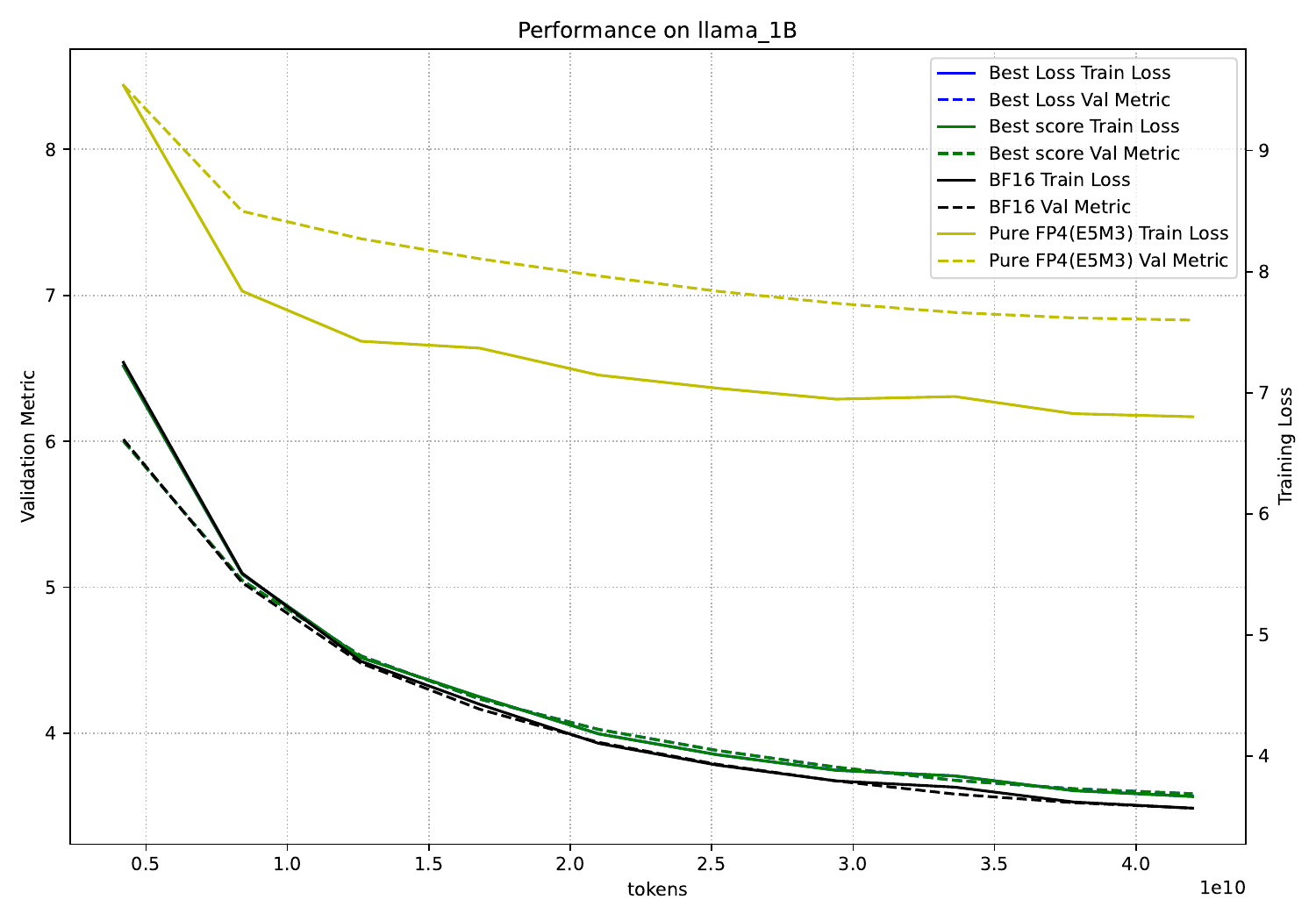}
    \caption{Llama 1B}
    \label{fig:llama_1b}
\end{subfigure}

\begin{subfigure}[b]{0.33\textwidth}
    \centering
    \includegraphics[width=\textwidth]{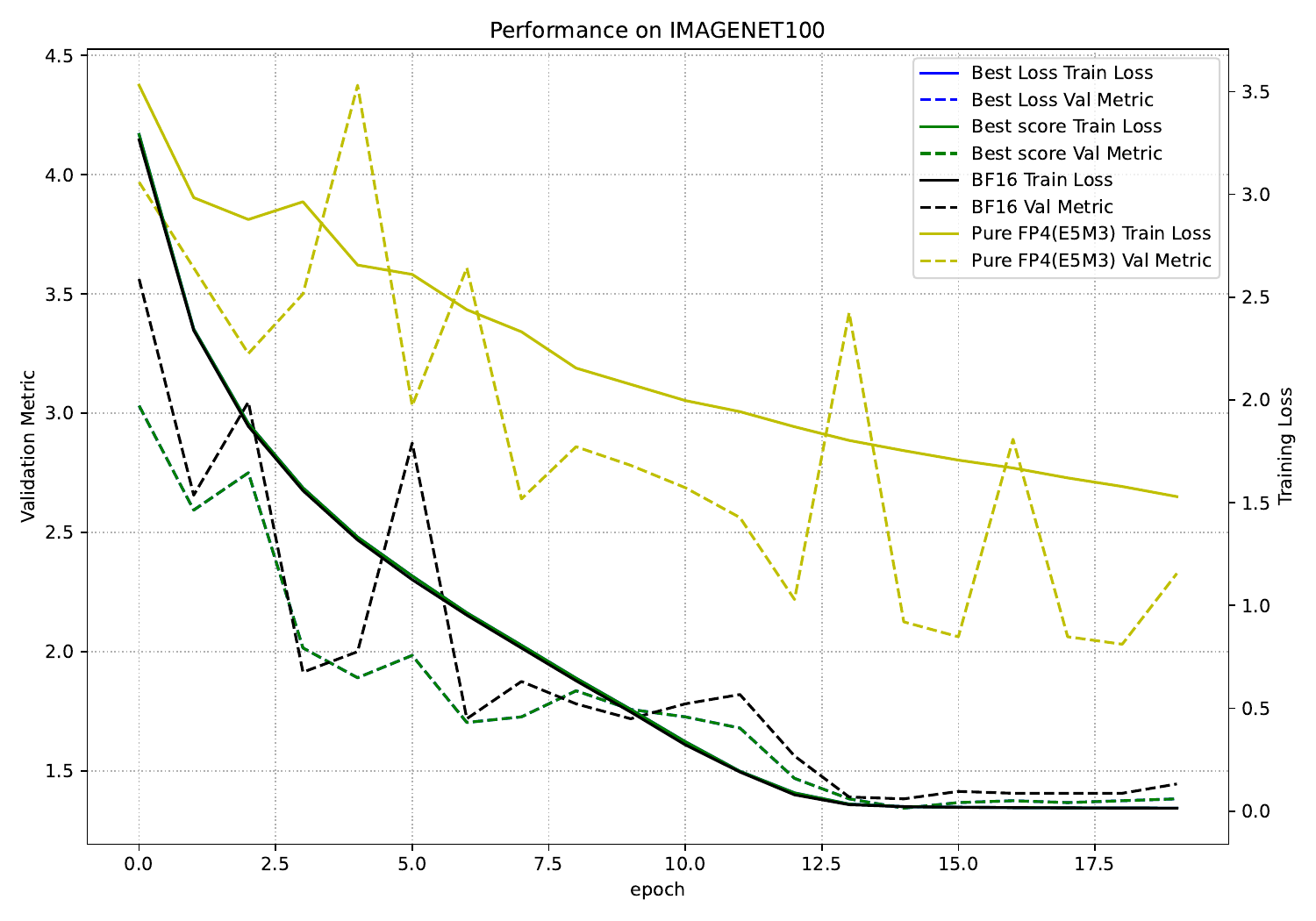}
    \caption{ImageNet-100}
    \label{fig:imagenet100}
\end{subfigure}%
\begin{subfigure}[b]{0.33\textwidth}
    \centering
    \includegraphics[width=\textwidth]{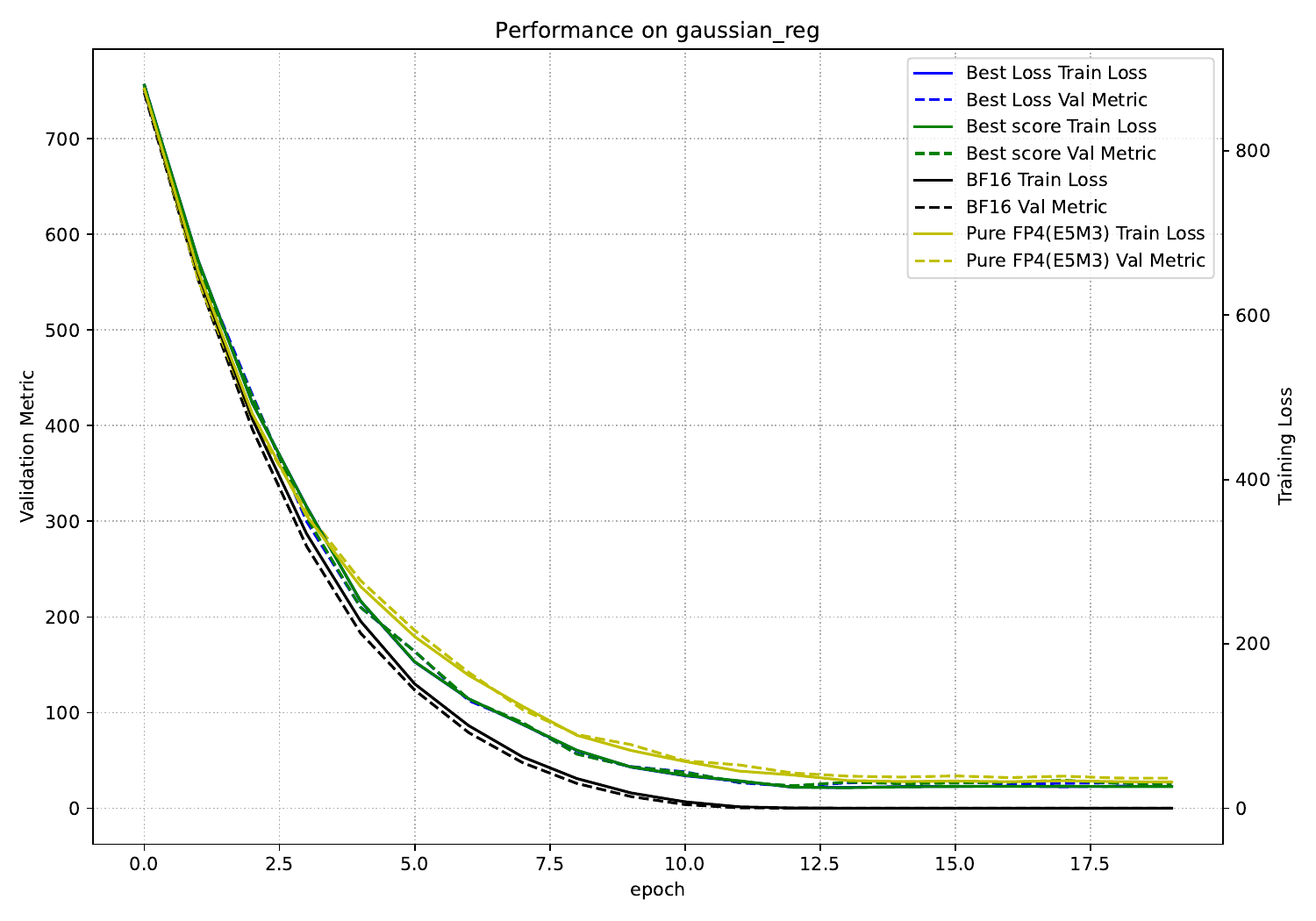}
    \caption{Gaussian Reg.}
    \label{fig:gaussian_reg}
\end{subfigure}%
\begin{subfigure}[b]{0.33\textwidth}
    \centering
    \includegraphics[width=\textwidth]{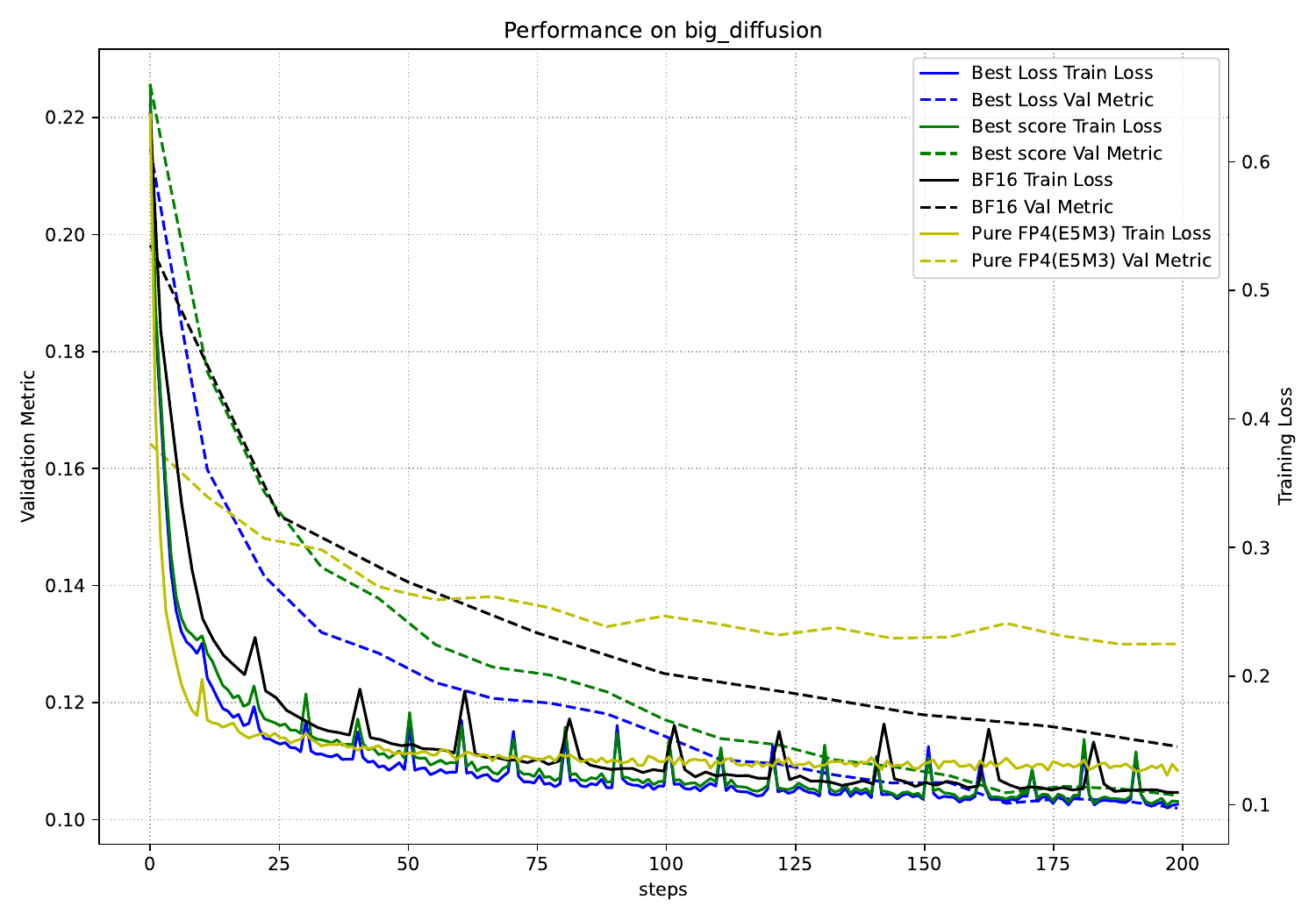}
    \caption{Big diffusion}
    \label{fig:big_diffusion}
\end{subfigure}

\end{figure}

\begin{figure}[htpb!]
\centering
\caption{Training and validation performance curves for additional dataset for the \texttt{UE5M3} scale}
\label{fig:other_models_ue5m3}

\begin{subfigure}[b]{0.25\textwidth}
    \centering
    \includegraphics[width=\textwidth]{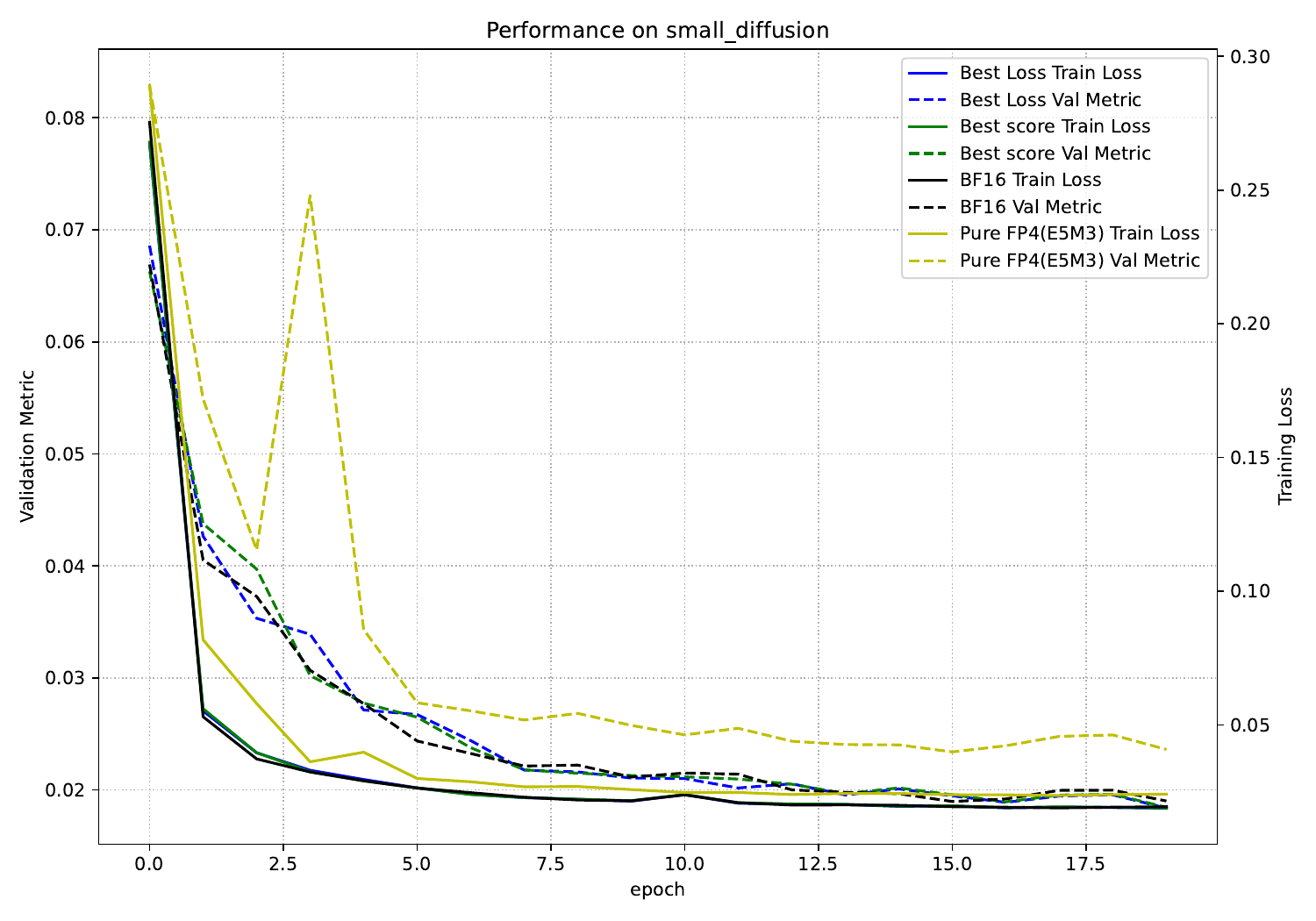}
    \caption{Small Diffusion}
    \label{fig:small_diffusion_ue5m3}
\end{subfigure}\hfill
\begin{subfigure}[b]{0.25\textwidth}
    \centering
    \includegraphics[width=\textwidth]{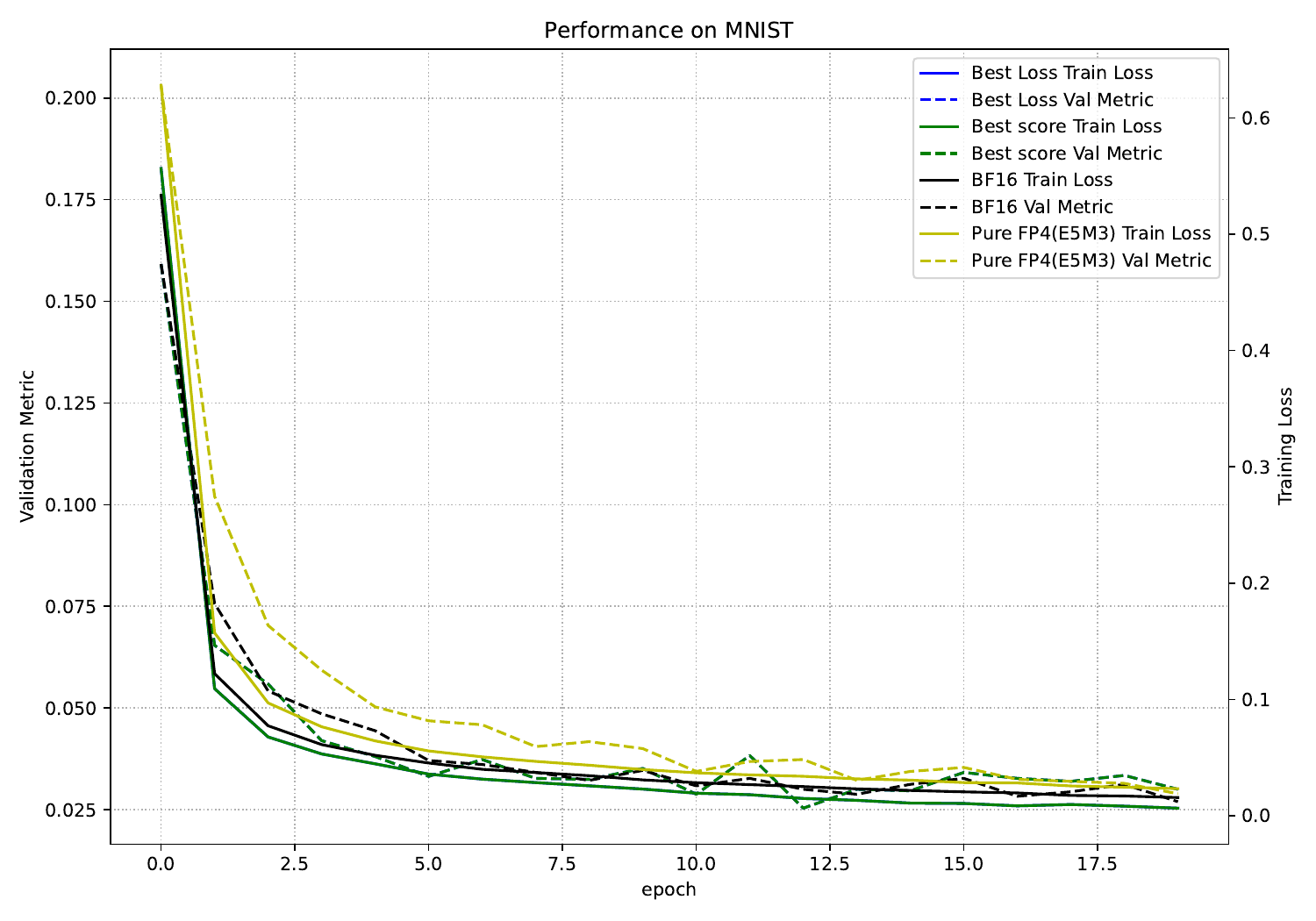}
    \caption{MNIST}
    \label{fig:mnist_ue5m3}
\end{subfigure}\hfill
\begin{subfigure}[b]{0.25\textwidth}
    \centering
    \includegraphics[width=\textwidth]{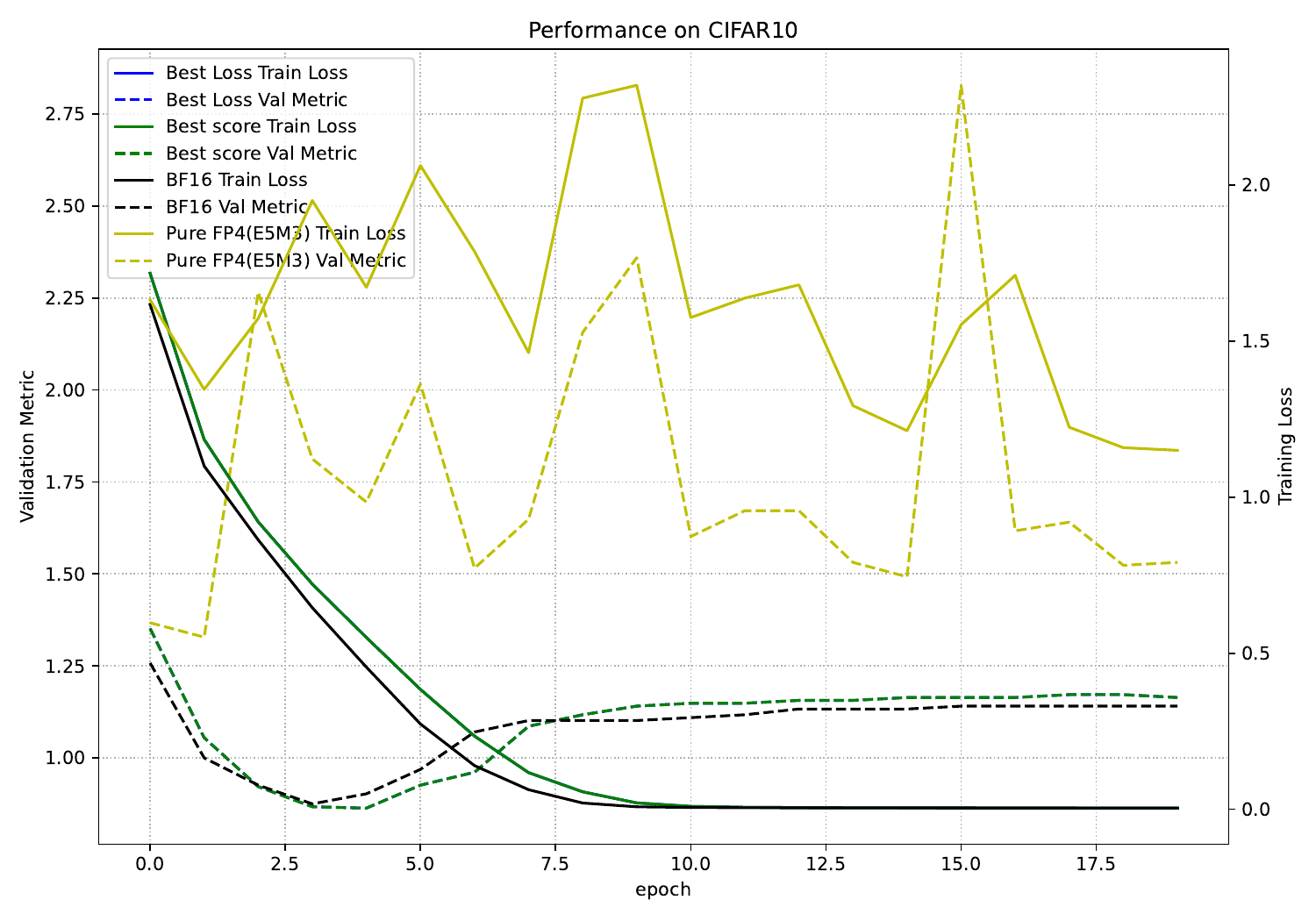}
    \caption{CIFAR-10}
    \label{fig:cifar10_ue5m3}
\end{subfigure}\hfill
\begin{subfigure}[b]{0.25\textwidth}
    \centering
    \includegraphics[width=\textwidth]{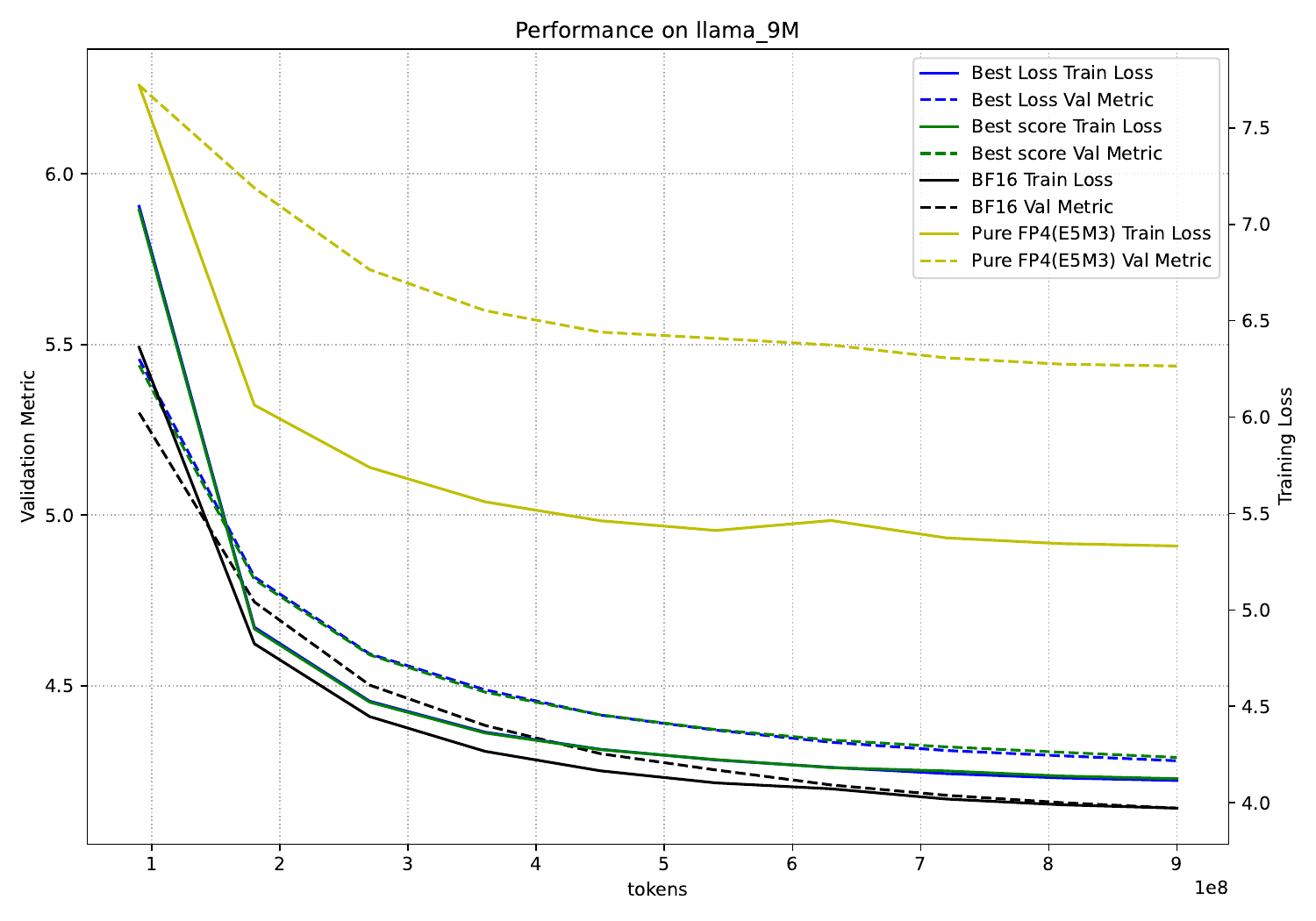}
    \caption{Llama 9M}
    \label{fig:llama_9m_ue5m3}
\end{subfigure}

\end{figure}

\begin{figure}[htp!]
\centering
\caption{Pareto-frontier plots for each dataset, \texttt{UE5M3} results. Note that we swept a smaller space compared to the main experiments.}
\label{fig:ue5m3_pareto}

\begin{subfigure}[b]{0.33\textwidth}
    \centering
    \includegraphics[width=\textwidth]{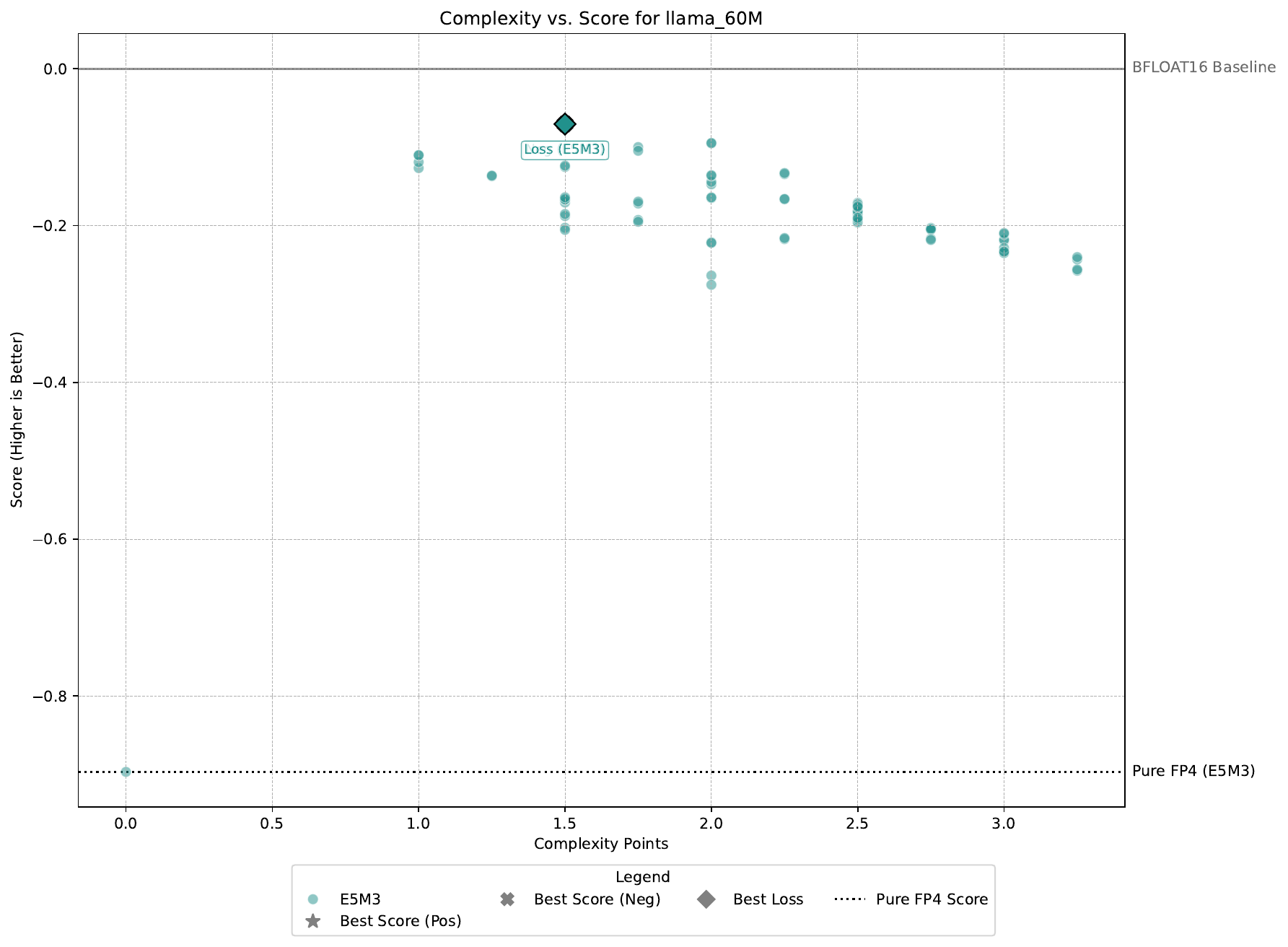}
    \caption{Llama 60M}
    \label{fig:llama_60m}
\end{subfigure}%
\begin{subfigure}[b]{0.33\textwidth}
    \centering
    \includegraphics[width=\textwidth]{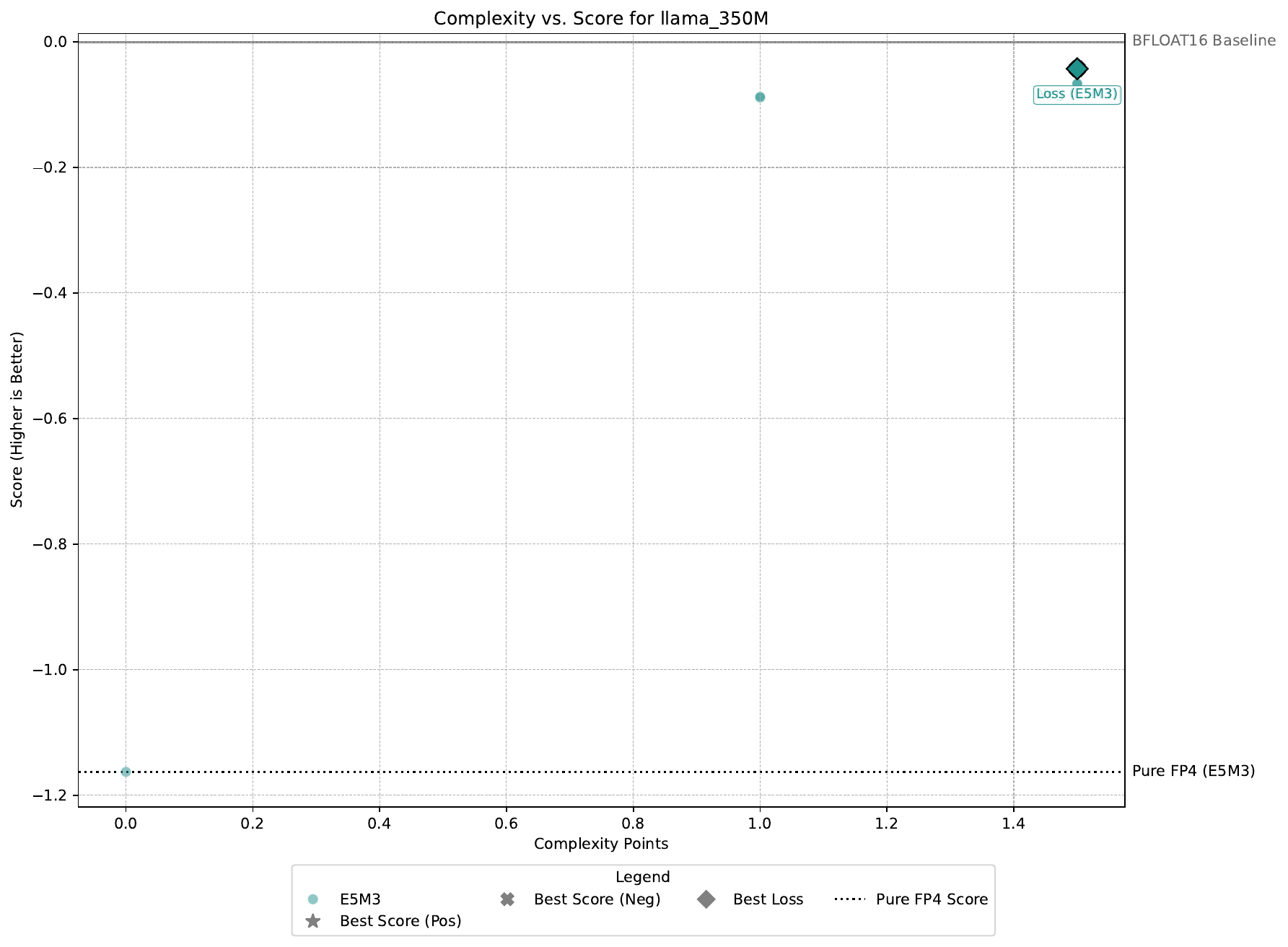}
    \caption{Llama 350M}
    \label{fig:llama_350m}
\end{subfigure}%
\begin{subfigure}[b]{0.33\textwidth}
    \centering
    \includegraphics[width=\textwidth]{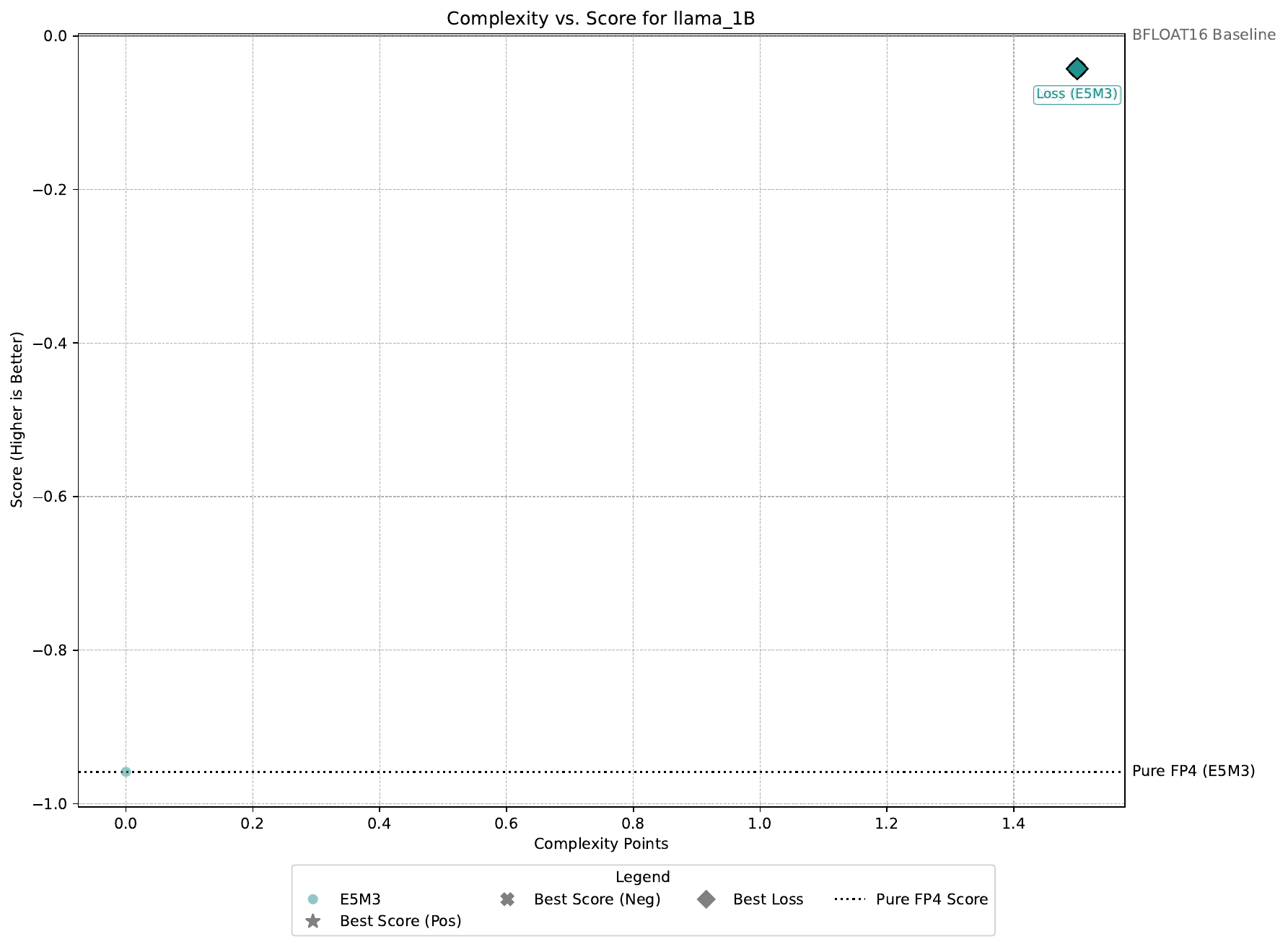}
    \caption{Llama 1B}
    \label{fig:llama_1b}
\end{subfigure}

\begin{subfigure}[b]{0.33\textwidth}
    \centering
    \includegraphics[width=\textwidth]{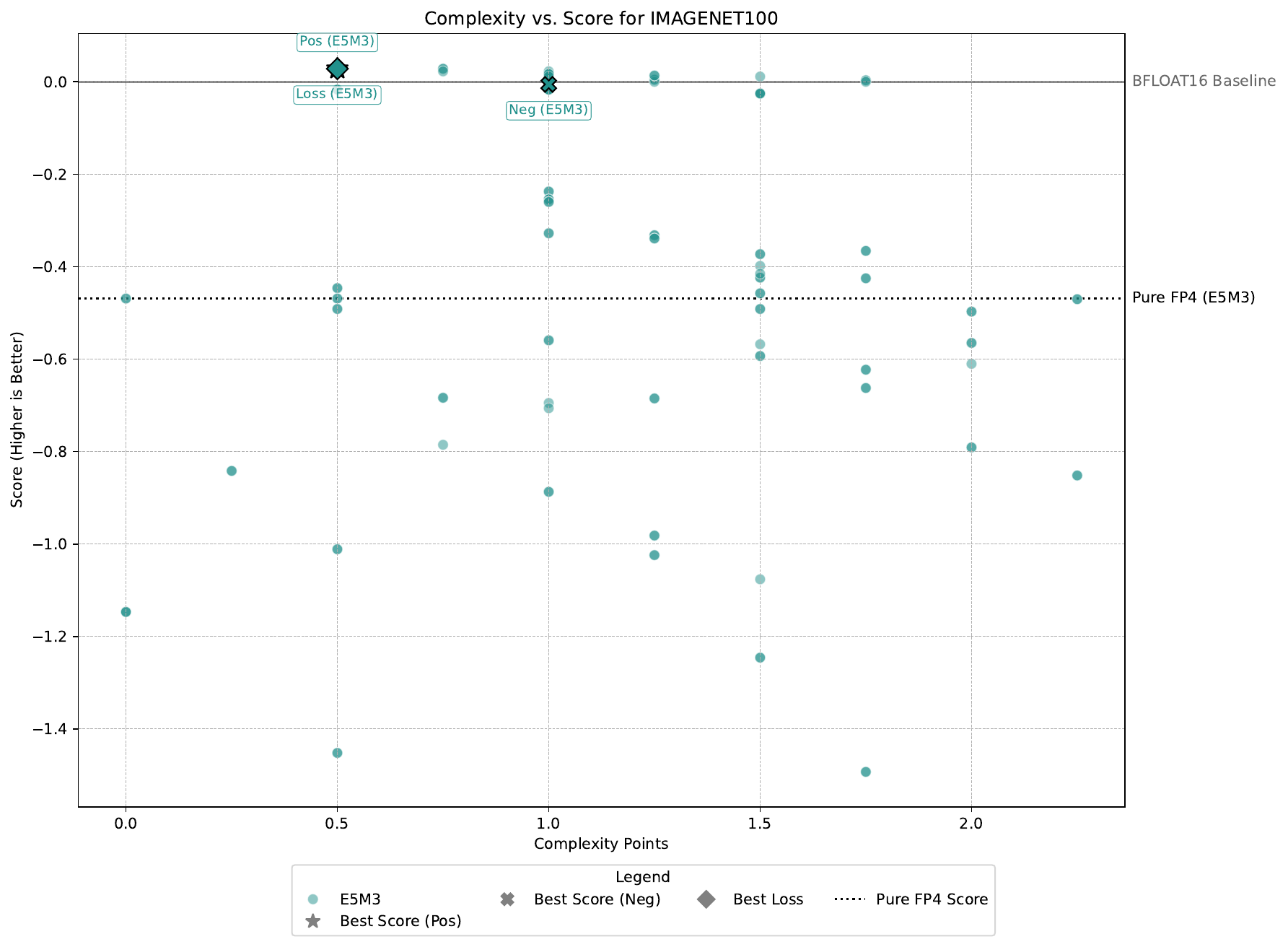}
    \caption{ImageNet-100}
    \label{fig:imagenet100}
\end{subfigure}%
\begin{subfigure}[b]{0.33\textwidth}
    \centering
    \includegraphics[width=\textwidth]{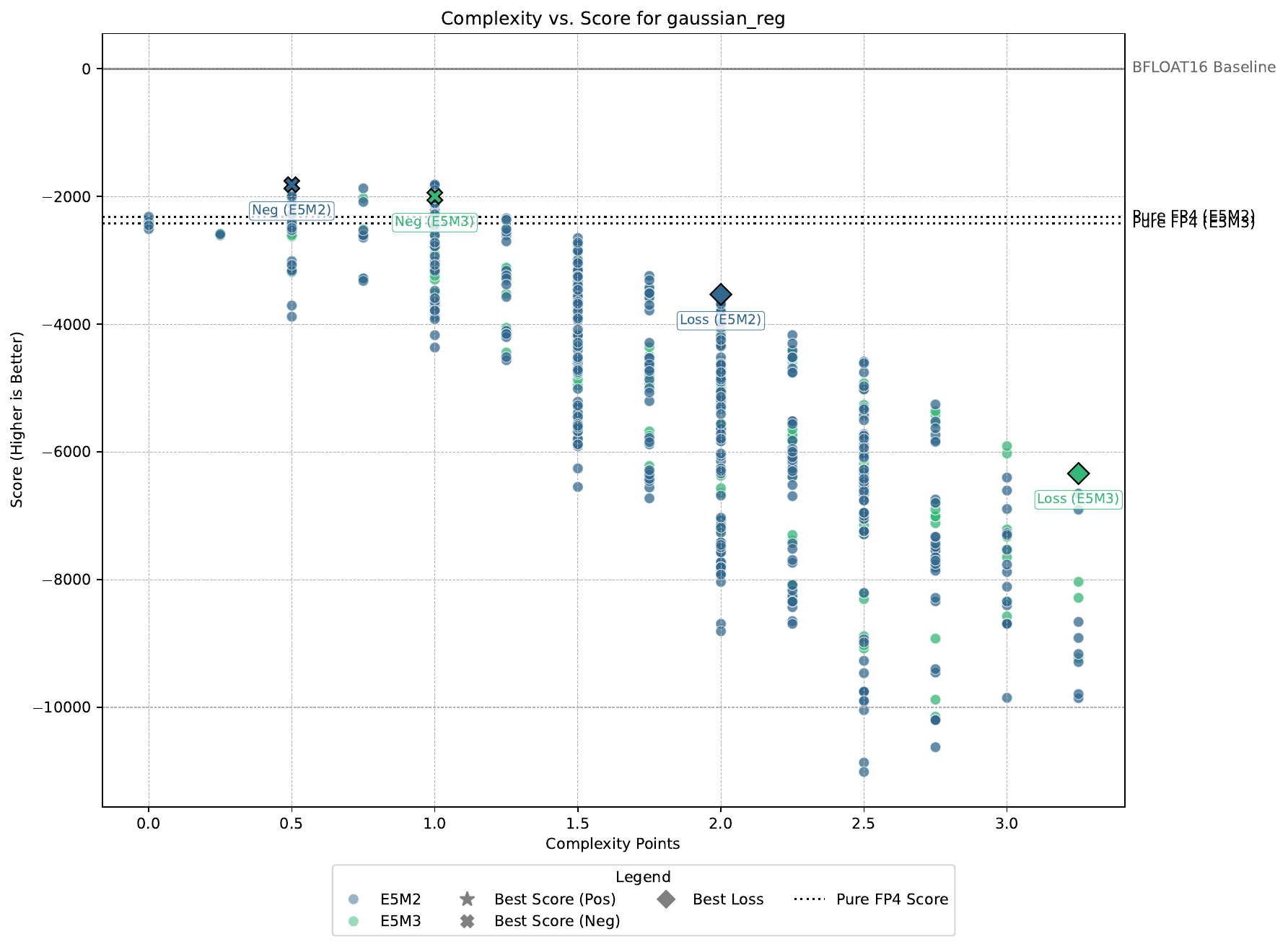}
    \caption{Gaussian Reg.}
    \label{fig:gaussian_reg}
\end{subfigure}%
\begin{subfigure}[b]{0.33\textwidth}
    \centering
    \includegraphics[width=\textwidth]{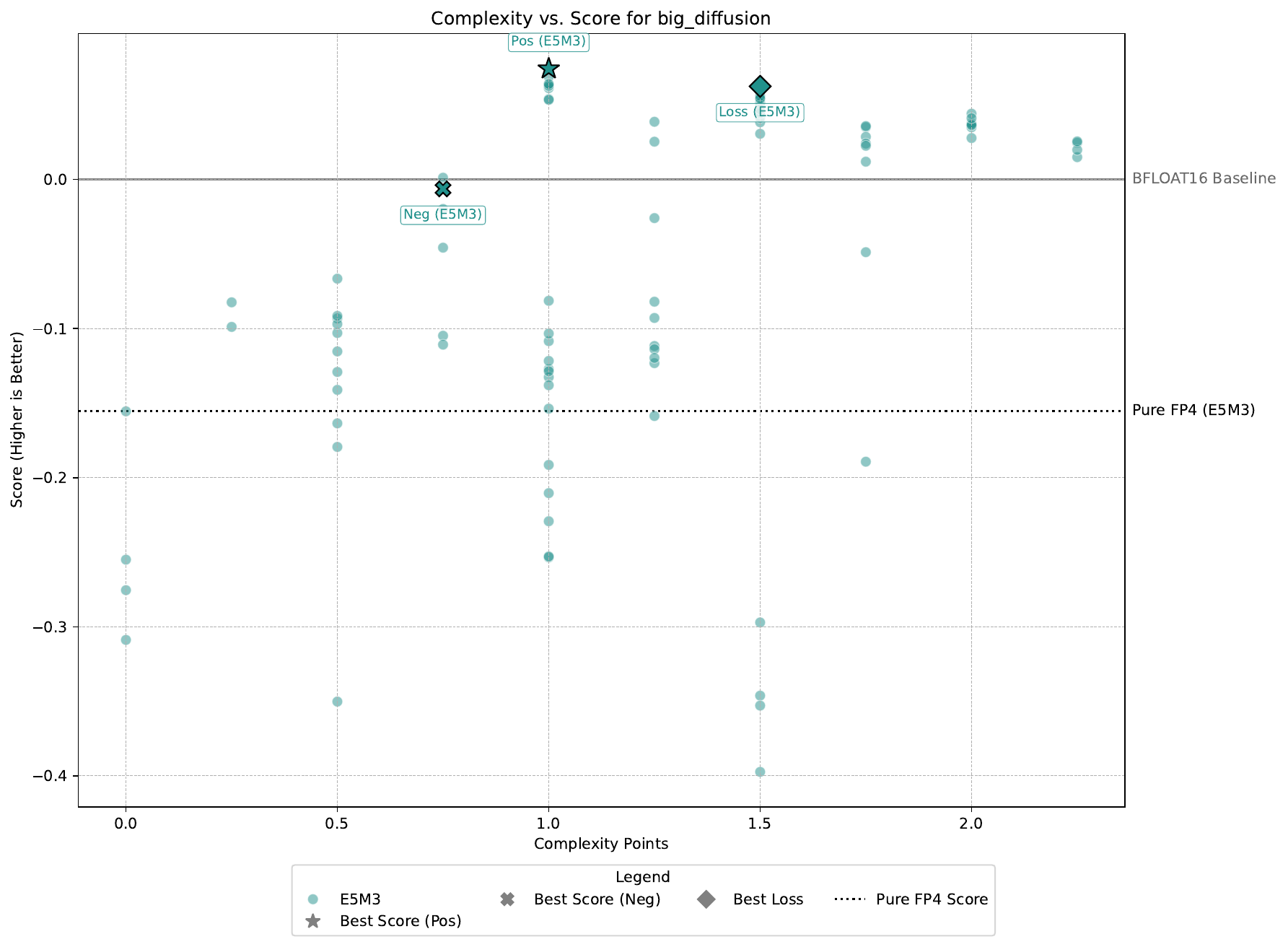}
    \caption{Big Diffusion}
    \label{fig:big_diffusion}
\end{subfigure}

\begin{subfigure}[b]{0.25\textwidth}
    \centering
    \includegraphics[width=\textwidth]{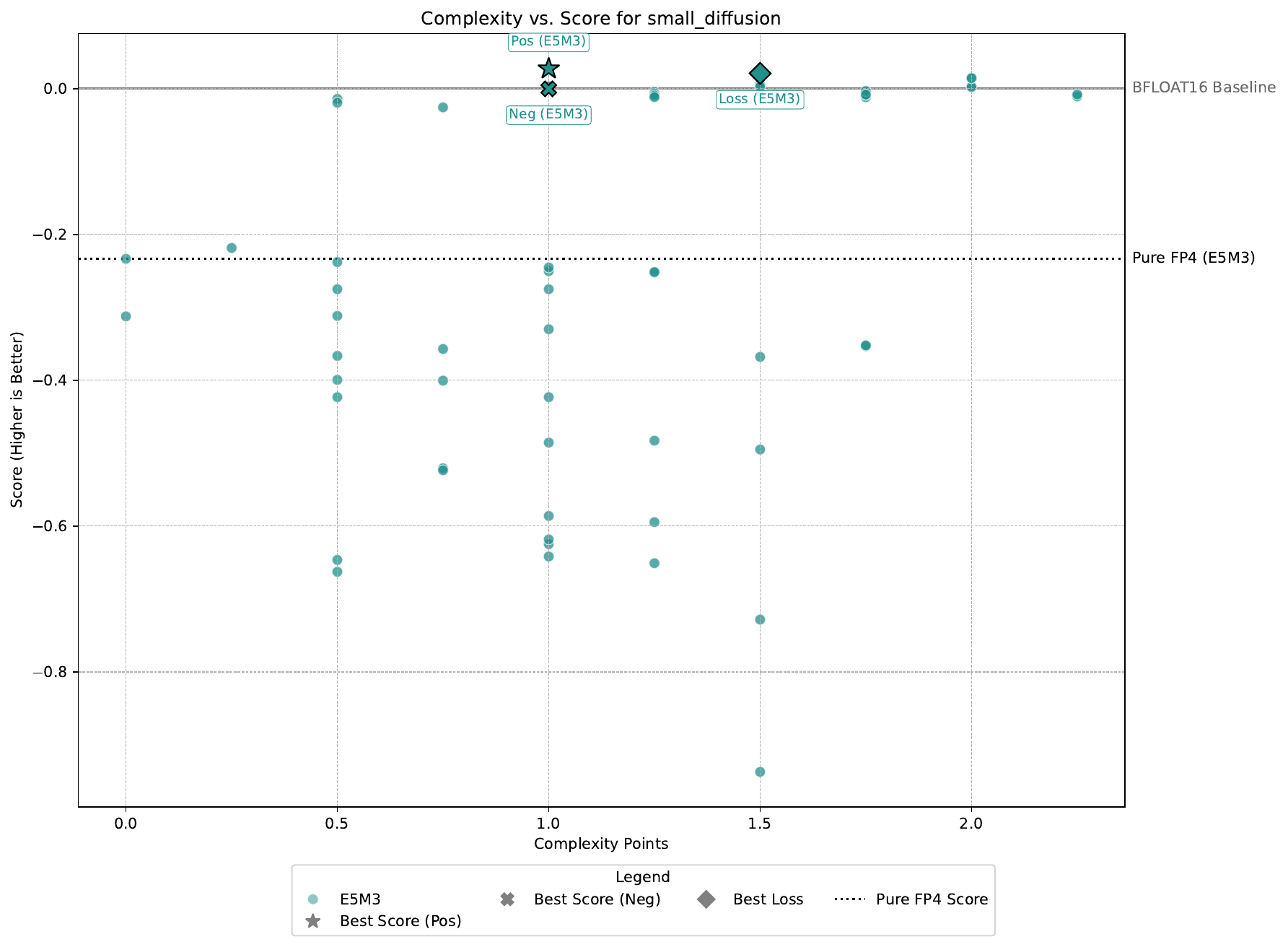}
    \caption{Small Diffusion}
    \label{fig:small_diffusion}
\end{subfigure}\hfill
\begin{subfigure}[b]{0.25\textwidth}
    \centering
    \includegraphics[width=\textwidth]{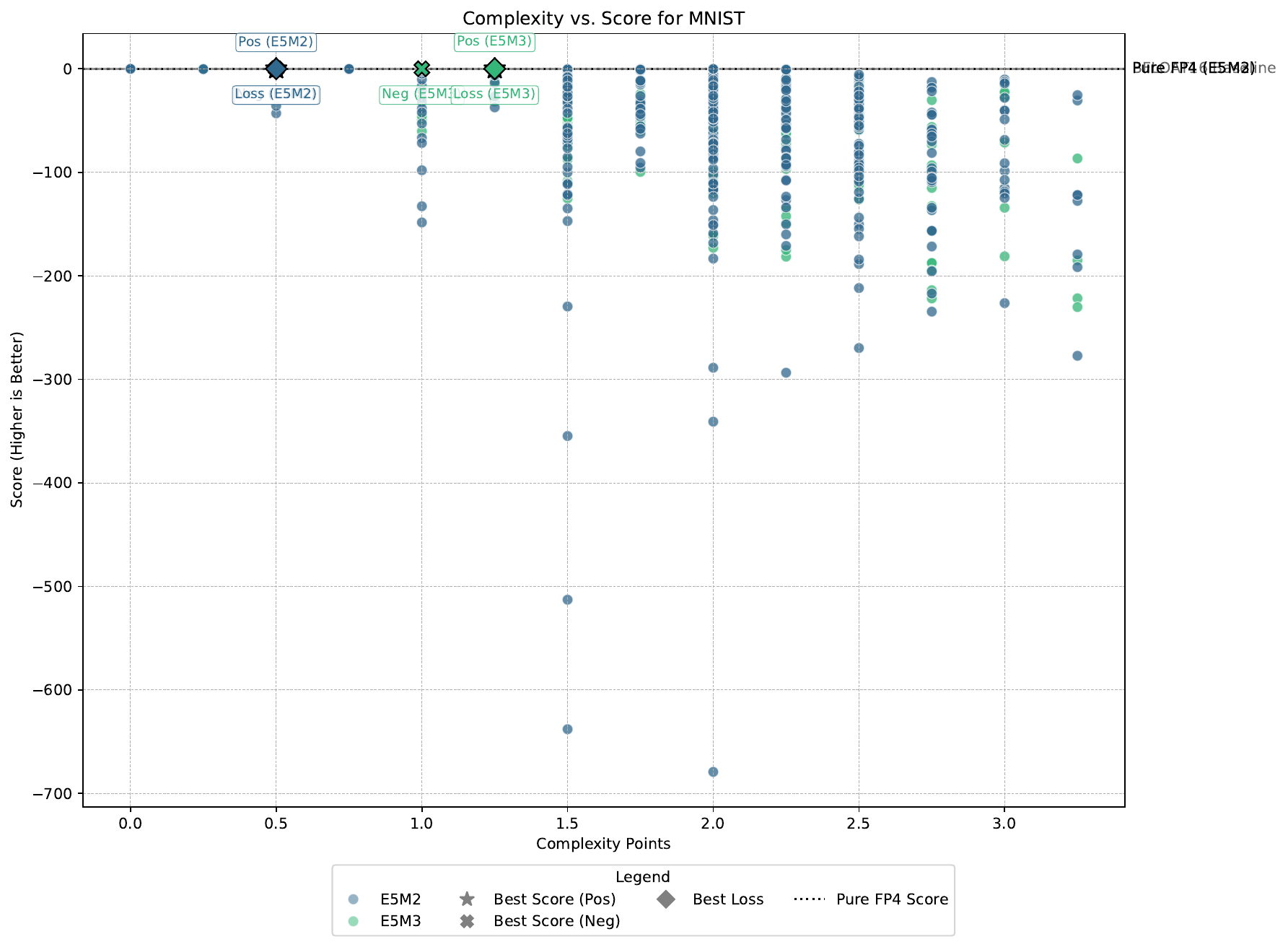}
    \caption{MNIST}
    \label{fig:mnist}
\end{subfigure}\hfill
\begin{subfigure}[b]{0.25\textwidth}
    \centering
    \includegraphics[width=\textwidth]{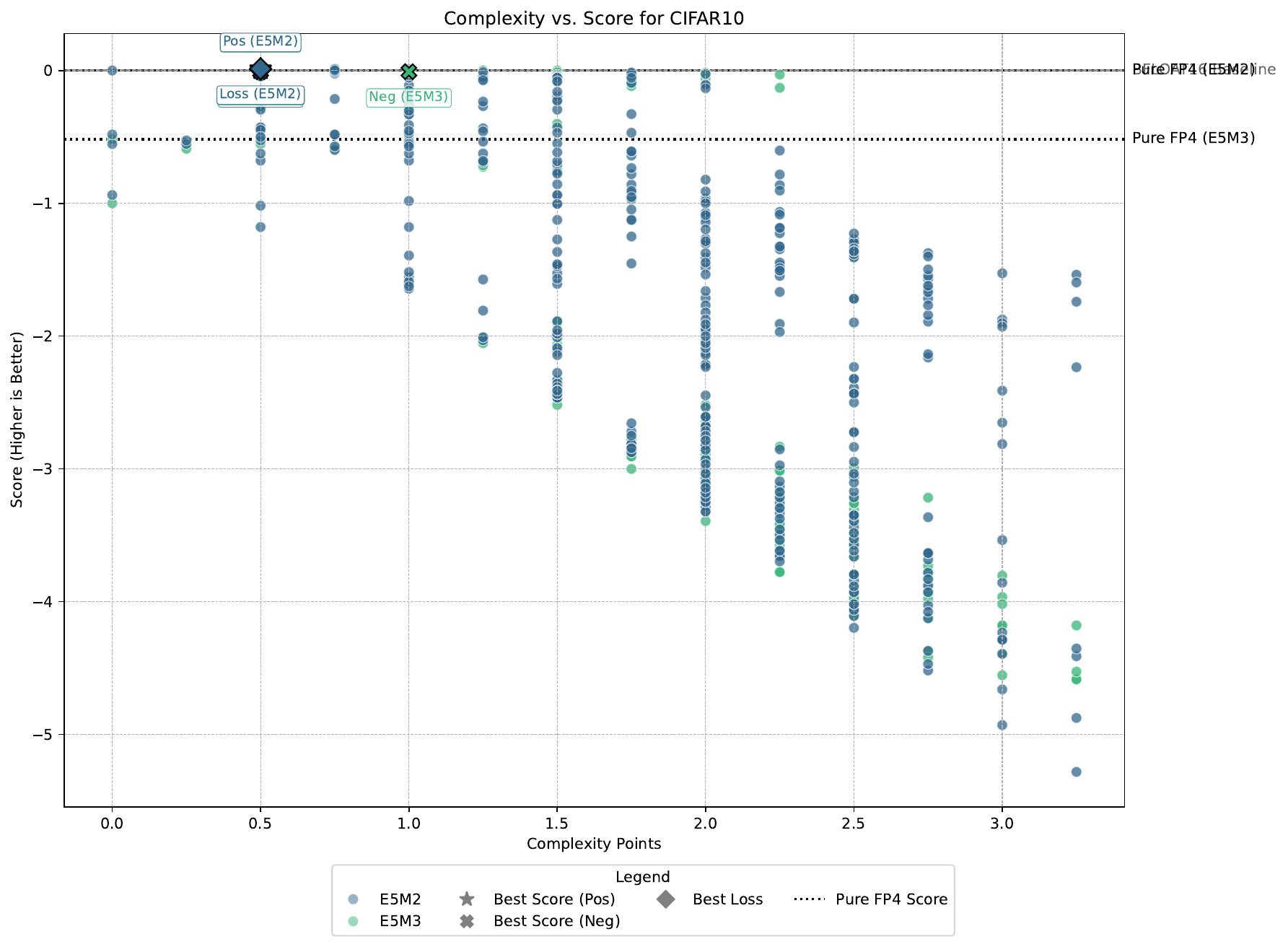}
    \caption{CIFAR-10}
    \label{fig:cifar10}
\end{subfigure}\hfill
\begin{subfigure}[b]{0.25\textwidth}
    \centering
    \includegraphics[width=\textwidth]{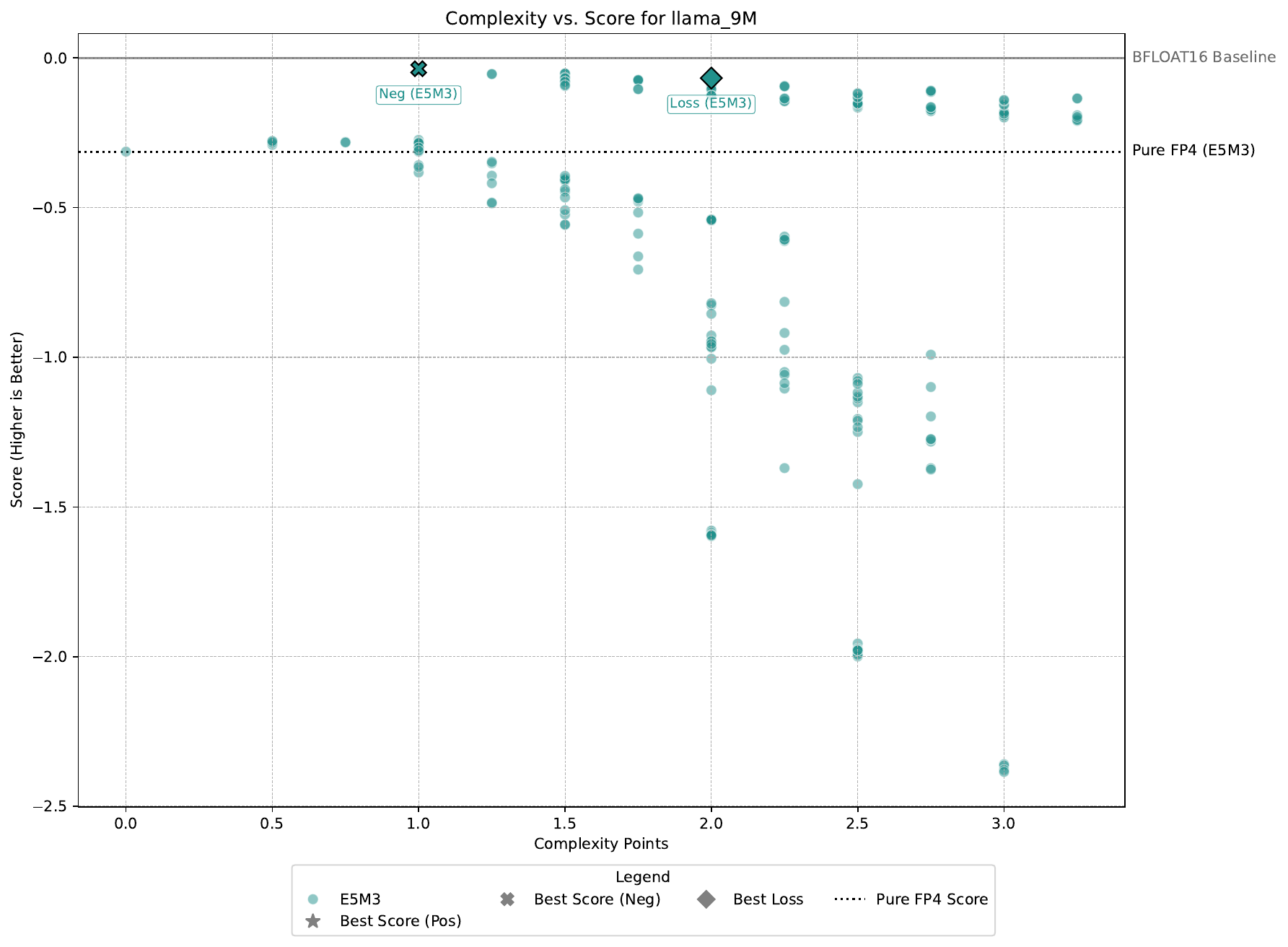}
    \caption{Llama 9M}
    \label{fig:llama_9m}
\end{subfigure}

\end{figure}

We present the \texttt{UE5M3} experiments in \Cref{tab:UE5M3}. We overall find that the \texttt{UE5M3} outperforms \texttt{MXFP4} when tensor scaling is applied. It should be noted that \texttt{UE5M3} will not work without any adjustments unlike \texttt{MXFP4}, implying that increased precision often comes with increased overhead. We visualise the training and validation curves in \Cref{fig:UE5M3_plots} and \Cref{fig:other_models_ue5m3}. We further provide the Pareto-frontier plots in \Cref{fig:ue5m3_pareto}, we note that generally lower complexity configurations achieve better scores.
\begin{table}[htbp!]
\caption{UE5M3 results}
\label{tab:UE5M3}
\resizebox{\textwidth}{!}{%
\begin{tabular}{llrrllllllllrlllrrl}
\toprule
Dataset & Source & \makecell{Val\\loss} & \makecell{Train\\loss} & Scale & \makecell{Block \\ size} & \makecell{Max \\ grad.} & \makecell{Quant.\\grad} & Hadamard & \makecell{Scale\\grad} & SR & Optimiser & \makecell{Loss\\scaling} & Round mode & \makecell{Tensor\\scaling} & \makecell{Tensor \\grad} & \makecell{Complexity\\points} & Score & \makecell{NaN\\mode} \\
\midrule
CIFAR10 & Baseline & 0.875 & 0.003 & N/A & N/A & N/A & N/A & N/A & N/A & N/A & Adam & False & N/A & N/A & N/A & N/A & N/A & N/A \\
CIFAR10 & Baseline & 0.895 & 0.027 & N/A & N/A & N/A & N/A & N/A & N/A & N/A & StableSPAM & False & N/A & N/A & N/A & N/A & N/A & N/A \\
CIFAR10 & Best Score (Neg) & 0.883 & 0.003 & E5M3 & 32 & STE & STE & None\_exact & STE & IntelFP4\_exact & Adam & True & TiesToEven & False & N/A & 1.000 & -0.009 & nearest\_subnormal \\
CIFAR10 & Best Score (Pos) & 0.867 & 0.003 & E5M3 & 32 & STE & STE & None\_exact & STE & IntelFP4\_exact & Adam & False & TiesToEven & False & N/A & 0.500 & 0.009 & nearest\_subnormal \\
CIFAR10 & Best loss E5M3 & 0.867 & 0.003 & E5M3 & 32 & STE & STE & None\_exact & STE & IntelFP4\_exact & Adam & False & TiesToEven & False & N/A & 0.500 & 0.009 & nearest\_subnormal \\
CIFAR10 & Pure FP4 & 0.875 & 0.005 & E5M2 & 32 & STE & STE & None\_exact & STE & None\_exact & Adam & False & TowardPositive & False & N/A & 0.000 & 0.000 & nearest\_subnormal \\
CIFAR10 & Pure FP4 & 1.328 & 1.150 & E5M3 & 32 & STE & STE & None\_exact & STE & None\_exact & Adam & False & TiesToEven & False & N/A & 0.000 & -0.518 & to\_one \\
IMAGENET100 & Baseline & 1.383 & 0.014 & N/A & N/A & N/A & N/A & N/A & N/A & N/A & Adam & False & N/A & N/A & N/A & N/A & N/A & N/A \\
IMAGENET100 & Baseline & 1.750 & 0.078 & N/A & N/A & N/A & N/A & N/A & N/A & N/A & StableSPAM & False & N/A & N/A & N/A & N/A & N/A & N/A \\
IMAGENET100 & Best Score (Neg) & 1.391 & 0.015 & E5M3 & 32 & STE & STE & None\_exact & STE & all\_activation\_exact & Adam & True & TowardPositive & False & N/A & 1.000 & -0.006 & nearest\_subnormal \\
IMAGENET100 & Best Score (Pos) & 1.344 & 0.014 & E5M3 & 32 & STE & STE & None\_exact & STE & IntelFP4\_exact & Adam & False & TiesToEven & False & N/A & 0.500 & 0.028 & nearest\_subnormal \\
IMAGENET100 & Best loss E5M3 & 1.344 & 0.014 & E5M3 & 32 & STE & STE & None\_exact & STE & IntelFP4\_exact & Adam & False & TiesToEven & False & N/A & 0.500 & 0.028 & to\_one \\
IMAGENET100 & Pure FP4 & 2.031 & 1.530 & E5M3 & 32 & STE & STE & None\_exact & STE & None\_exact & Adam & False & TowardPositive & False & N/A & 0.000 & -0.469 & nearest\_subnormal \\
MNIST & Baseline & 0.027 & 0.016 & N/A & N/A & N/A & N/A & N/A & N/A & N/A & Adam & False & N/A & N/A & N/A & N/A & N/A & N/A \\
MNIST & Baseline & 0.028 & 0.004 & N/A & N/A & N/A & N/A & N/A & N/A & N/A & StableSPAM & False & N/A & N/A & N/A & N/A & N/A & N/A \\
MNIST & Best Score (Neg) & 0.027 & 0.010 & E5M3 & 32 & STE & STE & None\_exact & STE & None\_exact & StableSPAM & False & TiesToEven & True & ignore & 1.000 & -0.005 & nearest\_subnormal \\
MNIST & Best Score (Pos) & 0.025 & 0.006 & E5M3 & 32 & STE & STE & None\_exact & STE & IntelFP4\_exact & StableSPAM & False & Stochastic & False & N/A & 1.250 & 0.047 & nearest\_subnormal \\
MNIST & Best loss E5M3 & 0.025 & 0.006 & E5M3 & 32 & STE & STE & None\_exact & STE & IntelFP4\_exact & StableSPAM & False & Stochastic & False & N/A & 1.250 & 0.047 & nearest\_subnormal \\
MNIST & Pure FP4 & 0.029 & 0.022 & E5M2 & 16 & STE & STE & None\_exact & STE & None\_exact & Adam & False & TowardPositive & False & N/A & 0.000 & -0.059 & nearest\_subnormal \\
MNIST & Pure FP4 & 0.029 & 0.023 & E5M3 & 32 & STE & STE & None\_exact & STE & None\_exact & Adam & False & TowardPositive & False & N/A & 0.000 & -0.072 & nearest\_subnormal \\
big\_diffusion & Baseline & 0.135 & 0.128 & N/A & N/A & N/A & N/A & N/A & N/A & N/A & Adam & False & N/A & N/A & N/A & N/A & N/A & N/A \\
big\_diffusion & Baseline & 0.113 & 0.110 & N/A & N/A & N/A & N/A & N/A & N/A & N/A & StableSPAM & False & N/A & N/A & N/A & N/A & N/A & N/A \\
big\_diffusion & Best Score (Neg) & 0.113 & 0.109 & E5M3 & 32 & STE & STE & None\_exact & STE & None\_exact & Adam & True & Stochastic & False & N/A & 0.750 & -0.006 & nearest\_subnormal \\
big\_diffusion & Best Score (Pos) & 0.104 & 0.100 & E5M3 & 32 & STE & STE & None\_exact & STE & all\_activation\_exact & StableSPAM & False & TowardPositive & False & N/A & 1.000 & 0.074 & to\_one \\
big\_diffusion & Best loss E5M3 & 0.102 & 0.097 & E5M3 & 32 & STE & STE & None\_exact & STE & all\_activation\_exact & StableSPAM & True & TiesToEven & False & N/A & 1.500 & 0.062 & to\_one \\
big\_diffusion & Pure FP4 & 0.130 & 0.123 & E5M3 & 32 & STE & STE & None\_exact & STE & None\_exact & Adam & False & TiesToEven & False & N/A & 0.000 & -0.155 & nearest\_subnormal \\
gaussian\_reg & Baseline & 25.250 & 25.224 & N/A & N/A & N/A & N/A & N/A & N/A & N/A & Adam & False & N/A & N/A & N/A & N/A & N/A & N/A \\
gaussian\_reg & Baseline & 0.013 & 0.013 & N/A & N/A & N/A & N/A & N/A & N/A & N/A & StableSPAM & False & N/A & N/A & N/A & N/A & N/A & N/A \\
gaussian\_reg & Best Score (Neg) & 25.875 & 26.250 & E5M3 & 32 & STE & STE & None\_exact & STE & None\_exact & StableSPAM & False & TiesToEven & True & ignore & 1.000 & -1998.698 & nearest\_subnormal \\
gaussian\_reg & Best loss E5M3 & 25.250 & 26.237 & E5M3 & 32 & STE & STE & backward\_exact & STE & IntelFP4\_exact & StableSPAM & True & Stochastic & True & ignore & 3.250 & -6338.788 & nearest\_subnormal \\
gaussian\_reg & Pure FP4 & 30.000 & 29.974 & E5M2 & 16 & STE & STE & None\_exact & STE & None\_exact & Adam & False & TiesToEven & False & N/A & 0.000 & -2317.491 & nearest\_subnormal \\
gaussian\_reg & Pure FP4 & 31.375 & 31.836 & E5M3 & 32 & STE & STE & None\_exact & STE & None\_exact & Adam & False & TiesToEven & False & N/A & 0.000 & -2423.755 & nearest\_subnormal \\
llama\_1B & Baseline & 3.578 & 3.682 & N/A & N/A & N/A & N/A & N/A & N/A & N/A & Adam & False & N/A & N/A & N/A & N/A & N/A & N/A \\
llama\_1B & Baseline & 3.487 & 3.569 & N/A & N/A & N/A & N/A & N/A & N/A & N/A & StableSPAM & False & N/A & N/A & N/A & N/A & N/A & N/A \\
llama\_1B & Best Score (Neg) & 3.586 & 3.666 & E5M3 & 32 & STE & STE & None\_exact & STE & IntelFP4\_exact & StableSPAM & False & TiesToEven & True & ignore & 1.500 & -0.043 & to\_one \\
llama\_1B & Best loss E5M3 & 3.586 & 3.666 & E5M3 & 32 & STE & STE & None\_exact & STE & IntelFP4\_exact & StableSPAM & False & TiesToEven & True & ignore & 1.500 & -0.043 & to\_one \\
llama\_1B & Pure FP4 & 6.830 & 6.802 & E5M3 & 32 & STE & STE & None\_exact & STE & None\_exact & Adam & False & TiesToEven & False & N/A & 0.000 & -0.959 & to\_one \\
llama\_350M & Baseline & 2.269 & 2.375 & N/A & N/A & N/A & N/A & N/A & N/A & N/A & Adam & False & N/A & N/A & N/A & N/A & N/A & N/A \\
llama\_350M & Baseline & 2.258 & 2.363 & N/A & N/A & N/A & N/A & N/A & N/A & N/A & StableSPAM & False & N/A & N/A & N/A & N/A & N/A & N/A \\
llama\_350M & Best Score (Neg) & 2.322 & 2.437 & E5M3 & 32 & STE & STE & None\_exact & STE & IntelFP4\_exact & StableSPAM & False & TiesToEven & True & ignore & 1.500 & -0.043 & to\_one \\
llama\_350M & Best loss E5M3 & 2.322 & 2.437 & E5M3 & 32 & STE & STE & None\_exact & STE & IntelFP4\_exact & StableSPAM & False & TiesToEven & True & ignore & 1.500 & -0.043 & to\_one \\
llama\_350M & Pure FP4 & 4.884 & 4.963 & E5M3 & 32 & STE & STE & None\_exact & STE & None\_exact & Adam & False & TiesToEven & False & N/A & 0.000 & -1.163 & to\_one \\
llama\_60M & Baseline & 2.665 & 2.657 & N/A & N/A & N/A & N/A & N/A & N/A & N/A & Adam & False & N/A & N/A & N/A & N/A & N/A & N/A \\
llama\_60M & Baseline & 2.983 & 3.028 & N/A & N/A & N/A & N/A & N/A & N/A & N/A & StableSPAM & False & N/A & N/A & N/A & N/A & N/A & N/A \\
llama\_60M & Best Score (Neg) & 2.791 & 2.788 & E5M3 & 32 & STE & STE & None\_exact & STE & IntelFP4\_exact & StableSPAM & False & TiesToEven & True & ignore & 1.500 & -0.071 & to\_one \\
llama\_60M & Best loss E5M3 & 2.791 & 2.788 & E5M3 & 32 & STE & STE & None\_exact & STE & IntelFP4\_exact & StableSPAM & False & TiesToEven & True & ignore & 1.500 & -0.071 & to\_one \\
llama\_60M & Pure FP4 & 5.056 & 5.050 & E5M3 & 32 & STE & STE & None\_exact & STE & None\_exact & Adam & False & TiesToEven & False & N/A & 0.000 & -0.897 & to\_one \\
llama\_9M & Baseline & 4.183 & 4.013 & N/A & N/A & N/A & N/A & N/A & N/A & N/A & Adam & False & N/A & N/A & N/A & N/A & N/A & N/A \\
llama\_9M & Baseline & 4.141 & 3.972 & N/A & N/A & N/A & N/A & N/A & N/A & N/A & StableSPAM & False & N/A & N/A & N/A & N/A & N/A & N/A \\
llama\_9M & Best Score (Neg) & 4.290 & 4.125 & E5M3 & 32 & STE & STE & None\_exact & STE & None\_exact & StableSPAM & False & TiesToEven & True & ignore & 1.000 & -0.036 & to\_one \\
llama\_9M & Best loss E5M3 & 4.280 & 4.115 & E5M3 & 32 & STE & STE & None\_exact & STE & IntelFP4\_exact & StableSPAM & True & TiesToEven & True & ignore & 2.000 & -0.067 & to\_one \\
llama\_9M & Pure FP4 & 5.437 & 5.332 & E5M3 & 32 & STE & STE & None\_exact & STE & None\_exact & Adam & False & TiesToEven & False & N/A & 0.000 & -0.313 & to\_one \\
small\_diffusion & Baseline & 0.029 & 0.029 & N/A & N/A & N/A & N/A & N/A & N/A & N/A & Adam & False & N/A & N/A & N/A & N/A & N/A & N/A \\
small\_diffusion & Baseline & 0.019 & 0.019 & N/A & N/A & N/A & N/A & N/A & N/A & N/A & StableSPAM & False & N/A & N/A & N/A & N/A & N/A & N/A \\
small\_diffusion & Best Score (Neg) & 0.019 & 0.019 & E5M3 & 32 & STE & STE & None\_exact & STE & None\_exact & StableSPAM & False & TiesToEven & True & ignore & 1.000 & -0.000 & to\_one \\
small\_diffusion & Best Score (Pos) & 0.018 & 0.019 & E5M3 & 32 & STE & STE & None\_exact & STE & IntelFP4\_exact & StableSPAM & False & TiesToEven & False & N/A & 1.000 & 0.027 & to\_one \\
small\_diffusion & Best loss E5M3 & 0.018 & 0.019 & E5M3 & 32 & STE & STE & None\_exact & STE & IntelFP4\_exact & StableSPAM & True & TiesToEven & False & N/A & 1.500 & 0.021 & nearest\_subnormal \\
small\_diffusion & Pure FP4 & 0.023 & 0.024 & E5M3 & 32 & STE & STE & None\_exact & STE & None\_exact & Adam & False & TiesToEven & False & N/A & 0.000 & -0.233 & to\_one \\
\bottomrule
\end{tabular}
}
\end{table}

\clearpage

\subsection{Experimentation details}

\paragraph{Complexity Score Calculation}
\label{par:appendix_complexity}

The complexity penalty $\Omega(c)$ is calculated based on the set of techniques $\mathcal{T}$ used in a configuration. The total set of techniques and their corresponding weights $w_t$ are detailed in Table~\ref{tab:complexity_weights}. A configuration's total complexity is the sum of weights for all techniques it employs, i.e., $\Omega(c) = \sum_{t \in \mathcal{T}_c} w_t$, where $\mathcal{T}_c \subseteq \mathcal{T}$.

\begin{table}[htpb!]
\centering
\caption{Complexity weights for non-baseline techniques.}
\label{tab:complexity_weights}
\begin{tabular}{l l c}
\toprule
\textbf{Technique ($t$)} & \textbf{Activation Condition} & \textbf{Weight ($w_t$)} \\
\midrule
Non-STE Smoothing & \texttt{smooth} $\neq$ 'STE' & 3.0 \\
Tensor Scaling Gradient Est. & \texttt{tensor\_scaling\_grad\_est} is active & 3.0 \\
\addlinespace 
Non-STE Step Gradient & \texttt{stepGradient} $\neq$ 'STE' & 2.0 \\
Hadamard Transform & \texttt{use\_hadamard} is active & 1.0 \\
\addlinespace
Non-STE Quantized Gradient & \texttt{qGradient} $\neq$ 'STE' & 1.5 \\
\addlinespace
Stochastic Rounding (SR) & \texttt{SR} is active & 0.5 \\
Tensor Scaling & \texttt{use\_tensor\_scaling} is active & 0.5 \\
Loss Scaling & \texttt{loss\_scaling} is True & 0.5 \\
SPAM Optimizer & 'SPAM' in \texttt{optimiser} name & 0.5 \\
Stochastic Rounding for scale & Scale rounding is \texttt{Stochastic}  & 0.25 \\

\bottomrule
\end{tabular}
\end{table}
We motivate the weight with reference to the added complexity and memory overhead, and fusability on a lower level language based on \Cref{tab:techniques_summary}.

\paragraph{Parameter sweeps}

\begin{table}[htbp!]
\centering
\caption{Summary of the Full Hyperparameter Sweep.}
\label{tab:full_sweep_compact}
\footnotesize
\begin{tabular}{@{}llp{0.5\linewidth}@{}}
\toprule
\textbf{Group} & \textbf{Parameter} & \textbf{Search Values} \\
\midrule
Quant. & Scale Format & \{\texttt{E8M0}, \texttt{E4M3}\} \\
       & Max Approx. & \{\texttt{STE}, \texttt{softsoftmax}, \texttt{hardsoftmax}, \texttt{absmax}\} \\
       & Scale Rounding & \{\texttt{TiesToEven}, \texttt{TowardPositive}, \texttt{SR}\} \\
\midrule
Gradient & Step Gradient & \{\texttt{STE}, \texttt{baseline}, \texttt{spline}\} \\
         & Scaling Quant. \textsuperscript{1} & \{\texttt{STE}, \texttt{baseline}, \texttt{spline}\} \\
         & Tensor Scale Grad.\textsuperscript{2} & \{\texttt{ignore}, \texttt{absmax}, \texttt{STE}\} \\
\midrule
Opt. & Optimizer & \{\texttt{Adam}, \texttt{StableSPAM}\} \\
     & Loss Scaling & \{\texttt{True}, \texttt{False}\} \\
     & Tensor Scaling & \{\texttt{True}, \texttt{False}\} \\
     & SR & \{\texttt{None}, \texttt{all act.}, \texttt{backward act.}\} \\
     & Hadamard  & \{\texttt{None}, \texttt{all}, \texttt{backward}\} \\
\bottomrule
\end{tabular}
\par
\parbox{\linewidth}{\footnotesize%
\textsuperscript{1}\textbf{Conditional}: Options for \texttt{Scaling Quant.} depend on the values of \texttt{Max Approx}.
}
\parbox{\linewidth}{\footnotesize%
\textsuperscript{2}\textbf{Conditional}: Options for \texttt{Tensor Scale Grad.} depend on the values of \texttt{Tensor Scaling} and \texttt{Max Approx}. 
}
\end{table}

\paragraph{Dataset descriptions}
\begin{table}[htbp!]
\centering
\caption{Summary of Experimental Setups}
\label{tab:exp_summary}
\small 
\resizebox{\textwidth}{!}{%

\begin{tabular}{@{}llccccc@{}}
\toprule
\textbf{Learning Task} & \textbf{Dataset / Model} & \textbf{Global Batch Size} & \textbf{Seq. Length} & \textbf{Grad. Accum.} & \textbf{Learning Rate} & \textbf{Training Duration} \\ \midrule
Regression & Synthetic Gaussian & 4096 & \multicolumn{2}{c}{-} & $1 \times 10^{-2}$ & 20 Epochs \\
Classification & MNIST & 512 & \multicolumn{2}{c}{-} & $1 \times 10^{-3}$ & 20 Epochs \\
Classification & CIFAR-10 & 512 & \multicolumn{2}{c}{-} & $1 \times 10^{-3}$ & 20 Epochs \\
Classification & ImageNet-100 & 512 & \multicolumn{2}{c}{-} & $1 \times 10^{-3}$ & 20 Epochs \\
Image Generation & CIFAR-10 (Small U-Net, small\_diffusion) & 512 & \multicolumn{2}{c}{-} & $1 \times 10^{-3}$ & 20 Epochs \\
Image Generation & FFHQ (Big U-Net, big\_diffusion) & 20 & \multicolumn{2}{c}{-} & $1 \times 10^{-4}$ & 3 Epochs \\ \midrule
Language Modeling & LLaMA-9M & 4096 & 128 & 1 & $1 \times 10^{-3}$ & 0.9B Tokens\textsuperscript{*} \\
Language Modeling & LLaMA-60M & 128 & 512 & 1 & $1 \times 10^{-4}$ & 6B Tokens\textsuperscript{*} \\
Language Modeling & LLaMA-350M & 16 & 1024 & 8 & $1 \times 10^{-4}$ & 14.7B Tokens\textsuperscript{*} \\
Language Modeling & LLaMA-1B & 4 & 1024 & 512 & $1 \times 10^{-4}$ & 42B Tokens\textsuperscript{*} \\
\bottomrule
\end{tabular}
}
\par
\vspace{0.5em} 
\parbox{\linewidth}{\footnotesize%
\textsuperscript{*}We use the WikiText dataset. Training duration is calculated based on parameter count (100x for models $<$350M, $\approx 42\times$ otherwise). We chose the token count based on \cite{fishman2025scalingfp8trainingtrilliontoken}, which show that training divergence in low precision usually happen around this amount of tokens relative to model size. Gradient accumulation is used for the 350M and 1B models. The 350M and 1B model configurations where taken directly from \cite{tseng2025trainingllmsmxfp4}.
}
\end{table}

\clearpage

\subsection{Derivation of Proposition 1}
\label{derivation_1}
The function for a single element is:
\[
f_{ij} = \frac{1}{s_q} Q(s_q \mathbf{X}_{ij})
\]
Since $s_q$ is a function of $\mathbf{X}_{ij}$, we must use the product rule on: $\left(\frac{1}{s_q}\right)$ and $\left(Q(s_q\cdot \mathbf{X}_{ij})\right)$.
\[
\frac{\partial f_{ij}}{\partial \mathbf{X}_{ij}} = \left( \frac{\partial}{\partial \mathbf{X}_{ij}} \frac{1}{s_q} \right) \cdot Q(s_q \mathbf{X}_{ij}) + \frac{1}{s_q} \cdot \left( \frac{\partial}{\partial \mathbf{X}_{ij}} Q(s_q \mathbf{X}_{ij}) \right)
\]

\paragraph{}{Term 1: Derivative of $\frac{1}{s_q}$}
This derivative depends on how $s_q$ is defined by $\mathbf{X}$. Let $s = s(\mathbf{X})$.
\[
\frac{\partial}{\partial \mathbf{X}_{ij}}\left(\frac{1}{s_q}\right) = -\frac{1}{s_q^2}\frac{\partial s_q}{\partial \mathbf{X}_{ij}} = -\frac{q'(s)}{s_q^2} \frac{\partial s}{\partial \mathbf{X}_{ij}}
\]

\paragraph{}{Term 2: Derivative of $Q(s_q \mathbf{X}_{ij})$}
 We apply the chain rule to $Q$, and then the product rule to its argument $(s_q \mathbf{X}_{ij})$.
\begin{align*}
    \frac{\partial}{\partial \mathbf{X}_{ij}} Q(s_q \mathbf{X}_{ij}) &= Q'(s_q \mathbf{X}_{ij}) \cdot \frac{\partial (s_q \mathbf{X}_{ij})}{\partial \mathbf{X}_{ij}} \\
    \frac{\partial (s_q \mathbf{X}_{ij})}{\partial \mathbf{X}_{ij}} &= \left(\frac{\partial s_q}{\partial \mathbf{X}_{ij}}\right) \mathbf{X}_{ij} + s_q \left(\frac{\partial \mathbf{X}_{ij}}{\partial \mathbf{X}_{ij}}\right) = \mathbf{X}_{ij} \frac{\partial s_q}{\partial \mathbf{X}_{ij}} + s_q = \mathbf{X}_{ij} q'(s) \frac{\partial s}{\partial \mathbf{X}_{ij}} + s_q
\end{align*}
Thus, the full derivative of the second term is:
\[
\frac{\partial}{\partial \mathbf{X}_{ij}} Q(s_q \mathbf{X}_{ij}) = Q'(s_q \mathbf{X}_{ij}) \left( \mathbf{X}_{ij} q'(s) \frac{\partial s}{\partial \mathbf{X}_{ij}} + s_q\right)
\]

\paragraph{Combining and Final Result}
We substitute the results for both terms back into the main equation:
\[
\frac{\partial f_{ij}}{\partial \mathbf{X}_{ij}} = \left( -\frac{q'(s)}{s_q^2} \frac{\partial s}{\partial \mathbf{X}_{ij}} \right) Q(s_q \mathbf{X}_{ij}) + \frac{1}{s_q} \left[ Q'(s_q \mathbf{X}_{ij}) \left( \mathbf{X}_{ij} q'(s) \frac{\partial s}{\partial \mathbf{X}_{ij}} + s_q\right) \right]
\]
Distributing the $\frac{1}{s_q}$ term:
\[
\frac{\partial f_{ij}}{\partial \mathbf{X}_{ij}} = -\frac{q'(s)}{s_q^2} Q(s_q \mathbf{X}_{ij}) \frac{\partial s}{\partial \mathbf{X}_{ij}} + \frac{\mathbf{X}_{ij}}{s_q} Q'(s_q \mathbf{X}_{ij}) q'(s) \frac{\partial s}{\partial \mathbf{X}_{ij}} + \frac{s_q}{s_q} Q'(s_q \mathbf{X}_{ij})
\]
Grouping the terms by their derivative component gives the final result for this fully general model:
\begin{equation}
\boxed{
\frac{\partial f_{ij}}{\partial \mathbf{X}_{ij}} = Q'(s_q \mathbf{X}_{ij}) + \frac{\partial s}{\partial \mathbf{X}_{ij}} \left[ \frac{q'(s)}{s_q} \left( \mathbf{X}_{ij} Q'(s_q \mathbf{X}_{ij}) - \frac{1}{s_q} Q(s_q \mathbf{X}_{ij}) \right) \right]
}
\end{equation}

\subsection{Theorem 2 derivation}
\label{derivation_2}

\begin{proof}
We want to find the partial derivative of $h_{ij}(\mathbf{X})$ with respect to an element $\mathbf{X}_{ij}$. The transformation is defined as:
\[
h_{ij}(\mathbf{X}) = g(\mathbf{X}) \cdot f_{ij}(\mathbf{U}_p)
\]
where $g(\mathbf{X}) = \text{absmax}(\mathbf{X})$ and $\mathbf{U}_p = \mathbf{X}_p / g(\mathbf{X})$. An element $\mathbf{X}_{ij}$ belongs to a specific block $p$.

\begin{enumerate}
    \item \textbf{Apply the Product Rule.} We treat $g$ and $f_{ij}$ as two functions of $\mathbf{X}$. The product rule states $(uv)' = u'v + uv'$.
    \[
    \frac{\partial h_{ij}}{\partial \mathbf{X}_{ij}} = \frac{\partial g}{\partial \mathbf{X}_{ij}} \cdot f_{ij}(\mathbf{U}_p) + g \cdot \frac{\partial f_{ij}(\mathbf{U}_p)}{\partial \mathbf{X}_{ij}}
    \]

    \item \textbf{Apply the Chain Rule.} The second term's derivative requires the chain rule because $f_{ij}$ is a function of $\mathbf{U}_{p,ij}$, which is a function of $\mathbf{X}_{ij}$.
    \[
    \frac{\partial f_{ij}(\mathbf{U}_p)}{\partial \mathbf{X}_{ij}} = \frac{\partial f_{ij}}{\partial \mathbf{U}_{p,ij}} \cdot \frac{\partial \mathbf{U}_{p,ij}}{\partial \mathbf{X}_{ij}}
    \]

    \item \textbf{Apply the Quotient Rule.} We find the derivative of $\mathbf{U}_{p,ij} = \mathbf{X}_{ij} / g$ with respect to $\mathbf{X}_{ij}$ using the quotient rule $(\frac{u}{v})' = \frac{u'v - uv'}{v^2}$.
    \[
    \frac{\partial \mathbf{U}_{p,ij}}{\partial \mathbf{X}_{ij}} = \frac{1 \cdot g - \mathbf{X}_{ij} \cdot \frac{\partial g}{\partial \mathbf{X}_{ij}}}{g^2} = \frac{1}{g} - \frac{\mathbf{X}_{ij}}{g^2} \frac{\partial g}{\partial \mathbf{X}_{ij}}
    \]

    \item \textbf{Substitute and Combine.} Now, substitute the result from step (3) into step (2), and then the result of that into step (1).
    \[
    \frac{\partial h_{ij}}{\partial \mathbf{X}_{ij}} = \frac{\partial g}{\partial \mathbf{X}_{ij}} f_{ij}(\mathbf{U}_p) + g \cdot \left[ \frac{\partial f_{ij}}{\partial \mathbf{U}_{p,ij}} \left( \frac{1}{g} - \frac{\mathbf{X}_{ij}}{g^2} \frac{\partial g}{\partial \mathbf{X}_{ij}} \right) \right]
    \]

    \item \textbf{Simplify and Rearrange.} Distribute the outer $g$ into the brackets.
    \[
    \frac{\partial h_{ij}}{\partial \mathbf{X}_{ij}} = \frac{\partial g}{\partial \mathbf{X}_{ij}} f_{ij}(\mathbf{U}_p) + \frac{g}{g} \frac{\partial f_{ij}}{\partial \mathbf{U}_{p,ij}} - \frac{g \cdot \mathbf{X}_{ij}}{g^2} \frac{\partial f_{ij}}{\partial \mathbf{U}_{p,ij}} \frac{\partial g}{\partial \mathbf{X}_{ij}}
    \]
    The terms simplify, and we can replace $\frac{\mathbf{X}_{ij}}{g}$ with its definition, $\mathbf{U}_{p,ij}$.
    \[
    \frac{\partial h_{ij}}{\partial \mathbf{X}_{ij}} = \frac{\partial g}{\partial \mathbf{X}_{ij}} f_{ij}(\mathbf{U}_p) + \frac{\partial f_{ij}}{\partial \mathbf{U}_{p,ij}} - \mathbf{U}_{p,ij} \frac{\partial f_{ij}}{\partial \mathbf{U}_{p,ij}} \frac{\partial g}{\partial \mathbf{X}_{ij}}
    \]
    Finally, we group the terms containing $\frac{\partial g}{\partial \mathbf{X}_{ij}}$ to arrive at the theorem's statement.
    \[
    \frac{\partial h_{ij}}{\partial \mathbf{X}_{ij}} = \frac{\partial f_{ij}}{\partial \mathbf{U}_{p,ij}} + \frac{\partial g}{\partial \mathbf{X}_{ij}} \left( f_{ij}(\mathbf{U}_p) - \mathbf{U}_{p,ij} \frac{\partial f_{ij}}{\partial \mathbf{U}_{p,ij}} \right)
    \]
\end{enumerate}
This completes the proof.
\end{proof}

\subsection{Propostion 1 Proof}
\label{derivation_3}

\begin{proof}
The result follows directly from applying the chain rule to $s(Z(\mathbf{X}))$.
\[
\frac{\partial s}{\partial \mathbf{X}_{ij}} = \frac{ds}{dZ} \cdot \frac{\partial Z}{\partial \mathbf{X}_{ij}} = \frac{d}{dZ}\left(\frac{\texttt{FP4 max}}{Z}\right)\frac{\partial Z}{\partial \mathbf{X}_{ij}} = -\frac{\texttt{FP4 max}}{Z(\mathbf{X})^2} \frac{\partial Z}{\partial \mathbf{X}_{ij}}
\]
\end{proof}

\subsection{Absmax gradient derivation}
\label{derivation_4}

\begin{proof}
Let $(i^*, j^*)$ be the index of the element with the maximum absolute value, such that $Z(\mathbf{X}) = |\mathbf{X}_{i^*j^*}|$. We first find the gradient of $Z(\mathbf{X})$. The derivative of the absolute value function is the sign function, $\frac{d|x|}{dx} = \sign(x)$. The derivative is non-zero only when we differentiate with respect to the element $\mathbf{X}_{i^*j^*}$ itself. This can be expressed precisely using the Kronecker delta:
\[
\frac{\partial Z}{\partial \mathbf{X}_{ij}} = \frac{\partial |\mathbf{X}_{i^*j^*}|}{\partial \mathbf{X}_{ij}} = \sign(\mathbf{X}_{i^*j^*}) \cdot \delta_{ii^*}\delta_{jj^*}
\]
Substituting this result into the formula from Theorem 2 completes the proof.
\[
\frac{\partial s}{\partial \mathbf{X}_{ij}} = -\frac{\text{FP4 max}}{Z(\mathbf{X})^2} \frac{\partial Z}{\partial \mathbf{X}_{ij}} = -\frac{\text{FP4 max}}{Z(\mathbf{X})^2} \left( \sign(\mathbf{X}_{i^*j^*}) \cdot \delta_{ii^*}\delta_{jj^*} \right)
\]
\end{proof}

\subsection{Softmax gradient derivation}
\label{derivation_5}

\begin{proof}
We first find the gradient of $Z(\mathbf{X})$ by applying the chain rule multiple times.
\begin{align*}
\frac{\partial Z}{\partial \mathbf{X}_{ij}} &= \frac{\partial}{\partial \mathbf{X}_{ij}} \left[ \frac{1}{\beta} \log\left(\sum_{k,l} e^{\beta|\mathbf{X}_{kl}|}\right) \right] \\
&= \frac{1}{\beta} \cdot \frac{1}{\sum_{k,l} e^{\beta|\mathbf{X}_{kl}|}} \cdot \frac{\partial}{\partial \mathbf{X}_{ij}}\left(e^{\beta|\mathbf{X}_{ij}|}\right) \\
&= \frac{1}{\beta} \cdot \frac{1}{\sum_{k,l} e^{\beta|\mathbf{X}_{kl}|}} \cdot \left(e^{\beta|\mathbf{X}_{ij}|} \cdot \beta \cdot \sign(\mathbf{X}_{ij})\right) \\
&= \frac{e^{\beta|\mathbf{X}_{ij}|}}{\sum_{k,l} e^{\beta|\mathbf{X}_{kl}|}} \cdot \sign(\mathbf{X}_{ij})
\end{align*}
The fractional term is the definition of the softmax function applied to the scaled, absolute values of the tensor elements. Thus:
\[
\frac{\partial Z}{\partial \mathbf{X}_{ij}} = \softmax(\beta|\mathbf{X}|)_{ij} \cdot \sign(\mathbf{X}_{ij})
\]
Substituting this dense gradient back into the formula from Theorem 2 completes the proof.
\[
\frac{\partial s}{\partial \mathbf{X}_{ij}} = -\frac{\texttt{FP4 max}}{Z(\mathbf{X})^2} \frac{\partial Z}{\partial \mathbf{X}_{ij}} = -\frac{\texttt{FP4 max}}{Z(\mathbf{X})^2} \left( \softmax(\beta|\mathbf{X}|)_{ij} \cdot \sign(\mathbf{X}_{ij}) \right)
\]
\end{proof}

\subsection{Tensor reconstruction error with MXFP4 format}

\begin{figure}[htp!]
    \centering
    \begin{subfigure}{0.49\textwidth}
        \includegraphics[width=\linewidth]{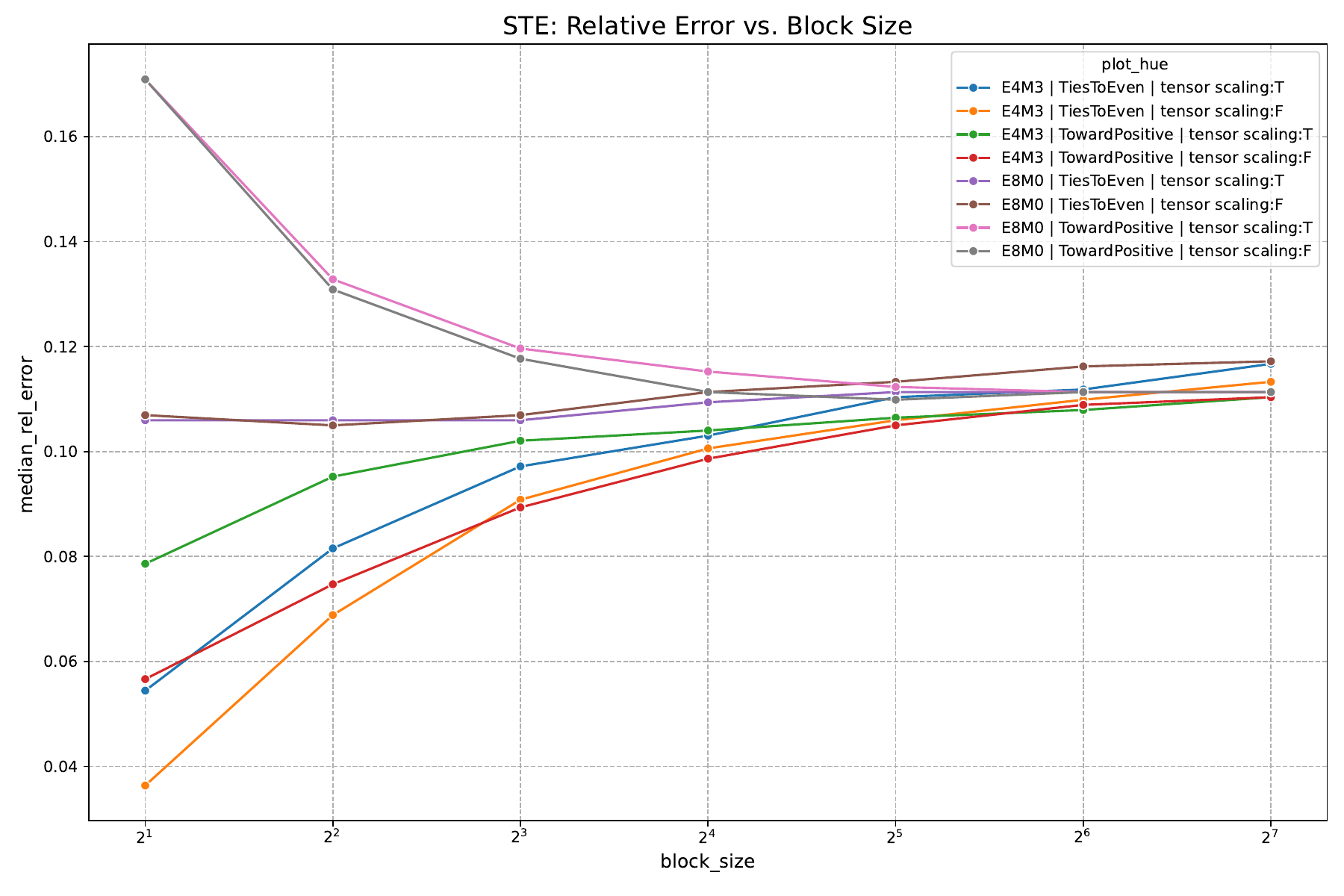}
        \caption{E4M3 vs E8M0}
        \label{fig:exp1-mean}
    \end{subfigure}
    \hfill
    \begin{subfigure}{0.49\textwidth}
        \includegraphics[width=\linewidth]{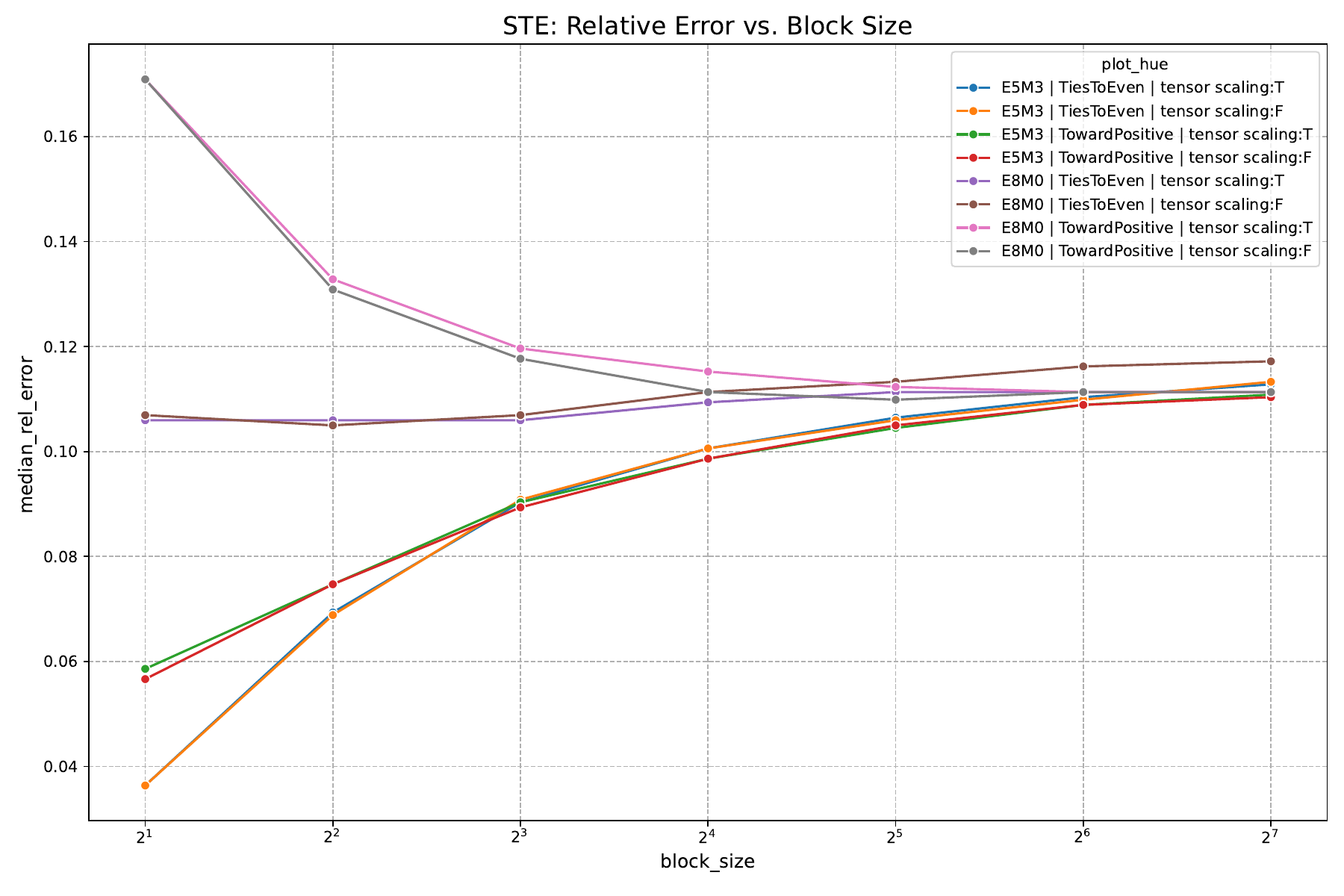}
        \caption{E8M0 vs UE5M3}
        \label{fig:exp1-median}
    \end{subfigure}
    \caption{Reconstruction error using STE as a function of block size.}
    \label{fig:exp1}
\end{figure}

\paragraph{Reconstruction error} We first consider the reconstruction error, i.e., $ \vert \mathbf{X}-\frac{1}{s_q} Q(s_q \cdot \mathbf{X})\vert$ for different choices of $k$, rounding modes of $s$, block sizes, and max functions $Z(\mathbf{X})$. We illustrate different slices of the relative error $\frac{\vert \mathbf{X}-\frac{1}{s_q} Q(s_q \cdot \mathbf{X})\vert}{\vert \mathbf{X}\vert}$. Figure~\ref{fig:exp1} shows the reconstruction error for the Straight-Through Estimator (STE) as a function of block size. As expected, the error decreases as the block size increases.

\begin{figure}[htp!]
    \centering
    \begin{subfigure}{0.49\textwidth}
        \includegraphics[width=\linewidth]{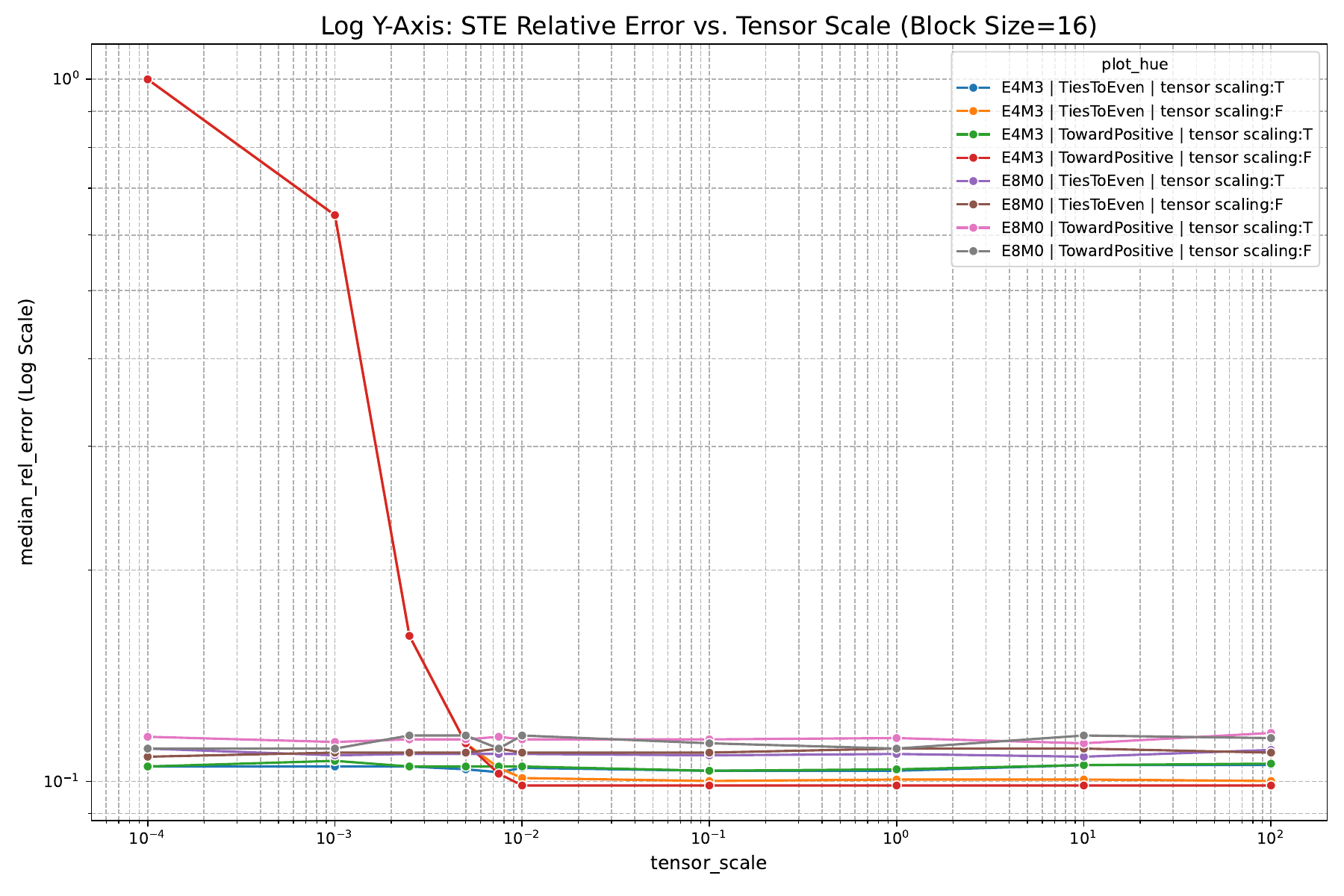}
        \caption{E4M3 vs E8M0}
        \label{fig:exp2-mean}
    \end{subfigure}
    \hfill
    \begin{subfigure}{0.49\textwidth}
        \includegraphics[width=\linewidth]{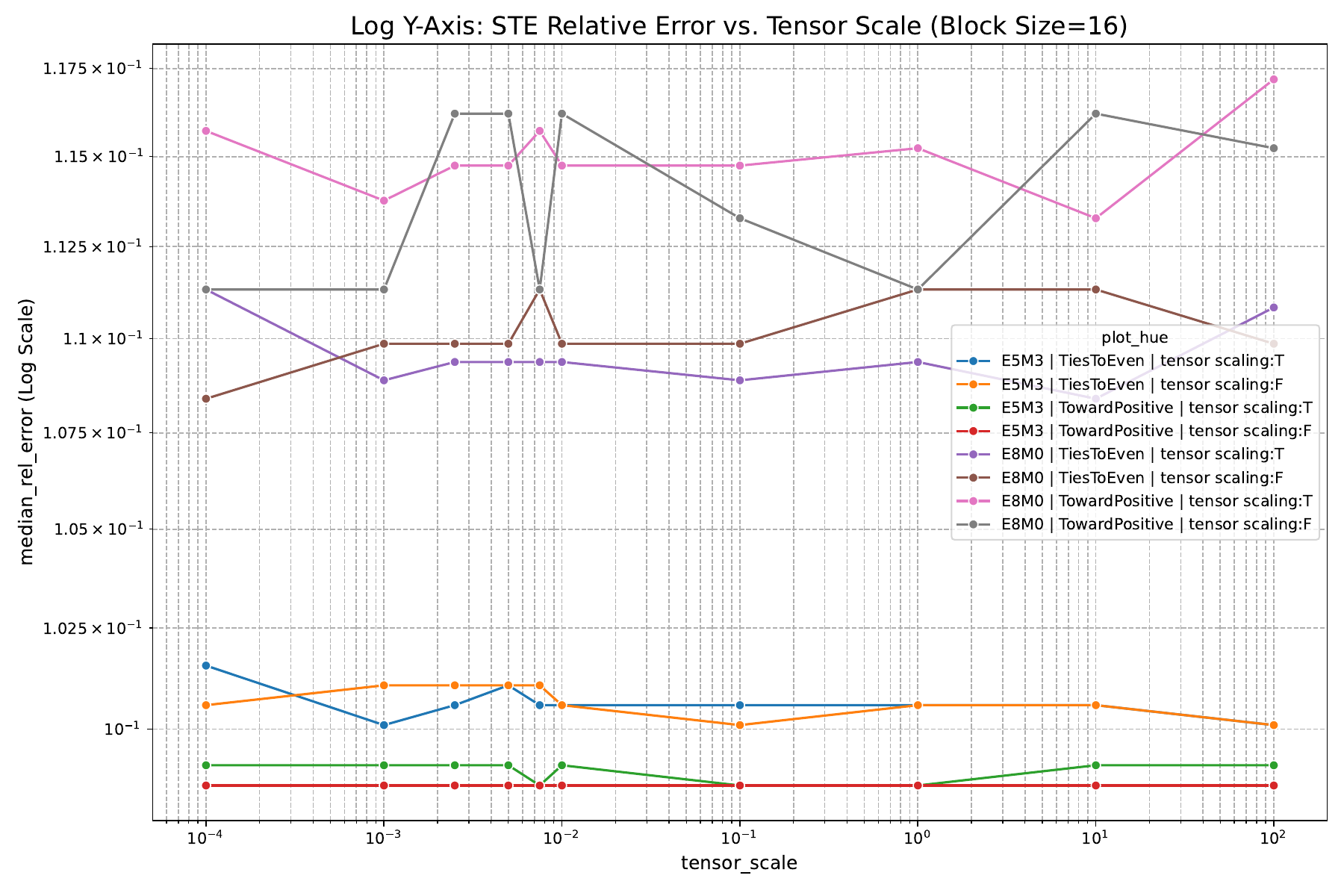}
        \caption{E8M0 vs UE5M3}
        \label{fig:exp2-median}
    \end{subfigure}
    \caption{Reconstruction error using STE as a function of tensor scale (Block Size = 16).}
    \label{fig:exp2}
\end{figure}

\paragraph{Experiment 2: STE Error vs. Tensor Scale} Figure~\ref{fig:exp2} illustrates the impact of the input tensor's scale on the STE reconstruction error, plotted on a log-log scale. These plots show comparisons for a fixed block size of 16.

\begin{figure}[htp!]
    \centering
    \begin{subfigure}{0.49\textwidth}
        \includegraphics[width=\linewidth]{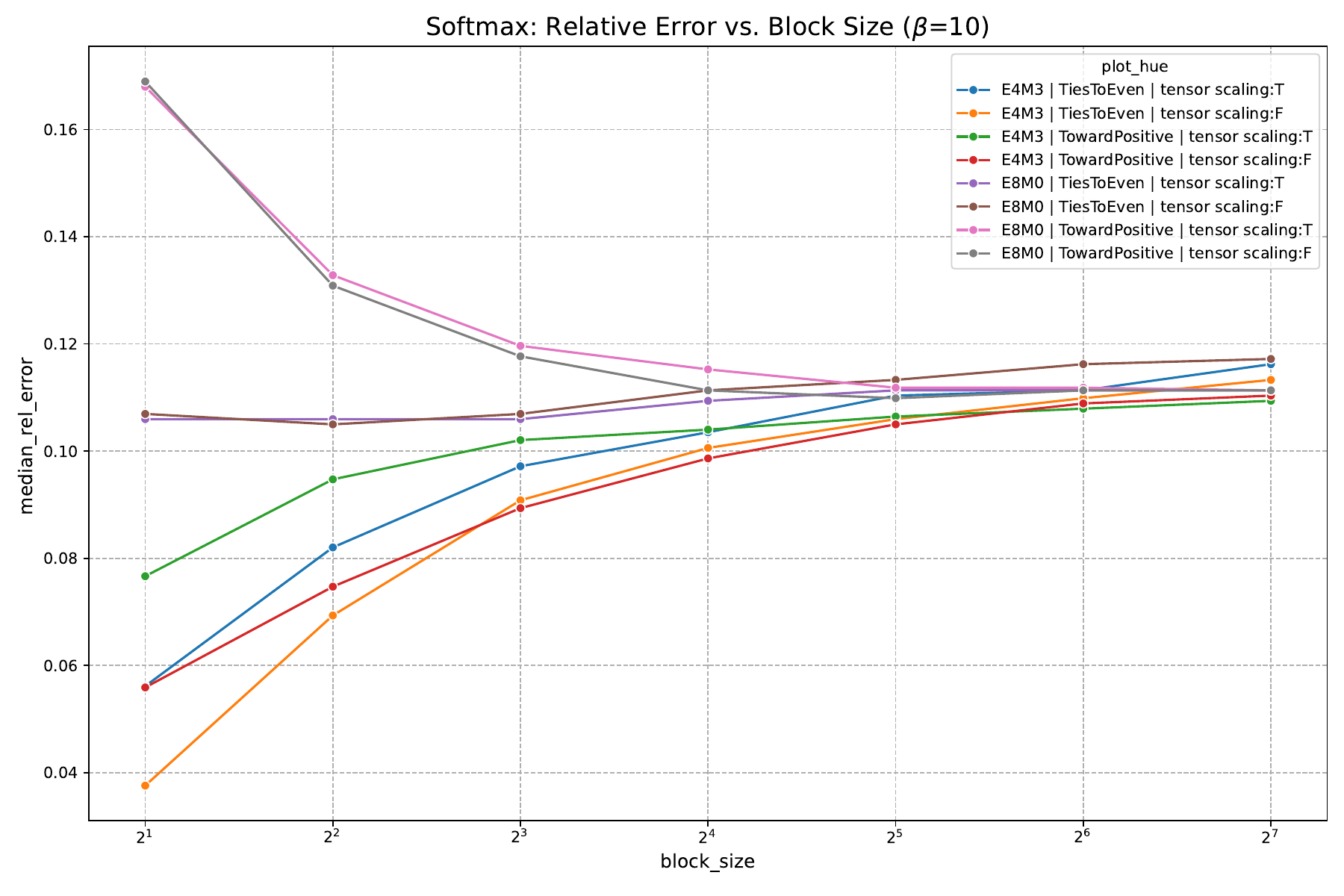}
        \caption{E4M3 vs E8M0}
        \label{fig:exp3-mean}
    \end{subfigure}
    \hfill
    \begin{subfigure}{0.49\textwidth}
        \includegraphics[width=\linewidth]{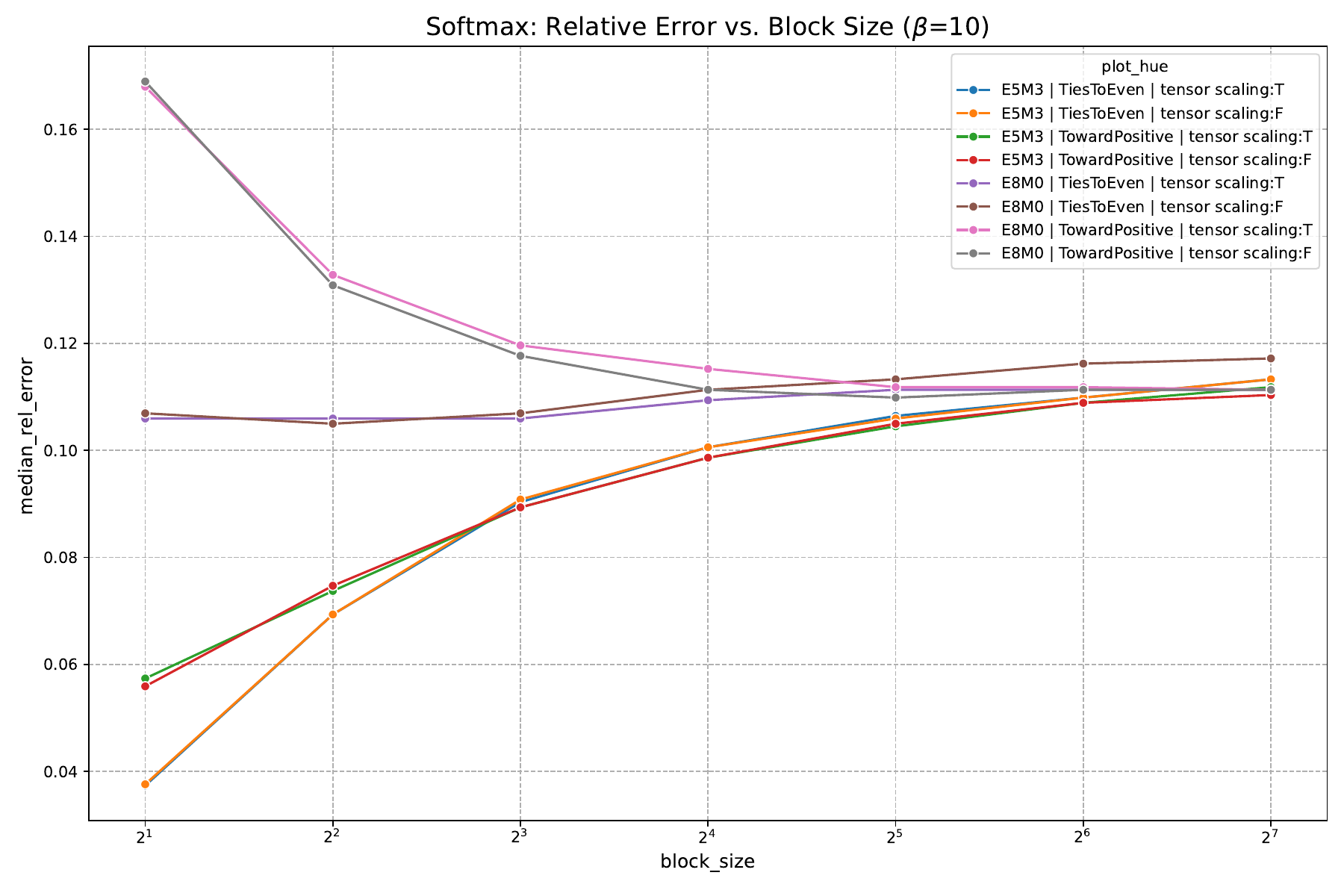}
        \caption{E8M0 vs UE5M3}
        \label{fig:exp3-median}
    \end{subfigure}
    \caption{Reconstruction error using Softmax approximation as a function of block size.}
    \label{fig:exp3}
\end{figure}

\paragraph{Experiment 3: Softmax Error vs. Block Size} Similar to the first experiment, Figure~\ref{fig:exp3} shows the reconstruction error for the Softmax approximation as a function of block size, comparing different data formats.

\newpage

\begin{figure}[htp!]
    \centering
    \begin{subfigure}{0.49\textwidth}
        \includegraphics[width=\linewidth]{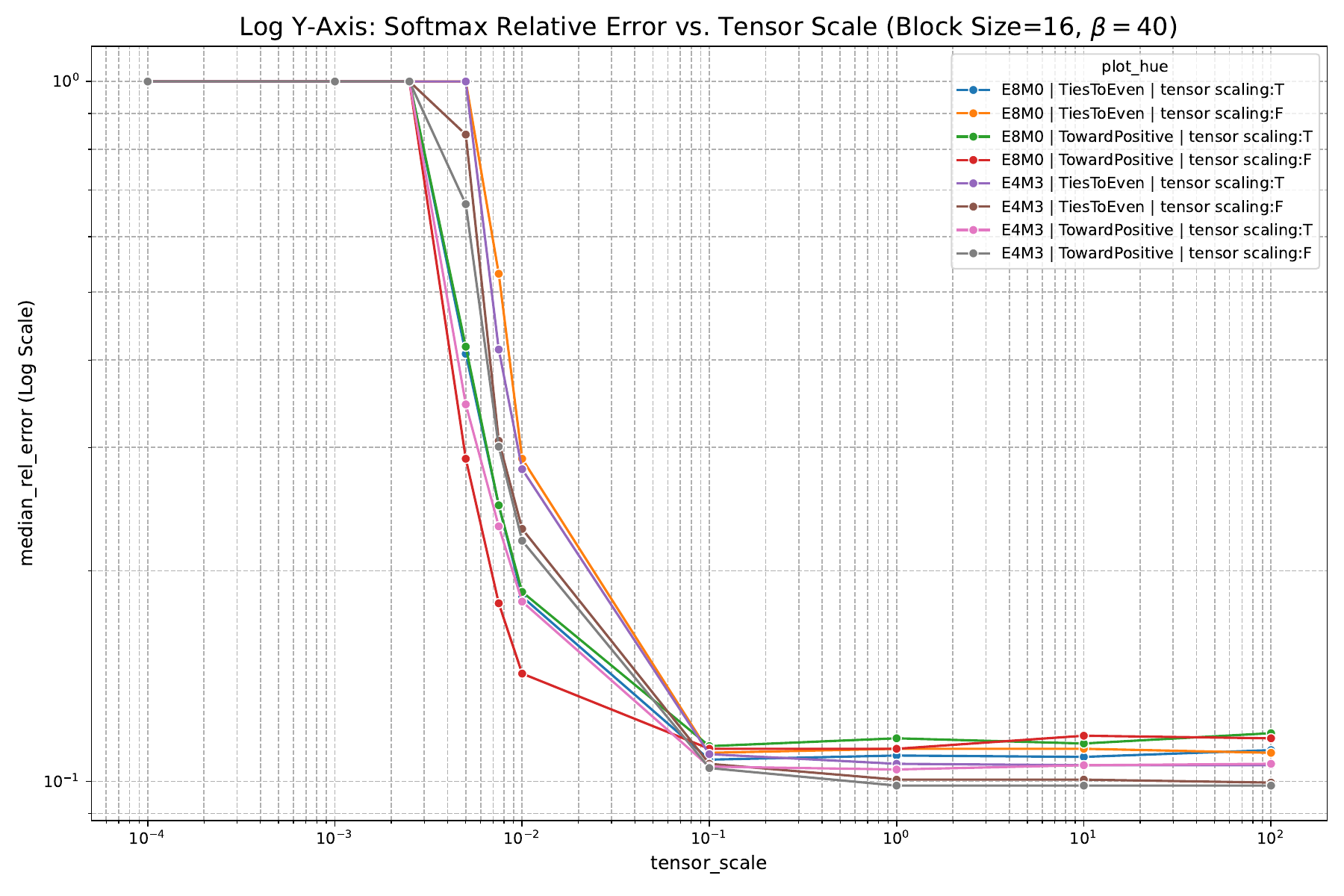}
        \caption{E4M3 vs E8M0}
        \label{fig:exp4-mean}
    \end{subfigure}
    \hfill
    \begin{subfigure}{0.49\textwidth}
        \includegraphics[width=\linewidth]{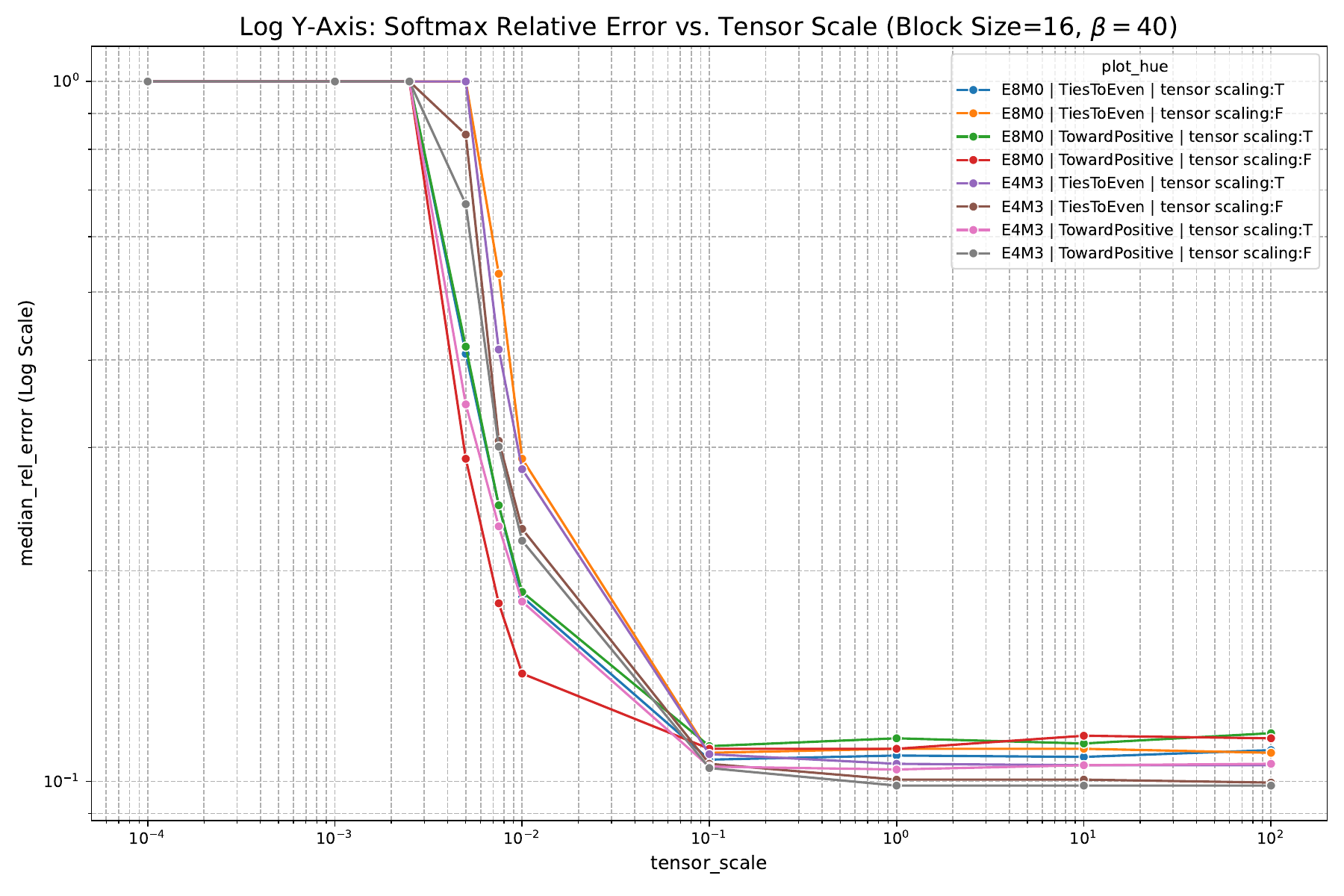}
        \caption{E8M0 vs UE5M3}
        \label{fig:exp4-median}
    \end{subfigure}
    \caption{Reconstruction error using Softmax approximation as a function of tensor scale (Block Size = 16, $\beta=40$).}
    \label{fig:exp4}
\end{figure}

\paragraph{Experiment 4: Softmax Error vs. Tensor Scale} Figure~\ref{fig:exp4} shows the effect of tensor scale on the Softmax approximation for a fixed block size of 16 and a $\beta$ value of 40.

\begin{figure}[htp!]
    \centering
    \begin{subfigure}{0.49\textwidth}
        \includegraphics[width=\linewidth]{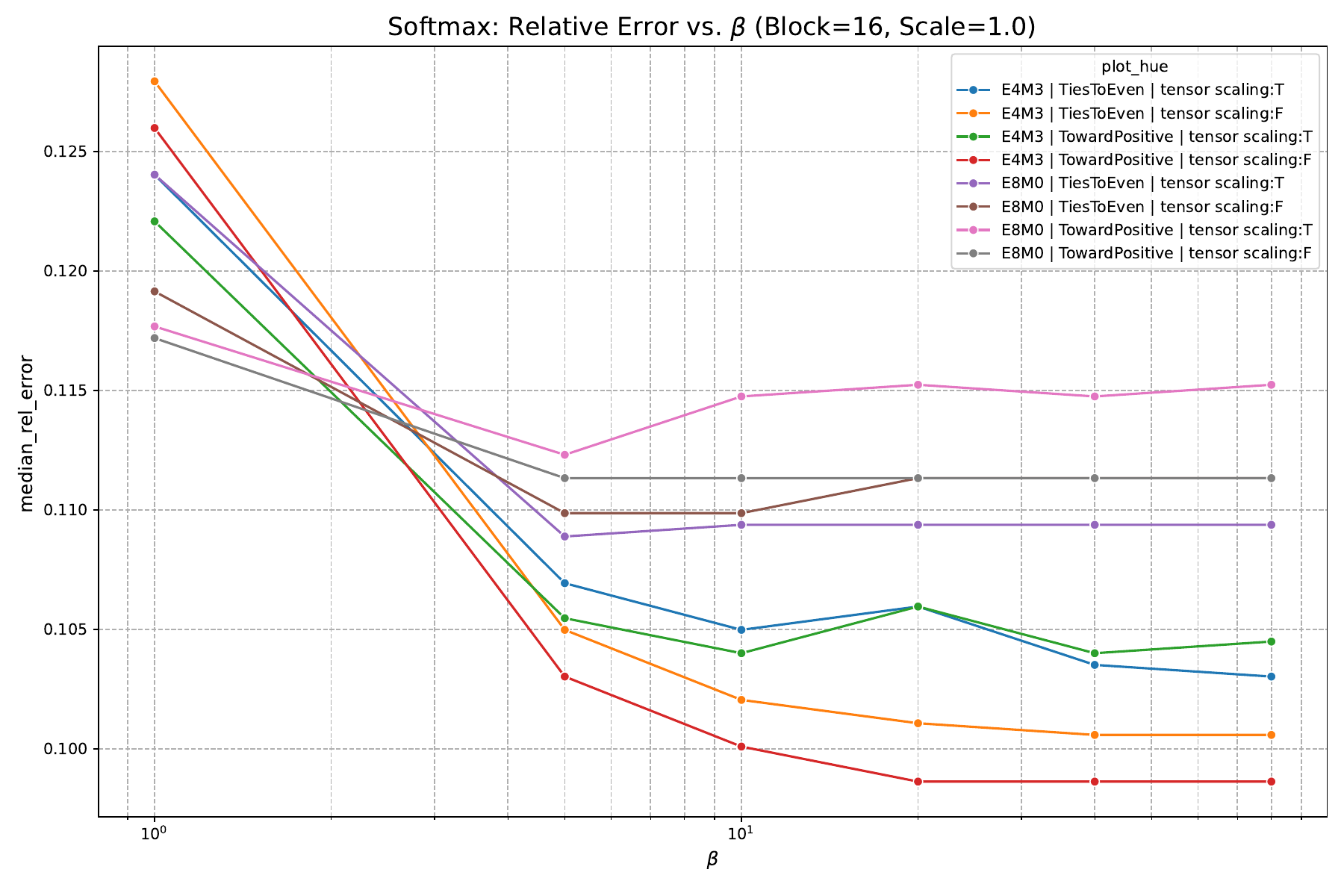}
        \caption{E4M3 vs E8M0}
        \label{fig:exp5-mean}
    \end{subfigure}
    \hfill
    \begin{subfigure}{0.49\textwidth}
        \includegraphics[width=\linewidth]{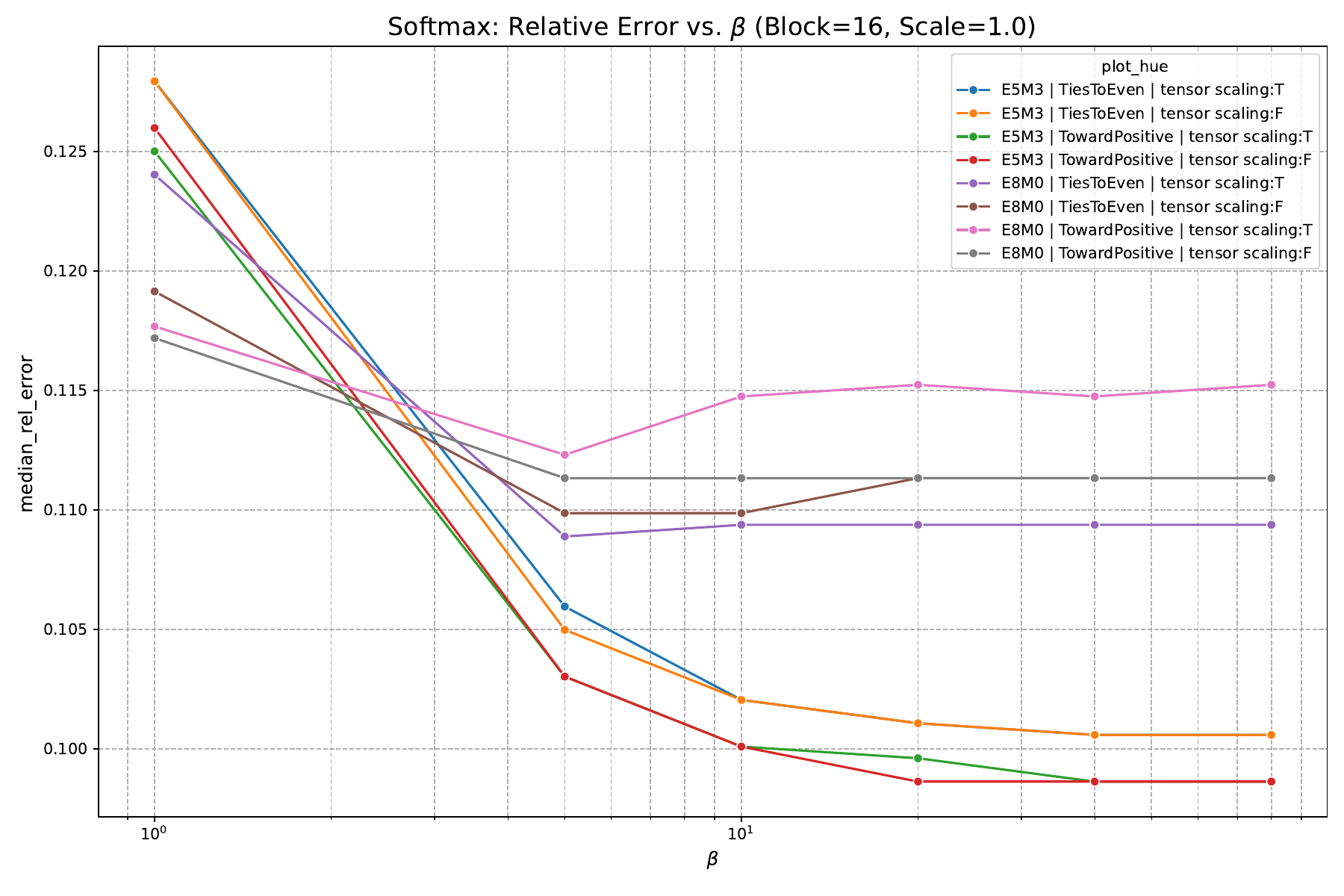}
        \caption{E8M0 vs UE5M3}
        \label{fig:exp5-median}
    \end{subfigure}
    \caption{Reconstruction error using Softmax approximation as a function of the $\beta$ parameter.}
    \label{fig:exp5}
\end{figure}

\paragraph{Experiment 5: Softmax Sensitivity to $\beta$} Finally, Figure~\ref{fig:exp5} analyzes the sensitivity of the Softmax approximation to the inverse temperature parameter, $\beta$. The comparison highlights how tuning $\beta$ affects the reconstruction error for different formats.

\subsection{Things we tried but didn't work}

\paragraph{Conditional Block-wise Scaling}
As scaling factors have limited range, we found in our initial experiments that the \texttt{E4M3} format tends to stall during training, which is caused by underflow due to its limited range compared to \texttt{E8M0}. We propose a conditional scaling strategy, where the choice is determined by comparing the dynamic range of the data's scales, $\text{DR}_{\text{data}} = \frac{g}{\tilde{g}}$, with the intrinsic dynamic range of the target scale format, $\text{DR}_{\text{format}} = \frac{\texttt{E4M3}_{\text{max}}}{\texttt{E4M3}_{\text{min}}}$. Here, $g = \max_p\{m_{p}\}$ and $\tilde{g} = \min_p\{m_{p}\}$.

\paragraph{Case 1: $\text{DR}_{\text{data}} \le \text{DR}_{\text{format}}$ (Ideal Multiplicative Scaling)}
If the data's dynamic range fits within the format's range, we can compute a single constant $C$ using the geometric mean to center the scales within the target range:
$$ C = \frac{\sqrt{\texttt{E4M3}_{\text{max}}\cdot {\tilde{g}} \cdot \texttt{E4M3}_{\text{min}}\cdot{g}}}{\texttt{FP4}_{\text{max}}} $$
The full forward pass for an element, including the final de-normalization, is:
$$ h_{ij}(\mathbf{X}) = \frac{  C}{q(C \cdot s_{p})} Q\left( \frac{q(C \cdot s_{p})}{C} \cdot \mathbf{X}_{ij} \right) $$
where $q(\cdot)$ is the quantization function for the scales (e.g., rounding to the nearest \texttt{E4M3} value). 

\paragraph{Case 2: $\text{DR}_{\text{data}} > \text{DR}_{\text{format}}$ (Affine Mapping fallback)}
If the scaled dynamic range is too wide, we resort to using an affine transformation to map $s_p \in [\frac{\texttt{FP4}_{\text{max}}}{g},\frac{\texttt{FP4}_{\text{max}}}{\tilde{g}}]$ to the range $[\texttt{E4M3}_{\text{min}}, \texttt{E4M3}_{\text{max}}]$. The affine parameters are:
$$ a = \frac{\texttt{E4M3}_{\text{max}} - \texttt{E4M3}_{\text{min}}}{\frac{\texttt{FP4}_{\text{max}}}{\tilde{g}} - \frac{\texttt{FP4}_{\text{max}}}{g}}, \quad b = \texttt{E4M3}_{\text{max}} - a \cdot \frac{\texttt{FP4}_{\text{max}}}{\tilde{g}} $$
The scale to be quantized is $\tilde s_{p} = a \cdot s_{p} + b$. We can then combine this with tensor scaling to achieve a reasonable quantisation:
$$ h_{ij}(\mathbf{X}) = \frac{g\cdot \texttt{E4M3}_{\text{max}}}{\texttt{FP4}_{\text{max}}\cdot q(\tilde s_{p})} Q\left( q(\tilde s_{p}) \cdot \frac{\mathbf{X}_{ij} \cdot \texttt{FP4}_{\text{max}}}{g\cdot \texttt{E4M3}_{\text{max}}} \right) $$ 

In the above setting, we're mapping the scale $s_p$ to the full range of \texttt{E4M3}, however due to the affine mapping we may lose precision for cases when $m_p \ll g$, since the term $\frac{\mathbf{X}_{ij} \cdot \texttt{FP4}_{\text{max}}}{g\cdot \texttt{E4M3}_{\text{max}}}$ will not have the full $\texttt{FP4}_{\text{max}}$ range. We motivate this trade-off with the observation that \texttt{NVFP4} has a block-size of 16, implying that having a well-represented scale outweighs the block accuracy.

When we tested the above on CIFAR10 as a unit test for \texttt{E4M3} we couldn't get anywhere near convergence.

\paragraph{Sigmoid approximation}

Let $\mathcal{V} = \{v_1, \dots, v_n\}$ denote FP4 (E2M1) levels. Define intervals $I_i = (v_i, v_{i+1}]$, $i=1,\dots,n-1$, with
\[
c_i = \frac{v_i + v_{i+1}}{2}, \quad 
\Delta_i = v_{i+1}-v_i, \quad 
\gamma_i = \frac{12}{\Delta_i}.
\]

For $x \in I_i$, let
\[
z_i(x) = \frac{(x-c_i)\gamma_i}{T}, \quad
w(x) = \sigma(z_i(x)) = \frac{1}{1+e^{-z_i(x)}}.
\]

\begin{proposition}[Smooth Quantization Properties]
Let $Q(x)$ be defined as above. Then:
\begin{enumerate}
    \item The forward mapping $Q(x)= v_i + w(x)\Delta_i$ is a smooth interpolation between $v_i$ and $v_{i+1}$ using a sigmoid.
    \item Its derivative is
    \[
    Q'(x) = \Delta_i \cdot \sigma(z_i)(1 - \sigma(z_i)) \cdot \frac{\gamma_i}{T} = \frac{12}{T} \, \sigma(z_i)(1-\sigma(z_i)).
    \]
    \item In the limit $T \to 0$, $Q(x)$ converges to the standard ties-to-even quantization:
    \[
    \lim_{T\to 0} Q(x) = \begin{cases} 
    v_i, & x \le c_i \\ 
    v_{i+1}, & x > c_i 
    \end{cases}.
    \]
\end{enumerate}

\begin{proof}
The forward mapping is linear in $v_i$ and $v_{i+1}$ with a weight $w(x) \in (0,1)$ from the sigmoid, so it is smooth and bounded by $v_i$ and $v_{i+1}$.  

For the derivative:
\[
Q'(x) = \frac{d}{dx}\big(v_i + w(x)\Delta_i\big) = \Delta_i \frac{dw}{dx} = \Delta_i \frac{dw}{dz_i}\frac{dz_i}{dx}.
\]
Since $w = \sigma(z_i)$, we have $\frac{dw}{dz_i} = \sigma(z_i)(1-\sigma(z_i))$, and $dz_i/dx = \gamma_i/T$, giving
\[
Q'(x) = \Delta_i \cdot \sigma(z_i)(1-\sigma(z_i)) \cdot \frac{\gamma_i}{T} = \frac{12}{T} \, \sigma(z_i)(1-\sigma(z_i)).
\]

Finally, as $T \to 0$, the sigmoid becomes a step function at $c_i$:
\[
\sigma\Big(\frac{(x-c_i)\gamma_i}{T}\Big) \to
\begin{cases} 0,& x<c_i \\ 1,& x>c_i \end{cases},
\]
so $Q(x)$ reduces to ties-to-even quantization:
\[
Q(x) \to
\begin{cases} v_i, & x \le c_i \\ v_{i+1}, & x > c_i \end{cases}.
\]
\end{proof}
\end{proposition}
We tried this gradient adjustment, we expected it would provide a significant performance benefit, however this was not the case in early experiments (MNIST, gaussian regression, CIFAR10, llama 9M). Hence on lower level implementations, the additional complexity is not justified. The additional complexity is   $\mathcal{O}(np\log k)$, with $\mathcal{O}(n)$ extra memory. $p$ denotes the number of polynomials used to evaluate the exponential function used in the sigmoid.

\end{document}